%% file: main.tex
\documentclass{article}

\usepackage[nonatbib,final]{neurips_2021}

\usepackage[utf8]{inputenc} 
\usepackage[T1]{fontenc}    
\usepackage[makeroom]{cancel}
\usepackage{graphicx}
\usepackage{hyperref}       
\usepackage{url}            
\usepackage{booktabs}       
\usepackage{amsfonts}       
\usepackage{nicefrac}       
\usepackage{microtype}      
\usepackage{xcolor}         
\usepackage{amsmath, amssymb}
\usepackage{dsfont}
\usepackage{wrapfig}
\usepackage{algorithmic}
\usepackage{algorithm}
\usepackage{enumitem}

\include{macro}

\title{Exponential Graph is Provably Efficient for Decentralized Deep Training}

\author{Bicheng Ying$^{1,3} $\thanks{Equal Contribution. Corresponding Author: Wotao Yin}, Kun Yuan$^{2*}$, Yiming Chen$^{2*}$, Hanbin Hu$^4$, Pan Pan$^2$, Wotao Yin$^2$ \\

$^1$ University of California, Los Angeles $^2$ DAMO Academy, Alibaba Group  \\

$^3$ Google Inc. $^4$ University of California, Santa Barbara \\
\texttt{ybc@ucla.edu, $\{$kun.yuan, charles.cym$\}$@alibaba-inc.com, } \\ 
\texttt{hanbinhu@ucsb.edu, $\{$panpan.pp, wotao.yin$\}$@alibaba-inc.com} \\

}

\begin{document}

\maketitle

\begin{abstract}
Decentralized SGD is an emerging training method for deep learning known for its much less (thus faster) communication per iteration, which relaxes the averaging step in parallel SGD to inexact averaging. The less exact the averaging is, however, the more the total iterations the training needs to take. Therefore, the key to making decentralized SGD efficient is to realize nearly-exact averaging using little communication. This requires a skillful choice of communication topology, which is an under-studied topic in decentralized optimization.

In this paper, we study so-called exponential graphs where every node is connected to $O(\log(n))$ neighbors and $n$ is the total number of nodes. This work proves such graphs can lead to both fast communication and effective averaging simultaneously. We also discover that a sequence of $\log(n)$ one-peer exponential graphs, in which each node communicates to one single neighbor per iteration, can together achieve exact averaging. This favorable property enables one-peer exponential graph to average as effective as its static counterpart but communicates more efficiently. We apply these exponential graphs in decentralized (momentum) SGD to obtain the state-of-the-art balance between per-iteration communication and iteration complexity among all commonly-used topologies. Experimental results on a variety of tasks and models demonstrate that decentralized (momentum) SGD over exponential graphs promises both fast and high-quality training. Our code is implemented through BlueFog and available at \url{https://github.com/Bluefog-Lib/NeurIPS2021-Exponential-Graph}.
\end{abstract}

\vskip -2mm
\section{Introduction}
\vskip -2mm

Efficient distributed training methods across multiple computing nodes are critical for large-scale modern deep learning tasks. 
Parallel stochastic gradient descent (SGD) is a widely-used approach, which, at each iteration, computes a globally averaged gradient either using Parameter-Server \cite{li2014scaling} or All-Reduce \cite{patarasuk2009bandwidth}. Such global coordination  across all nodes in parallel SGD results in either significant bandwidth cost or high latency, which can notably hamper the training scalability. 

Decentralized SGD \cite{nedic2009distributed, chen2012diffusion, lian2017can, assran2019stochastic}  based on \textit{partial averaging} has been one of the promising alternatives to parallel SGD in distributed deep training. Partial averaging, as opposed to the global averaging exploited in parallel SGD, only requires each node to compute the locally averaged model within its neighborhood. Decentralized SGD does not involve any global operations, so it has much lower communication overhead per iteration. The fewer neighbors each node needs to communicate, the more efficient the \textbf{per-iteration communication} is in decentralized SGD. 

\begin{wrapfigure}{R}{0.38\textwidth}
\vspace{-0.6cm}
\begin{center}
    \includegraphics[width=0.38\textwidth]{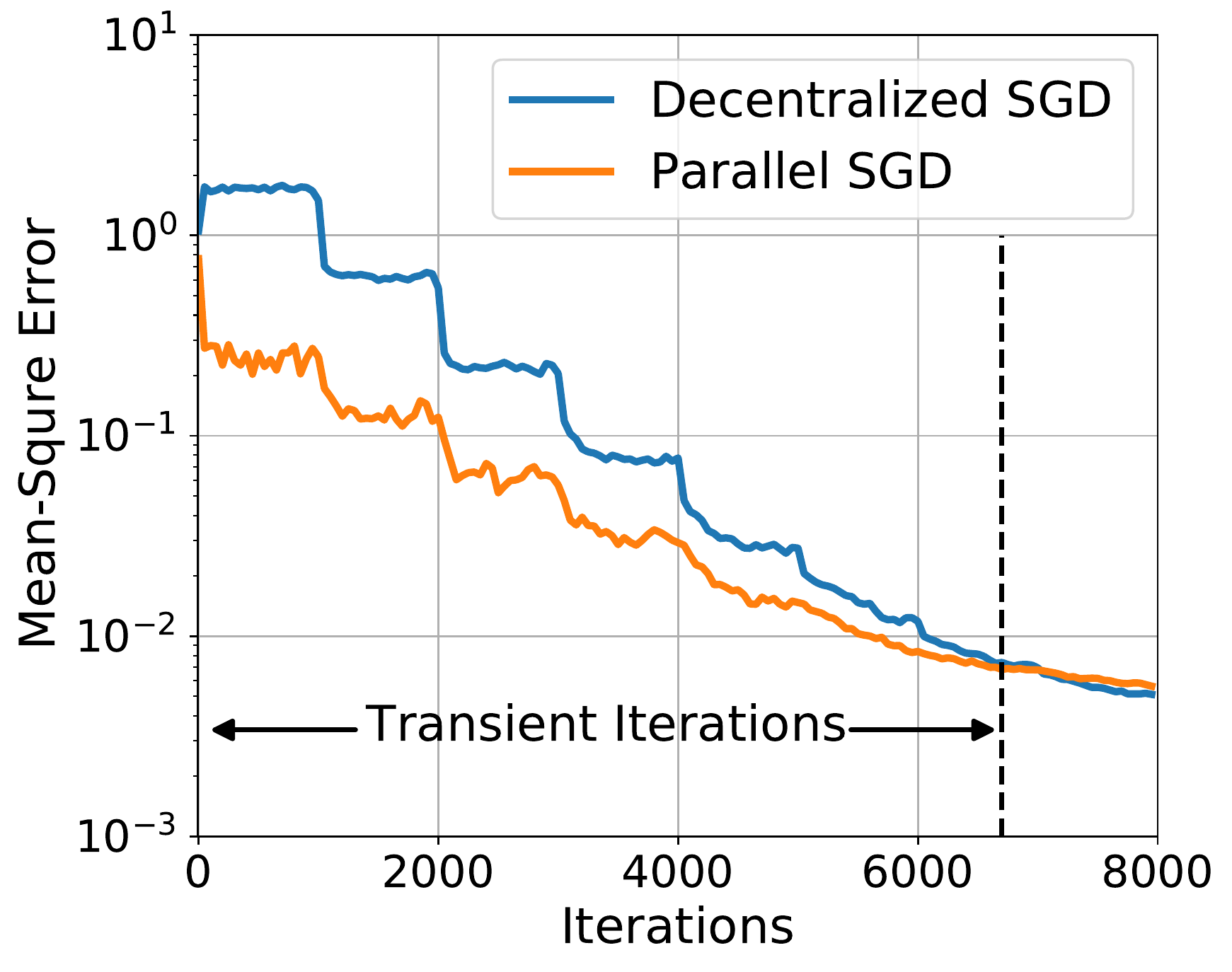}\hspace{-0.5mm}
\end{center}
\vspace{-5mm}
\caption{\small Illustration of transient iters. Experimental setting is in Appendix \ref{apx.sub.converg-compare}. }
\label{fig.tran-iter}
\vspace{-4mm}
\end{wrapfigure}

The reduced communication in decentralized SGD comes with a cost: slower convergence. While it can asymptotically achieve the same convergence linear speedup as parallel SGD \cite{lian2017can,assran2019stochastic,koloskova2020unified,yu2019linear}, i.e., the training speed increases proportionally to the number
of computing nodes (see the definition in Sec.~\ref{sec:dmsgd}),  decentralized SGD requires more iterations to reach that stage due to the ineffectiveness to aggregate information using partial averaging. We refer those iterations before decentralized SGD reaches its linear speedup stage as {\bf transient iterations} (see the definition in Sec.~\ref{sec:dmsgd}), which is an important metric to measure the influence of partial-averaging \cite{pu2019sharp,yuan2021removing} on convergence rate of decentralized SGD. The less effective the partial averaging is, the more transient iterations  decentralized SGD needs to take.  Fig.~\ref{fig.tran-iter} illustrates the transient iterations of decentralized SGD for the logistic regression problem. It is observed that decentralized SGD can asymptotically converge as fast as parallel SGD, but it requires more iterations (i.e., transient iterations) to reach that stage.



Per-iteration communication and transient iterations in decentralized SGD are determined by the network topology (we also use graph interchangeably with topology). The maximum degree of the graph decides the communication cost while the connectivity influences the transient iteration complexity. 
Generally speaking, a sparsely-connected topology communicates cheaply but endows decentralized SGD with more transient iterations due to the less effective information aggregation. A skillful choice of network topology, which  is critical to achieve balance between per-iteration communication and transient iteration complexity,  
is under-studied in literature.

\setlength{\tabcolsep}{4pt}
\begin{table}[t]
\centering
\caption{\small Comparison 
between decentralized (momentum) SGD over (some) various commonly-used topologes. The table assumes homogeneous data distributions across all nodes (which is practical for deep training within a data-center). The comparison for data-heterogeneous scenarios, and with more other topologies, is listed in Appendix \ref{apx.transient}. The smaller the transient iteration complexity is, the faster decentralized algorithms will converge.} 
\vspace{-1mm}

\begin{tabular}{lcccccc}
\toprule
\textbf{Topology}& {Ring } & {Grid }  & {Rand-Graph} & {Rand-Match} & {\bf Static Exp} &{\bf One-peer Exp} \\ \midrule
\textbf{Per-iter Comm. }  & $\Omega(2)$  & $\Omega(4)$  & $\Omega(\frac{n}{2})$  & ${\Omega(1)}$  & $\Omega(\log_2(n))$  & ${\Omega(1)}$  \\ [2mm]
\textbf{Trans. Iters. } & $\Omega(n^7)$ & $\Omega(n^5)$& $\Omega(n^3)$& $-$ & ${\Omega(n^3 \log_2^2(n))}$ & ${\Omega(n^3 \log_2^2(n))}$\\ 
\bottomrule

\end{tabular}
\vspace{-4mm}
\label{tb-main-result}
\end{table}
\setlength{\tabcolsep}{7pt}


This work studies \textbf{exponential graphs} which are empirically successful \cite{assran2019stochastic,wang2019slowmo,kong2021consensus,chen2021accelerating,yuan2021decentlam} but less theoretically understood in deep training. Exponential graphs have two variants. In a {static exponential graph}, each node communicates to {$\lceil \log_2(n) \rceil$ neighbors (see Sec.~\ref{sec:static-expo} and Fig.~\ref{fig:static-one-peer-graph})}. In {one-peer exponential graph}, however, each node cycles through all its neighbors, communicating, only, to a single neighbor per iteration (see Sec.~\ref{sec:dynamic-expo} and Fig.~\ref{fig:static-one-peer-graph}). This paper will first  clarify the connectivity and averaging effectiveness of these exponential graphs, and then apply them to decentralized momentum SGD to obtain the state-of-the-art balance between per-iteration communication and transient iteration complexity among all commonly-used topologies. Our main results (as well as our contributions) are: 

\begin{itemize}[leftmargin=20pt]
\vspace{-2mm}
    \item We prove that the spectral gap, which is used to measure the connectivity of the graph (see the definition in Sec.~\ref{sec:dmsgd}), of the static exponential graph is upper bounded by $O(1/\log_2(n))$. Before us, many literatures (e.g. \cite{kong2021consensus})  claimed its upper bound to be $O(1)$ incorrectly. 
    \vspace{-1mm}
    \item Since one-peer exponential graphs are time-varying, it is difficult to derive their spectral gaps. However, we establish that any $\log_2(n)$ consecutive sequence of one-peer exponential graphs can together achieve {\em exact averaging} 
    when $n$ is a power of $2$. 
    \vspace{-1mm}
    \item With the above results, we establish that one-peer exponential graph, though much sparser than its static counterpart, surprisingly endows decentralized momentum SGD with the \textbf {same} convergence rate as static exponential graph in terms of the best-known bounds.
    \vspace{-1mm}
    \item 
    We derive that exponential graphs achieve $\tilde{\Omega}(1)$\footnote{Notation $\tilde{\Omega}(\cdot)$ hides all logarithm factors.} per-iteration communication and $\tilde{\Omega}(n^3)$ transient iterations, both of which are nearly the best  among other known topologies, see Table \ref{tb-main-result}. The one-peer exponential graph is particularly recommended for decentralized deep training. 
    \vspace{-1mm}
    
    \item We conduct extensive industry-level experiments across different tasks and models with various decentralized methods, graphs, and network size to validate our theoretical results.
\end{itemize}

\vskip -2mm
\section{Revisit Decentralized Momentum SGD and Related Works}
\vskip -2mm
\label{sec:dmsgd}
This section reviews basic concepts and existing results on decentralized momentum SGD.

\textbf{Problem.} Suppose $n$ computing nodes cooperate to solve the distributed optimization problem:
\begin{align}\label{dist-opt}
{\color{black} \min_{x \in \mathbb{R}^d}\  f(x)=\frac{1}{n}\sum_{i=1}^n f_i(x) \quad \mbox{where} \quad f_i(x): = \mathbb{E}_{\xi_i \sim D_i} F(x;\xi_i).}
\end{align}
Function $f_i(x)$ is local to node $i$, and random variable $\xi_i$ denotes the local data that follows distribution $D_i$. We do not assume each distribution $D_i$ is the same across all nodes. 

\begin{figure}[t]
    \centering
    \includegraphics[width=0.85\textwidth]{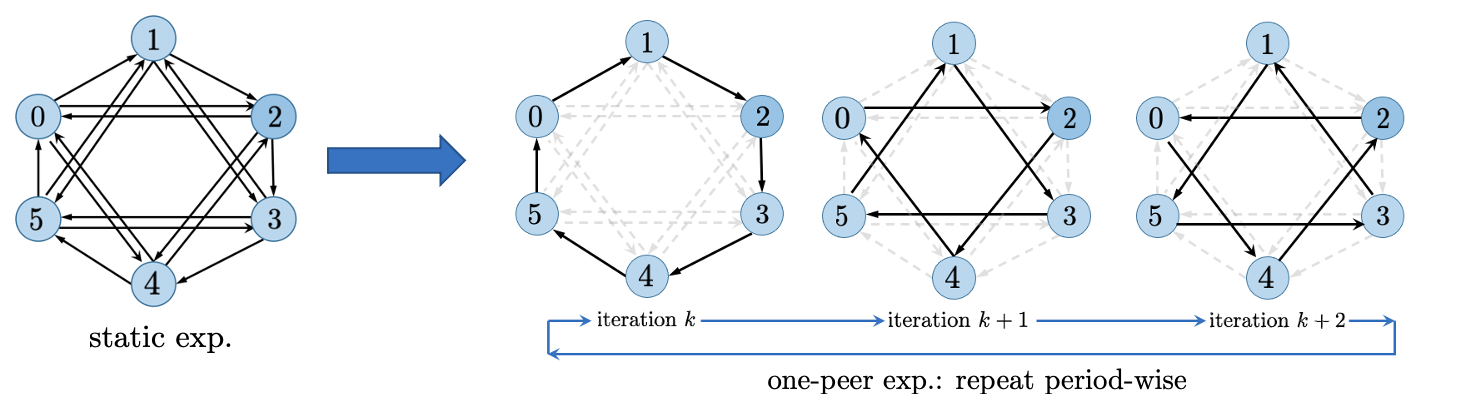} \vspace{-2mm}
    \caption{\small Illustration of the static and one-peer exponential graph. } \vspace{-5mm}
    \label{fig:static-one-peer-graph}
\end{figure}

\noindent \textbf{Network topology and weights.} Decentralized methods are based on partial averaging within neighborhood that is defined by the network topology (see the figure~\ref{fig:static-one-peer-graph} as an example of six nodes). We assume all computing nodes are connected by a (directed or undirected) network topology. 
We define $w_{ij}$, the weight to scale information flowing from node $j$ to node $i$, as follows:
\begin{align}\label{wij}
w_{ij}\left\{
\begin{aligned}
> 0 & \;\;\;\;\;\mbox{if node $j$ is connected to $i$, or $i=j$;} \\
= 0 & \;\;\;\;\;\mbox{otherwise.}\vspace{-2mm}
\end{aligned}\right.
\end{align}
$\mathcal{N}_i:=\{j|w_{ij} > 0\}$ is defined as the set of neighbors of node $i$ which also includes node $i$ itself and the {\em weight matrix} $W := [w_{ij}]_{i,j=1}^{n} \in \mathbb{R}^{n\times n}$ are denoted as a matrix that stacks the weights of all nodes. This matrix $W$ characterizes the sparsity and connectivity of the underlying network topology. 

\begin{wrapfigure}{R}{0.5\textwidth}
\vspace{-0.8cm}
\begin{minipage}{0.5\textwidth}
\begin{algorithm}[H]
        \caption{DmSGD}
        \begin{algorithmic}
        \STATE \textbf{Initialize} $\gamma$, $x^{(0)}_{i}$; let  $m^{(0)}_{i}\hspace{-0.3mm}=\hspace{-0.3mm}0, \beta\hspace{-0.3mm}\in\hspace{-0.3mm}(0,1)$ \vspace{1mm}
        \STATE \textbf{For} $k=0, 1,2,...,T-1$, every node $i$ \textbf{do} \vspace{1mm}
        \STATE $\quad$ Sample weight matrix $W^{(k)}$; \vspace{0.5mm}
        \STATE $\quad$ {\color{black}Update gradient $g_i^{(k)} = \nabla F(x_i^{(k)};\xi_i^{(k)})$;} \vspace{0.5mm}
        \STATE $\quad m^{(k+1)}_i = \sum_{j\in\cN_i} w^{(k)}_{ij} \big(\beta m^{(k)}_{j} +  g_j^{(k)} \big)$;\vspace{0.5mm}
        \STATE $\quad x^{(k+1)}_i  = \sum_{j\in \mathcal{N}_i} w^{(k)}_{ij} \big(x^{(k)}_{j} - \gamma m^{(k)}_j\big)$;
        \end{algorithmic}
        \label{Algorithm: DmSGD}
      \end{algorithm}

 		
 		
 		
\end{minipage}
\vspace{-0.3cm}
\end{wrapfigure}

\noindent \textbf{Decentralized momentum SGD (DmSGD).} There are many variants of decentralized momentum SGD \cite{assran2019stochastic,gao2020periodic,lin2021quasi,yuan2021decentlam}.
This paper will focus on the one proposed by \cite{yu2019linear} (listed in Algorithm \ref{Algorithm: DmSGD}), which imposes an additional partial-averaging over the momentum to achieve further speed up. The topology is allowed to change with iterations.  When $W^k \equiv W$, topology and weight matrix will remain static. 

\noindent \textbf{Assumptions.} We introduce several standard assumptions to facilitate future analysis:


\noindent \textbf{A.1} [{\sc Smoothness}] Each $f_i(x)$ is $L$-smooth, i.e., $\|\nabla f_i(x) - \nabla f_i(y)\| \le L \|x - y\|$ for any $x,y\in \mathbb{R}^d$.

\noindent \textbf{A.2} [{\sc Gradient noise}] The random sample  $\xi_i^{(k)}$ is independent of each other for any $k$ and $i$. We also assume $\mathbb{E}[{\nabla}F(x;\xi_{i})] = \nabla f_i(x)$ and $\mathbb{E}\|{\nabla}F(x;\xi_{i}) - \nabla f_i(x) \|^2 \le  \sigma^2$.

\noindent \textbf{A.3} [{\sc Data heterogeneity}] 
It holds that $\frac{1}{n}\sum_{i=1}^n\|\nabla f_i(x) - \nabla f(x)\|^2 \le b^2$ for any $x\in \RR^d$.

\noindent \textbf{A.4} [{\sc Weight matrix}] The weight matrix $W^{(k)}$ is doubly-stochastic, i.e. $W^{(k)} \mathds{1} = \mathds{1}$ and $\mathds{1}^T W^{(k)}  = \mathds{1}^T$. {\color{black} If $W^{(k)} \equiv W$, we assume $ \rho(W) := \max_{\lambda_i(W)\neq 1}\{|\lambda_i(W)|\} \in (0, 1)$, where $\lambda_i(W)$ is the $i-$th eigenvalue of the matrix $W$.\footnote{When there is no ambiguity, we  simply use $\rho$ instead of $\rho(W)$. Throughout the paper we do NOT use $\rho$ as spectral radius. Instead, it is the second largest eigenvalue in magnitude. Note we cannot directly sort the eigenvalues since $W$ is not necessarily symmetric and the eigenvalue can be a complex number.}}

{\color{black} The quantity $1-\rho$, which is also referred to as the spectral gap of the weight matrix $W$,  measures how well the topology is connected \cite{Seneta1981nonnegative}. }
In the large and sparse topology which is most valuable to deep training, it typically holds that $1 - \rho \to 0$. 

\textbf{Communication overhead.} According to \cite{ben2019demystifying}, global averaging across $n$ nodes either incurs $\Omega(n)$ bandwidth cost via Parameter-Server, or $\Omega(n)$ latency via Ring-Allreduce. In either way, it takes $\Omega(n)$ per-iteration communication time, which is proportional to the network size $n$. As to decentralized methods, we will similarly assume the per-iteration communication time to be $\Omega(\mbox{maximum degree})$. \vspace{-2mm}


\noindent \textbf{Convergence.} Under Assumptions A.1--A.4, DmSGD with static topology will converge at \cite{yu2019linear,koloskova2020unified}:
\begin{align}\label{dmsgd-convergence}
\frac{1}{T}\sum_{k=1}^T  \Ex \|\nabla f(\bvx^{(k)})\|^2  =  O\left( \frac{\sigma^2}{\sqrt{nT} } + \frac{n\sigma^2}{T(1-\rho)} + \frac{ n b^2}{T(1-\rho)^2}\right)  \quad
\end{align}
in which $\bar{x}^{(k)}=\frac{1}{n}\sum_{i=1}^n x^{(k)}_i$. It is worth noting that no analysis in literature, to our knowledge, exists for DmSGD over time-varying topologies with non-convex costs.

\noindent \textbf{Linear speedup.} When $T$ is sufficiently large, the first term $1/\sqrt{nT}$ dominates \eqref{dmsgd-convergence}. This also applies to parallel  SGD. Decentralized and parall SGDs all require $T = \Omega(1/(n\epsilon^2))$ iterations to reach a desired accuracy $\epsilon$, which is inversely proportional to $n$. Therefore, an algorithm is in its linear-speedup stage at $T$th iteration if, for this $T$, the term involving $nT$ is dominating the rate.

\noindent \textbf{Transient iterations}. Transient iterations are referred to those iterations before an algorithm reaches  linear-speedup stage, that is when $T$ is relatively small so non-$nT$ terms still dominate the rate (see illustration in Appendix \ref{apx.transient}). To reach linear speedup, $T$ has to satisfy (derivation in Appendix \ref{apx.transient})
\begin{align}
\mbox{Homogeneous data:}\quad  T = \Omega\left(\frac{n^3}{(1-\rho)^2}\right) \quad \quad  \quad  \mbox{Heterogeneous data:}\quad  T = \Omega\left(\frac{n^3}{(1-\rho)^4}\right) \quad \label{eq-tran-iter-hetero}
\end{align}
which corresponds to the transient iteration complexity in the homo/hetero-geneous data scenarios.

\subsection{Related Works}

\textbf{Decentralized deep training.} Decentralized optimization originates from the control and signal processing community. The first decentralized algorithms on general optimization problems include  decentralized gradient descent \cite{nedic2009distributed}, diffusion \cite{chen2012diffusion,sayed2014adaptive} and dual averaging \cite{duchi2011dual}. 
In the deep learning regime, decentralize SGD, which was established in \cite{lian2017can} to achieve the same linear speedup as parallel SGD in convergence rate, has attracted a lot of attentions. 
Many efforts have been made to extend the algorithm to directed topologies \cite{assran2019stochastic,nedic2014distributed}, time-varying topologies \cite{koloskova2020unified,nedic2014distributed}, asynchronous settings \cite{lian2018asynchronous}, and data-heterogeneous scenarios \cite{tang2018d,xin2020improved,lin2021quasi,yuan2021decentlam}. Techniques such as quantization/compression \cite{alistarh2017qsgd,bernstein2018signsgd,koloskova2019decentralized,koloskova2019decentralized2,tang2019doublesqueeze,liu2020linear}, periodic updates \cite{stich2019local,koloskova2020unified,yu2019linear}, and lazy communication \cite{liu2021decentralized,liu2019communication,chen2018lag} were also integrated into decentralized SGD to further reduce communiation overheads. 

\textbf{Topology influence.} The influence of network topology on decentralized SGD was extensively studied in \cite{koloskova2020unified,sayed2014adaptive,yuan2020influence,nedic2009distributed,nedic2014distributed,kong2021consensus}. 
All these works indicate that a well-connected topology will significantly accelerate decentralized SGD. 
Two directions have been explored to relieve the influence of network topology. 
One line of research proposes new algorithms that are less sensitive to  topologies. For example, \cite{yuan2020influence, huang2021improving,yuan2021removing,tang2018d,alghunaim2021unified} removed data heterogeneity with bias-correction techniques in \cite{yuan2017exact1,li2017decentralized,xin2020improved,lu2020decentralized,zhang2019decentralized}, and \cite{chen2021accelerating,wang2019slowmo,berahas2018balancing,kong2021consensus} utilized periodic global averaging or multiple partial averaging steps. All these methods have improved topology dependence.
 The other line is to investigate topologies that enable   communication-efficient decentralized optimization. \cite{nedic2018network,chow2016expander} examined various topologies (such as ring, grid, torus, expander, etc.) on averaging effectiveness, which, however, are either communication-costly or averaging-ineffective compared to exponential graphs studied in this paper. {\color{black}\cite{nachmias2008critical,benjamini2014mixing,beveridge2016best,boyd2005mixing} studied random graphs (such as Erdos-Renyi random graph and random geometric graph) in which each edge is activated randomly. The randomness of the edge activation can cause a highly unbalanced 
degrees of each node in the graph, which may significantly affect the efficiency in per-iteration communication.}

\textbf{Algorithms with time-varying topologies.} 
Many previous works have studied decentralized algorithms with time-varying topologies. \cite{nedic2014distributed} and \cite{nedic2017achieving} examined the convergence of decentralized (deterministic) gradient descent and gradient tracking under convex scenarios. \cite{di2016next,scutari2019distributed} investigated gradient tracking under non-convex scenarios, but it did not clarify the influence of the time-varying graphs on convergence rate. In the stochastic scenario,  \cite{koloskova2020unified} illustrates how decentralized SGD is influenced by time-varying  topologies in the non-convex scenario. However, its analysis cannot be directly extended to the decentralized momentum SGD studied in this work. 

Another related work is the \textbf{Matcha} method \cite{wang2019matcha} based on disjoint matching decomposition sampling. 
While similar to Matcha, decentralized SGD over one-peer exponential graphs has several fundamental differences. First, one-peer exponential graph is directed while Matcha only supports {\em undirected} and {\em symmetric} matching decomposition. Second, the favorable periodic exact-average property of one-peer exponential graphs only holds when sampled {\em cyclicly}. However, Matcha only supports {\em independent} and {\em random} matching samples in analysis. For these reasons, Matcha cannot cover one-peer exponential graphs (especially when momentum is utilized in decentralized SGD).

\textbf{Note.} This paper considers
deep training within \textbf{high-performance data-center clusters},
in which all GPUs are connected with high-bandwidth channels
and the network topology can be fully controlled. It is \textbf{not} for the wireless network setting in which the topology cannot be changed freely.

\section{Spectral Gap of Static Exponential Graph}
\label{sec:static-expo}

As discussed above, the graph maximum degree decides the per-iteration communication cost while the spectral gap determines the transient iteration complexity (see \eqref{eq-tran-iter-hetero}). It is critical to seek topologies that are both sparse and with large spectral gap $1-\rho$ simultaneously. In this section, we will establish that the static exponential graph, which was first introduced in \cite{assran2019stochastic,lian2017can}, 
is one of such topologies. 

In a static exponential graph, each node is assigned an index from $0$ to $n-1$ and will communicate to neighbors that are $2^0, 2^1,\cdots, 2^{\lfloor \log_2(n-1) \rfloor}$ hops away. The left plot in Fig.~\ref{fig:static-one-peer-graph} illustrates a directed $6$-node exponential network topology. With maximum degree $\lceil \log_2(n) \rceil$ neighbors, partial averaging over the static exponential graph will take $\Omega(\log_2(n))$ communication time per iteration. However, it remains unclear what the spectral gap is for this topology. 



\textbf{Weight matrix associated with static exponential graph} is defined as follows:   
\begin{align}\label{wij-static-exp}
w^{\rm exp}_{ij} = 
\begin{cases}
\frac{1}{\lceil \log_2(n) \rceil + 1} & \mbox{if $\log_2(\mathrm{mod}(j-i,n))$ is an integer or $i=j$} \\
\quad 0 & \mbox{otherwise.}\vspace{-2mm}
\end{cases}
\end{align}
An example weight matrix associated with the static exponential graph in Fig.~\ref{fig:static-one-peer-graph} is in Appendix \ref{apx.sub.exp-weight}. 
The following proposition evaluates the spectral gap $1-\rho$ for weight matrix in \eqref{wij-static-exp}.
\begin{proposition}[\sc Spectral Gap of static Expo]
\label{prop-spectral-gap-expo}
The spectral gap of matrix \eqref{wij-static-exp}, which can also be interpreted as the second largest magnitude of eigenvalues, satisfies {(Proof is in Appendix \ref{apx.sub.exp-spectral-gap})}
\begin{align}\label{sp-gap-static-expo}
1 - \rho(W^{\rm exp}) \, \left\{
\begin{aligned}
    =& \; \frac{2}{1+ \lceil \log_2(n) \rceil},  \;\; \mbox{when n is even number} \\
    <& \; \frac{2}{1+ \lceil \log_2(n) \rceil},  \;\; \mbox{when n is odd number} 
\end{aligned}\right.
\end{align}
In addition, we have $\|W^{\rm exp}-\frac{1}{n}\mathds{1}\mathds{1}^T\|_2 = \rho(W^{\rm exp})$.
\end{proposition}

\begin{remark} \color{black}
    For a general non-symmetric matrix $W$, it typically holds that $\|W-\frac{1}{n}\mathds{1}\mathds{1}^T\|_2 \neq \rho(W)$. Proposition \ref{prop-spectral-gap-expo} establishes  $\|W^{\rm exp}-\frac{1}{n}\mathds{1}\mathds{1}^T\|_2 = \rho(W^{\rm exp})$ for exponential graph.
\end{remark}

\begin{remark} \color{black} The hypercube graph is established in \cite[Chapter  16]{trevisan2017lecture} to have the spectral gap as $1-\rho(W^{\rm HyperCube})=2/(1+\log_2(n))$. While such spectral gap is on the same order as the exponential graph, there are two fundamental differences between these two graphs: (a) the hypercube graph has to be undirected and the corresponding $W$ is symmetric; (b) the number of vertices of hypercube must be a power of $\, 2$, i.e., $n = 2^\tau$ for some positive integer $\tau$. In comparision, the exponential graph is more flexible in the size of the graph structure. 
\end{remark}

\begin{remark} \color{black}
Proposition \ref{prop-spectral-gap-expo} clarifies the spectral gap of the static exponential graph. Many literatures before this work (such as \cite{kong2021consensus}) claimed the spectral gap to be $O(1)$, which is not accurate. 
\end{remark}

\begin{wrapfigure}{R}{0.4\textwidth}
\begin{center}
    \includegraphics[width=0.4\textwidth]{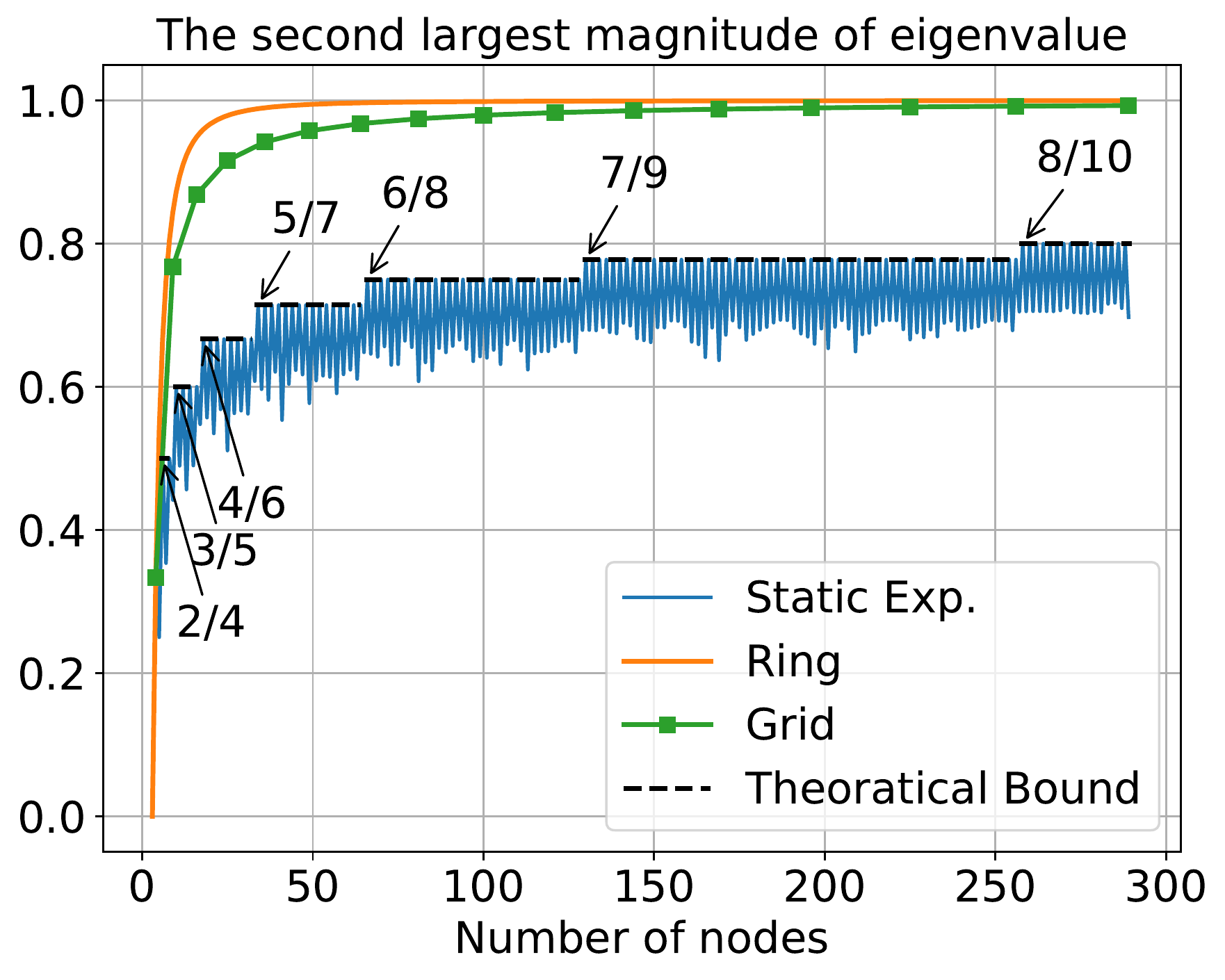}\hspace{-0.5mm}
\end{center}
\vspace{-5mm}
\caption{\small Spectral gap of some topologies.}
\label{fig.exp-eig-comparison-n0}
\vspace{-5mm}
\end{wrapfigure}

The theoretical analysis of Proposition \ref{prop-spectral-gap-expo} is non-trivial. To evaluate the spectral gap, for any network size $n$, we have to derive the analytical expression for each eigenvalue using Fourier transform and calculate the magnitudes. The most tricky part is to assert which eigenvalue expression attains the {\it second} largest value.

We now numerically validate the established spectral gap. In  Fig.~\ref{fig.exp-eig-comparison-n0}, we plotted the spectral gap of the static exponential graph with $n$ ranging from $4$ to $290$. It is observed that the derived gap $\rho = 1 - 2/(1 + \lceil \log_2(n) \rceil )$ is very  tight (see the black dashed line). In fact, it exactly matches the numerical spectral gap when $n$ is even. Moreover, it is also observed the spectral gap of static exponential graph is much smaller than that of ring or grid.

Finally, we compare the spectral gap and maximum degree of the static exponential graph with all other common graphs in Appendix \ref{apx.sub.static-compare}. 
It is observed that static exponential graph,  while with a sightly larger maximum degree, has a significantly smaller spectral gap than ring and grid.


\vskip -2mm
\section{One-Peer Exponential Graph Achieves Periodic Exact-Averaging}
\label{sec:dynamic-expo}
\vskip -2mm

Static exponential graph incurs  $\Omega(\log_2(n))$ communication overhead per iteration. 
To overcome this issue, \cite{assran2019stochastic} proposes to decompose the static exponential graph into a sequence of one-peer graphs, 
in which each node cycles through all its neighbors, communicating, only, to a single neighbor per iteration, see the right plot in Fig.~\ref{fig:static-one-peer-graph}. 
Apparently, each one-peer realization incurs $\Omega(1)$  communication cost, which matches with ring or grid. Since each realization is sparser than the static graph, one may expect DmSGD with one-peer exponential graphs are less effective in aggregating information. 
In the following, we will establish an interesting result: one-peer is very effective in averaging.


\textbf{Time-varying weight matrix.} We let $\tau = \lceil \log_2(n) \rceil$. The weight matrix at iteration $k$ is 
\begin{align}\label{wij-one-exp}
w^{(k)}_{ij} = 
\begin{cases}
\frac{1}{2} & \mbox{if $\log_2(\mathrm{mod}(j-i,n)) = \mathrm{mod}(k,\tau)$} \vspace{1mm}\\
\frac{1}{2} & \mbox{if $i=j$} \\
 0 & \mbox{otherwise.}\vspace{-2mm}
\end{cases}
\end{align}
The weight matrix for each realization of the one-peer exponential graphs in Fig.~\ref{fig:static-one-peer-graph} is in Appendix \ref{apx.sub.one-peer-weight}. Since each node communicates to one single neighbor per iteration, the resulting weight matrix is very sparse, with only one non-zero element in the non-diagonal positions per row and column. \vspace{-1mm}




\textbf{Periodic exact-averaging.} The periodic exact-averaging property, which was observed by \cite{assran2019stochastic} without theoretical justifications, is fundamental to clarify the averaging effectiveness of one-peer exponential graphs. The following lemma proves that the property holds when $n$ is a power of 2. \vspace{-1mm}

\begin{lemma}[\sc Periodic Exact Averaging]
\label{lm-perioidc-exact-averaging}
Suppose $\tau = \log_2(n)$ is a positive integer. If $W^{(k)}$ is the weight matrix generated by \eqref{wij-one-exp} over the one-peer exponential graphs, it then holds that each $W^{(k)}$ is doubly-stochastic, i.e. $W^{(k)} \mathds{1} = \mathds{1}$ and $\mathds{1}^T W^{(k)} = \mathds{1}^T$. Furthermore, it holds that\vspace{-1mm}
\begin{align}\label{finite-time-average-main-body}
W^{(k+\ell)} W^{(k+\ell-1)} \cdots W^{(k+1)} W^{(k)} = \frac{1}{n}\mathds{1}\mathds{1}^T
\end{align}
for any integer $k \ge0$ and $\ell \ge \tau$. And equivalently, the consensus residue form holds that\vspace{-1mm}
\eq{
    \big(W^{(k+\ell)} - \frac{1}{n}\mathds{1}\mathds{1}^T\big)
\big(W^{(k+\ell-1)} - \frac{1}{n}\mathds{1}\mathds{1}^T\big) \cdots  \big(W^{(k)} - \frac{1}{n}\mathds{1}\mathds{1}^T\big)=  0
}(Proof is in Appendix \ref{apx.sub.exact-avg-one-peer}). 
\end{lemma}

\begin{remark}\label{rm-power-2}
The assumption that $\log_2(n)$ is a positive integer seems  necessary. We numerically tested various one-peer exponential graphs with non-integer $\log_2(n)$. None of them is endowed with the periodic exact-average property. 
\end{remark}

\begin{remark}\label{rm-random-sampling}
When $\log_2(n)$ is a positive integer and each realization of the one-peer exponential graph is sampled \textbf{without replacement}, it is easy to verify that the periodic exact-averaging property still holds. However, if each realization is sampled with replacement, the periodic exact-averaging property generally does not hold unless all realizations are occasionally sampled without repeating.
\end{remark}

{\color{black}
\begin{remark}
It is worth noting that an one-peer variant of the hypercube graph is established to achieve exact averaging with $\tau = \log_n(n)$ steps \cite{shi2015finite}. Such one-peer hypercube is undirected and symmetric, which is different from the one-peer exponential graph which is directed and asymmetric. 
\end{remark}
}

\begin{wrapfigure}{R}{0.4\textwidth}
\begin{center}
    \includegraphics[width=0.4\textwidth]{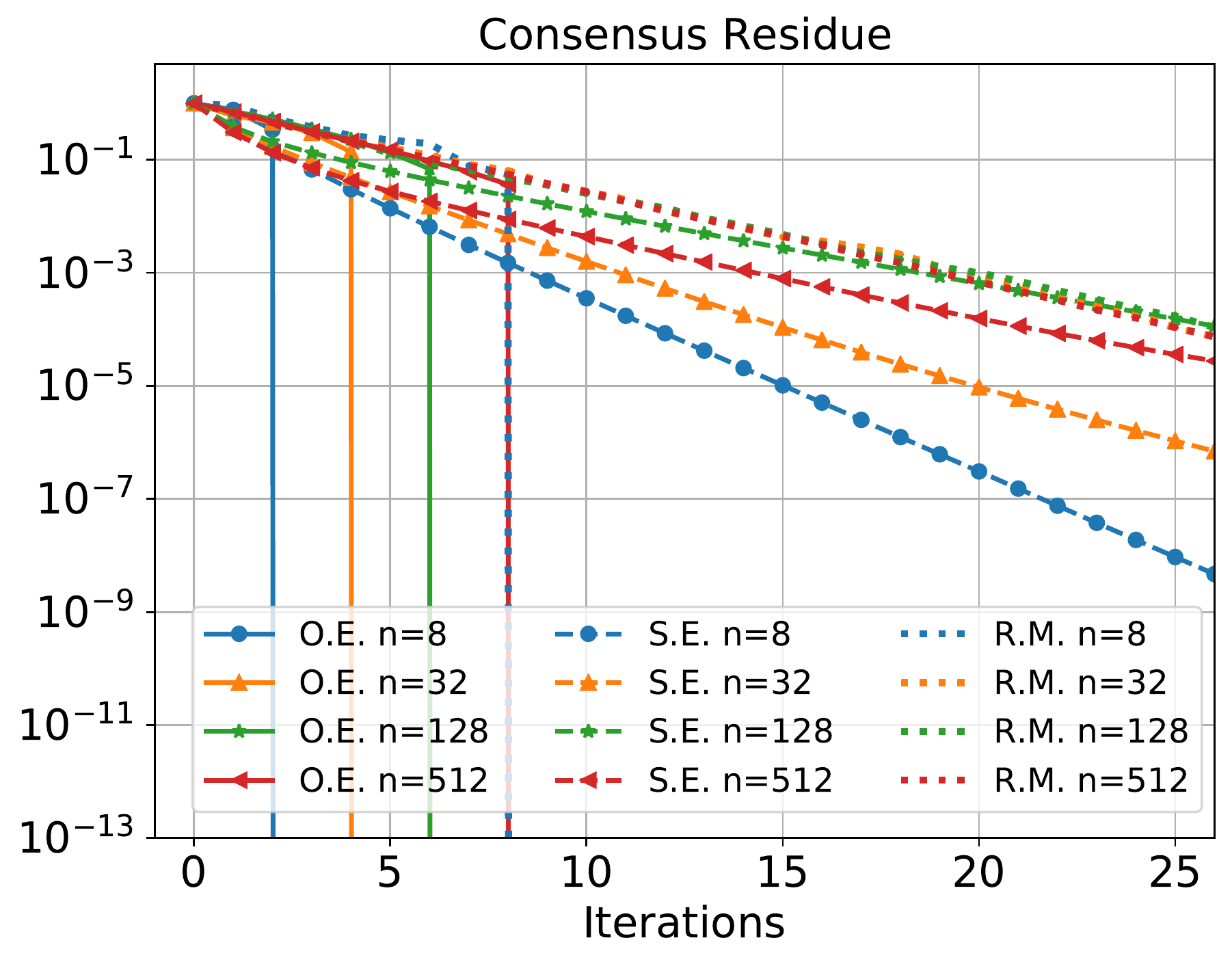}\vspace{-3mm}
\end{center}
\vspace{-2mm}
\caption{\small Illustration of how consensus residues decay with iterations for various graphs. O.E. and S.E.  denote one-peer and static exponential graphs, and R.M. denotes bipartite random match graph.}
\label{fig:period-averaging}
\vspace{-15mm}
\end{wrapfigure}


We now numerically validate Lemma \ref{lm-perioidc-exact-averaging}. To this end, we initialize a vector $x\in \RR^d$ arbitrarily, and examine how $\|(\Pi_{\ell=0}^k W^{(\ell)} - \frac{1}{n}\mathds{1}\mathds{1}^T)x\|$ decreases with iteration $k$. The weight matrix $W^{(k)}$ is either static or samples from one-peer exponential graph or bipartite random match graph. In Fig.~\ref{fig:period-averaging}, it is observed that one-peer exponential graphs can achieve exact average after $\log_2(n)$ steps, which coincides with the results in Lemma \ref{lm-perioidc-exact-averaging}. In contrast, the static exponential and bipartite random match graphs can only achieve the global average \textbf{asymptotically}. The justification for Remarks \ref{rm-power-2} and \ref{rm-random-sampling} is in Appendix \ref{apx.sub.one-peer-any-n}.

\section{DmSGD with Exponential Graphs}
\label{sec:DmSGD-analysis}  

With the derived property in Sec.~\ref{sec:static-expo} and \ref{sec:dynamic-expo}, this section will examine the convergence of DmSGD with static and one-peer exponential graphs. 

\textbf{DmSGD with static exponential graph.} Based on Proposition \ref{prop-spectral-gap-expo}, we can achieve the convergence rate and transient iterations, by following analysis in  \cite{yu2019linear}, of DmSGD with static exponential graph. 

\begin{corollary}\label{coro-tran-iter-se}
Under Assumptions A.1--A.4, if $\gamma = \frac{\sqrt{n(1-\beta)^{3}}}{\sqrt{T}}$, DmSGD (Algorithm \ref{Algorithm: DmSGD}) will converge at
\begin{align}\label{eq-thm-rate-one-peer}
\frac{1}{T}\sum_{k=1}^T  \Ex \|\nabla f(\bvx^{(k)})\|^2  =  O\left( \frac{\sigma^2}{\sqrt{(1-\beta) n T} } \hspace{-0.3mm}+\hspace{-0.3mm} \frac{n\log_2(n) (1 \hspace{-0.3mm}-\hspace{-0.3mm}\beta)\sigma^2}{T} \hspace{-0.3mm}+\hspace{-0.3mm} \frac{ n(1\hspace{-0.3mm}-\hspace{-0.3mm}\beta) b^2 \log^2_2(n)}{T}\right) 
\end{align}
Furthermore, the transient iteration complexity of DmSGD over static exponential graph is $O(n^3\log_2^2(n))$ for data-homogeneous scenario and $O(n^3\log_2^4(n))$ for data-heterogeneous scenario. 
\end{corollary}

\textbf{DmSGD with one-peer exponential graph.} With each realization being sparser than its static counterpart, one-peer exponential graph is believed to converge slower. However, the periodic exact-averaging property can help DmSGD achieve the same convergence rate as its static counterpart. Note that DmSGD with one-peer exponential graph is an  \textbf{one-loop} algorithm, see Algorithm \ref{Algorithm: DmSGD}. The DmSGD updates start immediately after sampling one weight matrix. 

\begin{theorem}
\label{thm-convergence-one-peer}
We assume $\tau = \log_2(n)$ is a positive integer, and the time-varying weight matrix is generated by \eqref{wij-one-exp} over one-peer exponential graphs. Under Assumptions A.1--A.4 and $\gamma = \frac{\sqrt{n(1-\beta)^{3}}}{\sqrt{T}}$, DmSGD (Algorithm \ref{Algorithm: DmSGD}) will converge at (Proof is in Appendix \ref{apx.sub.proof-dmsgd-convergence-1}-\ref{apx.sub.proof-dmsgd-convergence-3}). 
\begin{align}\label{eq-thm-rate-one-peer}
\frac{1}{T}\sum_{k=1}^T  \Ex \|\nabla f(\bvx^{(k)})\|^2  =  O\left( \frac{\sigma^2}{\sqrt{(1-\beta) n T} } + \frac{n(1-\beta)\sigma^2\tau}{T} + \frac{ n(1-\beta) b^2 \tau^2}{T}\right).
\end{align}
Furthermore, the transient iteration complexity of DmSGD over one-peer exponential graph is $O(n^3\log_2^2(n))$ for data-homogeneous scenario and $O(n^3\log_2^4(n))$ for data-heterogeneous scenario. 
\end{theorem}

\begin{remark}\label{rm-same-rate}
Comparing \eqref{eq-thm-rate-one-peer} with $(10)$, and noting that $\tau = \log_2(n)$, we conclude that \textbf{DmSGD with one-peer graphs converge exactly as fast as with the static counterpart in terms of the established rate bounds.} In addition, both graphs endow DmSGD with the same transient iteration complexity. 
\end{remark}
\begin{remark}
We can also achieve the convergence rate for decentralized SGD (i.e., DSGD without momentum acceleration) with one-peer exponential graph by setting $\beta = 0$. It is easy to verify that \textbf{DSGD with one-peer graphs can also converge as fast as with the static exponential graph}.
\end{remark}
\begin{remark}
{\color{black} The convergence rate and transient iteration complexity of DSGD with general mixing matrices sampling strategy are also studied in \cite{koloskova2020unified}. However, the results in reference \cite{koloskova2020unified} does not cover the scenario with momentum acceleration.} As we show in the proof details, it is highly non-trivial to handle momentum. 

\end{remark}
It is worth noting that the analysis for the above theorem is non-trivial. While it targets on the one-peer exponential graph, the analysis techniques can be extended to the general time-varying topologies. To our best knowledge, it establishes the first result for DSGD with momentum acceleration, over the time-varying topologies,  and in the non-convex settings. Existing analysis either focuses on DSGD without momentum \cite{koloskova2020unified}, or DmSGD with static topologies \cite{yu2019linear}. In addition, the last two terms in \eqref{eq-thm-rate-one-peer}, actually, can be further tightened by the spectral gap of one-peer exponential graphs. Since the tightened terms are rather complicated, we leave them to the discussion in Appendix \ref{apx.sub.converg-theorem}.

\textbf{State-of-the-art balance between communication and convergence.} Table \ref{tb-main-result} (and tables in Appendix \ref{apx.sub.converg-compare}) summarize the per-iteration communication time and transient iteration complexity for all commonly-used topologies. When $n$ is sufficiently large, the term $\log_2(n)$ can be ignored. In this scenario, the exponential graphs (including both static and one-peer variants) achieve state-of-the-art  $\tilde{\Omega}(1)$ per-iteration communication and $\tilde{\Omega}(n^3)$ transient iterations, in which $\tilde{\Omega}(\cdot)$ hides all logarithm factors. In Appendix \ref{apx.sub.converg-compare}, we numerically validate that exponential graphs have smaller transient iteration complexity than ring or grid graph as predicted in Table \ref{tb-main-result}. {\color{black}The comparison between exponential graph with random graphs \cite{nachmias2008critical,benjamini2014mixing,beveridge2016best,boyd2005mixing} (such as the Erdos-Renyi graph and geometric random graph) is discussed in Appendix~\ref{app-compare-random-graph}.}

\textbf{One-peer exponential graph is recommended for decentralized deep training}. It is because one-peer exponential graph endows DmSGD with the same convergence rate as its static counterpart, but incurs strictly less communication overhead per iteration. \vspace{-2mm}

\vskip -2mm
\section{Experiments}
\vskip -2mm
\label{sec:experiments}

This section will validate our theoretical results by extensive deep learning experiments. First, we  evaluate how DmSGD with exponential graphs perform against other commonly-used graphs with varying network size. Second, we examine whether one-peer exponential graphs achieve the same convergence rate and accuracy as its static counterpart across different tasks, models, and algorithms. 

\textbf{Metrics.} Training time and validation accuracy are two critical metrics to examine the effectiveness of a distributed training algorithm in deep learning. These two metrics are typically evaluated after the algorithm completes a fixed number of epochs (say, $90$ epochs). Training time can reflect the communication efficiency while  accuracy, though might not be precise, can roughly measure the convergence rate (or iteration complexity). These two metrics are used in most of our experiments.
\vspace{-3mm}



\vskip -2mm
\subsection{Setup}

\label{subsec-setup}
\vskip -2mm

We implement all decentralized algorithms with PyTorch \cite{paszke2019pytorch} 1.8.0 using NCCL 2.8.3 (CUDA 10.1) as the communication backend. For parallel SGD, we used PyTorch's native Distributed Data Parallel (DDP) module. For the implementation of decentralized methods, we utilize \textbf{BlueFog} \cite{bluefog}, which is a high-performance decentralized deep training framework, to facilitate the topology organization, weight matrix generation, and efficient partial averaging. We also follow DDP's design to enable computation and communication overlap. Each server contains 8 V100 GPUs in our cluster and is treated as one node. The inter-node network fabrics are 25 Gbps TCP as default, which is a common distributed training platform setting.


\vskip -2mm
\subsection{Exponential graphs enable efficient and high-quality training}\label{sec-experiment-implementaion}
\vskip -2mm

In this subsection we evaluate how DmSGD with exponential graphs perform against other commonly-used topologies in the task of image classification. 

\noindent \textbf{Implementation.} We conduct a series of image classification experiments with the ImageNet-1K \cite{deng2009imagenet}, which consists of 1,281,167 training images and 50,000 validation images in 1000 classes. We train classification models with different topologies and numbers of nodes to verify our theoretical findings. The training protocol in \cite{goyal2017accurate} is used. In details, we train total 90 epochs. The learning rate is warmed up in the first 5 epochs and is decayed by a factor of 10 at 30, 60 and 80-th epoch. The momentum SGD optimizer is used with linear learning rate scaling by default. Experiments are trained in the mixed precision using Pytorch native amp module. We implement DmSGD with all  graphs listed in Table \ref{tb-main-result}. The details of each graph is described in Appendix \ref{apx.experiment}.
For each graph, we test the training time and validation accuracy for DmSGD with GPU numbers ranging from $32$ to $256$. 

\noindent \textbf{Experiment results.} The comparison between different graphs (with varying size) in top-1 validation accuracy and training time after 90 epochs is listed in Table \ref{table-topo-nodes-comparison}. Major observations are:

\begin{table}[t!]
\vspace{-6mm}
\caption{\small Comparison of top-1 validation accuracy(\%) and training time (hours) with different topologies.}
\vspace{-5mm}
\label{table-topo-nodes-comparison}

\begin{center}
\begin{small}
\begin{sc}
\setlength{\tabcolsep}{0.8mm}{
\begin{tabular}{ccccccccc}
\toprule
     nodes & \multicolumn{2}{c}{4(4x8 GPUs)} & \multicolumn{2}{c}{8(8x8 GPUs)} & \multicolumn{2}{c}{16(16x8 GPUs)} & \multicolumn{2}{c}{32(32x8 GPUs)} \\
     topology & acc. & time & acc. & time & acc. & time & acc. & time    \\ 
\midrule 
Ring &  76.13 ±0.023 & 11.6 & 76.07 ±0.013 & 6.5 & 76.08 ±0.026 & 3.3 & 75.58 ±0.021 & 1.8 \\
Grid & 76.08 ±0.007 & 11.6 & 76.35 ±0.037 & 6.7 & 75.88 ±0.011 & 3.4 &  75.76 ±0.022 & 2.0 \\
Bi-Rand. Match. & 75.96 ±0.032 & {\bf 11.1} & 76.26 ±0.027 & {\bf 5.7} & 76.07 ±0.012 & {\bf 2.8} & 75.83 ±0.029 & {\bf 1.5} \\
Random Graph  & 75.97 ±0.028 & 11.5 & 76.01 ±0.033 & 7.1 & 76.18 ±0.008 & 6.7 & 76.24 ±0.018 & 4.7 \\
static exp. & 76.21 ±0.028 & 11.6 & 76.32 ±0.037 & 6.9 &  76.30 ±0.007 & 4.1 & 76.28 ±0.020  & 2.5 \\
one-peer exp. & {\bf 76.28 ±0.063} & {\bf 11.1} & {\bf 76.47 ±0.035} & {\bf 5.7} & {\bf76.42 ±0.030} & {\bf 2.8} & 
{\bf 76.30 ±0.062} & {\bf 1.5}

\\ \bottomrule
\end{tabular}}
\end{sc}
\end{small}
\end{center}
\vskip -0.25in
\end{table}




\begin{enumerate}[align=left,label=\textbf{[\arabic{*}]},leftmargin=0pt,labelwidth=!,labelsep=.5em]
    \vspace{-2mm}
    \item All graphs (except the random graph) endows DmSGD with training time linear speedup. Among them, bipartite random matching and one-peer exponential graphs achieve the best linear speedup due to their efficient per-iteration communication. However, the accuracy of the matching graph cannot match one-peer exponential graph.
    The random graph fails to achieve linear speedup because of its extremely expensive communication overheads.  \vspace{-1mm}
    
    \item In the $32\times 8$ GPUs scenario, the training time to finish all $90$ epochs can be sorted as follows: one-peer $\approx$ Bi-RandMatch $<$ Ring $<$ Grid $<$ static exponential $<$ random graph, which coincides with the per-iteration communication time listed in Table \ref{tb-main-result}. \vspace{-1mm}
    
    \item In the $32\times 8$ GPUs scenario, the training accuracy achieved by each graph after $90$ epochs is sorted as follows: random graph $\approx$ static exponential $\approx$ one-peer $>$ Bi-RandMatch $>$ Grid $>$ Ring, which coincides with the transient iteration complexity listed in Table  \ref{tb-main-result}. {\color{black}Note that the random graph is rather dense (see the detail in Appendix~\ref{app-details-topo}) so it has good accuracy but consumes significant wall-clock time in training.}
\end{enumerate}
With the second and third observations, we can find exponential graphs (especially the one-peer exponential graph) can enable both fast and high-quality training performance. We also examined the performance of exponential graphs when $n$ is not a power of $2$, see Appendix \ref{apx.sub.experiment-not-power2}.


\vskip -3mm
\subsection{One-peer exponential graph v.s. static exponential graph}
\label{sec-experiment-two-graphs-compare}
\vskip -2mm
In this subsection we will focus on the two exponential graphs studied in this paper. In particular, we will validate that one-peer exponential graph endows DmSGD with the same convergence rate as its static counterpart (i.e., the conclusion in Remark \ref{rm-same-rate}) across different tasks, models, and algorithms. 

\textbf{Comparison across models and algorithms.} Now we compare one-peer and static exponential graphs with different neural network architectures and algorithms. The task is image classification and the setting is the same as in Sec.~\ref{sec-experiment-implementaion}. We test both graphs for ResNet \cite{he2016deep}, MobileNetv2 \cite{sandler2018mobilenetv2} and EfficientNet \cite{tan2019efficientnet}, which are widely-used models in industry. In addition to the DmSGD algorithm (Algorithm \ref{Algorithm: DmSGD}) studied in this paper, we also examine how exponential graphs perform with other commonly-used decentralized momentum method: the vanilla DmSGD \cite{assran2019stochastic} which does not exchange momentum between neighbors, and QG-DmSGD \cite{lin2021quasi} which adds a quasi-global momentum to relieve the influence of data heterogeneity. We do not examine DecentLaM \cite{yuan2021decentlam} and D$^2$ \cite{tang2018d} because both methods require {\em symmetric} weight matrix during the training process which exponential graphs cannot provide. We also list the performance of parallel SGD using global averaging as one baseline. 

\begin{wrapfigure}{R}{0.65\textwidth}
\vspace{-0.2cm}
\begin{center}
    \includegraphics[width=0.65\textwidth]{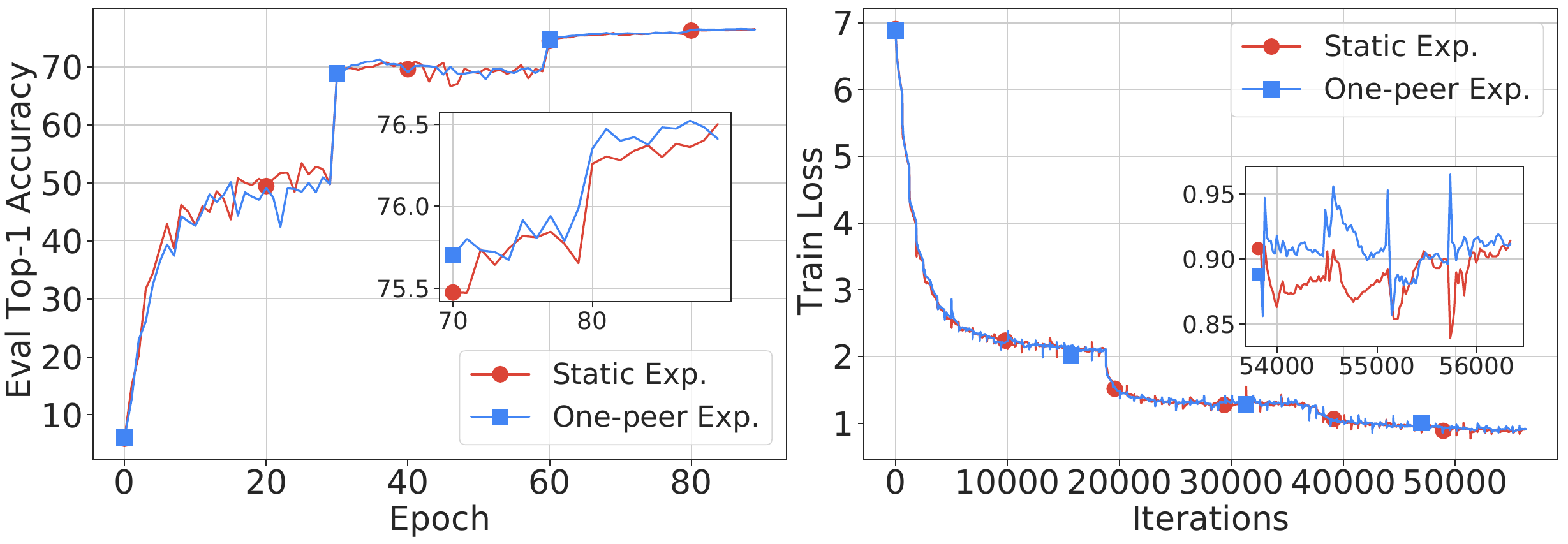}\vspace{-3mm}
\end{center}
\vspace{-2mm}
\caption{\small Convergence curves on the ImageNet (ResNet-50) in terms of training loss and validation top-1 accuracy . Network size is $8\times 8$ GPUs.}
\label{fig:Imagenet-speed}
\vspace{-3mm}
\end{wrapfigure}

Table \ref{tb:8*8} lists the top-1 validation accuracy comparison across all models and algorithms. In all scenarios, it is observed that both graphs can lead to roughly the same accuracy across models and algorithms. The accuracy difference ({\sc DIFF}) is marginal. We also depict the convergence curves in training loss and accuracy for DmSGD with both graphs in Fig.~\ref{fig:Imagenet-speed}. It shows both curves evolve closely to each other, indicating that one-peer exponential graph enables DmSGD with the same convergence rate as its static counterpart. This is consistent with Theorem \ref{thm-convergence-one-peer} and Remark \ref{rm-same-rate}. Since one-peer is more communication-efficient than static exponential graph (see Table \ref{table-topo-nodes-comparison}), it is recommended to utilize one-peer exponential graph in decentralized deep training. In addition, we observe that decentralized methods, while utilizing partial-averaging during training process, has no significant accuracy degradation compared parallel SGD. Decentralized SGD can even be superior sometimes. 


\begin{table}[t]

\begin{center}
\vskip -2mm
\caption{\small \color{black} Top-1 validation accuracy and wall-clock time(in hours) comparison with different models and algorithms on ImageNet dataset over static/one-peer exponential graphs (8x8 GPUs). }
\vskip -1mm

\label{tb:8*8}
\begin{small}
\begin{sc}
\setlength{\tabcolsep}{1.2mm}{
\begin{tabular}{ccccccc}
\toprule
    model &  \multicolumn{2}{c}{ResNet-50} & \multicolumn{2}{c}{MobileNet-v2} & \multicolumn{2}{c}{EfficientNet} \\
    Topology & static & one-peer & static & one-peer  & static & one-peer \\
    \midrule
    Parallel SGD &  76.21 (7.0) & - &    70.12 (5.8) & -  & 77.63 (9.0) & -  \\ 
    vanilla DmSGD & 76.14 (6.6) & 76.06 (5.5) & 69.98 (5.6) & 69.81 (4.6)  & 77.62 (8.4) & 77.48 (6.9)  \\
    DmSGD & 76.50 (6.9) & 76.52(5.7)  & 69.62 (5.7) & 69.98 (4.8) & 77.44 (8.7) & 77.51 (7.1)  \\
    QG-DmSGD & 76.43 (6.6) & 76.35(5.6)  & 69.83 (5.6) & 69.81 (4.6)  & 77.60 (8.4) & 77.72 (6.9)  \\
\bottomrule
\end{tabular}}
\end{sc}
\end{small}
\end{center}

\vskip -5mm
\end{table}


\textbf{Comparison across different tasks.} 
We next compare the aforementioned algorithms with one-peer and static exponential graphs in another well-known task: \textbf{object detection}. We will test the following widely-used models: Faster-RCNN \cite{ren2016faster} and RetinaNet \cite{lin2017focal} on popular PASCAL VOC \cite{everingham2010pascal} and COCO \cite{lin2014microsoft} datasets. We adopt the MMDetection  \cite{chen2019mmdetection} framework as the building blocks and utilize ResNet-50 with FPN \cite{lin2017feature} as the backbone network. We choose mean Average Precision (mAP) as the evaluation metric for both datesets. We used 8 GPUs (which are connected by the static or dynamic exponential topology) and set the total batch size as 64 in all detection experiments. 

Table \ref{table-detection-map} compares the performance of decentralized training across different object detection models and datasets. Similar to the above experiment, it is observed that both graphs enable decentralized algorithms with almost the same performance in each scenario. This again illustrates the value of one-peer exponential graph in deep learning tasks - it endows decentralized deep training with both fast training speed and satisfactory accuracy. 


\begin{table*}[h!]
\vskip -2mm
\caption{\small Comparison of different methods and models on PASCAL VOC and COCO datasets.}
\vskip 0.2mm
\begin{small}
\begin{sc}
\setlength{\tabcolsep}{1.25mm}{
\begin{tabular}{ccccccccc}
\toprule
    Dataset & \multicolumn{4}{c}{PASCAL VOC} & \multicolumn{4}{c}{COCO} \\
    Model & \multicolumn{2}{c}{RetinaNet}  & \multicolumn{2}{c}{Faster RCNN} & \multicolumn{2}{c}{RetinaNet} & \multicolumn{2}{c}{Faster RCNN} \\
    topology & static & one-peer & static & one-peer& static & one-peer& static & one-peer    \\ 
\midrule
Parallel SGD & 79.0 & - & 80.3 & - & 36.2 & - & 37.2 & -   \\ 
Vanilla DmSGD &  79.0 & 79.1 & 80.7 & 80.5 & 36.3 & 36.1 & 37.3 & 37.2  \\
DmSGD & 79.1 & 79.0 & 80.4 & 80.5 & 36.4 & 36.4 & 37.1 & 37.0 \\
QG-DmSGD & 79.2 & 79.1 & 80.8 & 80.4 & 36.3 & 36.2 & 37.2 & 37.1 \\
\bottomrule
\end{tabular}}
\end{sc}
\end{small}
\label{table-detection-map}
\vskip -5mm
\end{table*}

\section{Conclusion and Future Works}\vspace{-2mm}
In this paper, we establish the spectral gap of static exponential graph and prove  that any $\log_2(n)$ consecutive one-peer exponential graphs can together achieve exact averaging when $n$ is a power of 2. With these results, we reveal that  one-peer exponential graphs endow DmSGD with the same convergence rate as their static counterpart. We also establish that exponential graphs achieve nearly minimum per-iteration communication time and transient iteration complexity simultaneously when $n$ is large. All conclusions are thoroughly examined with  industrial-standard benchmarks. As the  future work, we will investigate symmetric time-varying graphs that can perform as well as one-peer exponential graph. Symmetric graphs are critical for D$^2$ and DecentLaM algorithms. 

{\color{black}
\section*{Acknowledgements}
The authors are grateful to Dr. Sai Praneeth Karimireddy from EPFL for the helpful discussions on the hypercube graph. 
}

{
\small
\bibliographystyle{ieee_fullname}
\bibliography{references}
}

\include{dyna_momentum}
\end{document}

%% file: macro.tex
\newcommand{\m}{{\boldsymbol{m}}}

\newcommand{\x}{{\boldsymbol{x}}}
\newcommand{\y}{{\boldsymbol{y}}}

\newcommand{\vg}{{\mathbf{g}}}

\newcommand{\vm}{{\mathbf{m}}}

\newcommand{\vx}{{\mathbf{x}}}
\newcommand{\vy}{{\mathbf{y}}}
\newcommand{\vz}{{\mathbf{z}}}

\newcommand{\cF}{{\mathcal{F}}}

\newcommand{\cN}{{\mathcal{N}}}

\newcommand{\cU}{{\mathcal{U}}}

\newcommand{\bx}{{\boldsymbol{x}}}

\newcommand{\cf}{{\mathcal{F}}}

\newcommand{\bvg}{{\bar{\mathbf{g}}}}
\newcommand{\bvm}{{\bar{\mathbf{m}}}}
\newcommand{\bvx}{{\bar{\mathbf{x}}}}

\newcommand{\bvz}{{\bar{\mathbf{z}}}}

\newcommand{\bxi}{{\boldsymbol{\xi}}}

\newcommand{\ww}{{\widehat{W}}}

\newcommand{\RR}{\mathbb{R}}

\newcommand{\vvvert}{{\vert\kern-0.25ex\vert\kern-0.25ex\vert}}
\newcommand{\eq}[1]{\begin{align}#1\end{align}}
\newcommand{\ExNorm}[1]{\mathbb{E}\,\left\|#1\right\|^2}
\newcommand{\nn}{\nonumber}

\newtheorem{lemma}{Lemma}
\newtheorem{proposition}{Proposition}
\newtheorem{theorem}{Theorem}
\newtheorem{corollary}{Corollary}
\newtheorem{remark}{Remark}

\newcommand{\qed}{\hfill\blacksquare}
\newcommand{\Ex}{\mathbb{E}\,}
\newcommand{\one}{\mathds{1}}

\usepackage{listings}
\usepackage{xcolor}

\definecolor{codegreen}{rgb}{0,0.6,0}
\definecolor{codegray}{rgb}{0.5,0.5,0.5}
\definecolor{codepurple}{rgb}{0.58,0,0.82}
\definecolor{backcolour}{rgb}{1.0, 1.0, 1.0}
\definecolor{dkgreen}{rgb}{0,0.6,0}
\definecolor{gray}{rgb}{0.5,0.5,0.5}
\definecolor{mauve}{rgb}{0.58,0,0.82}

%% file: dyna_momentum.tex
\appendix

\section{Static  Exponential Graph} \label{apx.static-exp}
\subsection{Weight matrix example} \label{apx.sub.exp-weight}

\begin{figure}[h!]
	\centering
	\includegraphics[width=0.8\textwidth]{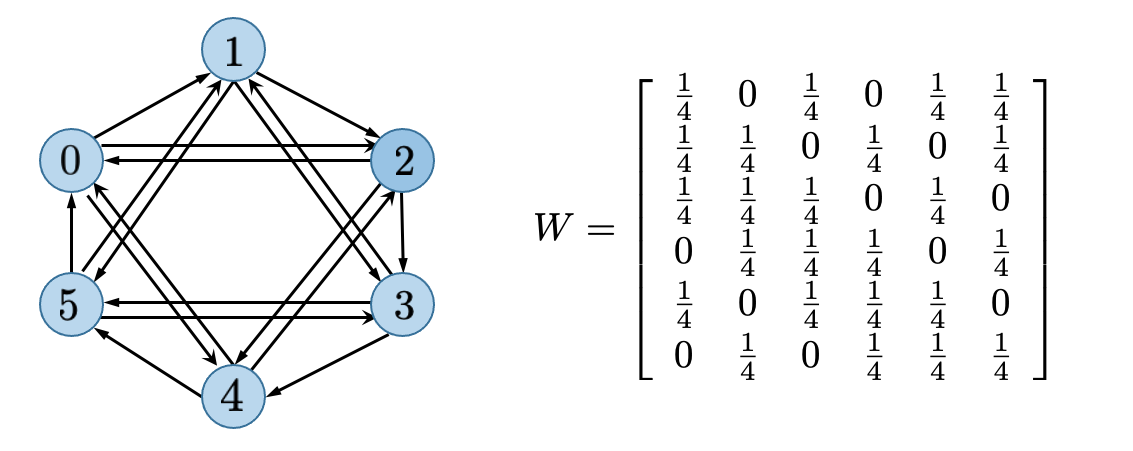} 
	\caption{\small Illustration of the $6$-node static exponential graph and its associated weight matrix.}
	\label{Fig:static-exp-weight-matrix}
\end{figure}

Fig.~\ref{Fig:static-exp-weight-matrix} illustrates the weight matrix $W$ defined in \eqref{wij-static-exp} for the $6$-node static exponential graph. The four nonzero entries in the first \emph{column} of $W$ correspond to the three outgoing neighbors of node $0$ and the node itself; the four nonzero entries on the first \emph{row} corrspond to the three incoming neighbors of node $0$ and the node itself.

\subsection{Spectral gap of static exponential graph} \label{apx.sub.exp-spectral-gap}
Before we present the proof of the spectral gap of static, we first need to review the Discrete Fourier Transform (DFT) and its connection to circulant matrix, which plays the critical role in the proof. We let ${\rm Circ}(c_0, c_1, \ldots, c_{n-1})$ denote a circulant matrix, which has the form:
\eq{
    C = {\rm Circ}(c_0, c_1, \ldots, c_{n-1}) \triangleq
    \begin{bmatrix}
        c_0     & c_{n-1}  &  c_{n-2}  & \ldots & c_{1} \\[1mm]
        c_{1} & c_0 &   c_{n-1}  &     & c_{2} \\
        \vdots & c_{1} &  c_0  & \ddots  & c_{n-2} \\
        c_{n-1}  &    &  \ddots  & \ddots  & c_{n-1} \\[1mm]
        c_{n} & c_{n-1} &  \ldots  & c_{1} & c_{0} \\
    \end{bmatrix}
}
and we call the circulant matrix $C$ is generated by the vector $c=(c_0, c_1,c_2,\ldots, c_{n-1} )$. With this notation, the circulant convolution can be equivalently re-written as the matrix-vector multiplication. Suppose we have two vectors $c\in \mathbb{C}^n$ and $v\in \mathbb{C}^n$:
\eq{
    c\otimes v = Cv
}
where $\otimes$ means the $n$-point circular convolution and $C$ is the circulant matrix generated by vector $c$.

\begin{lemma}[Eigenvalue of circulant matrix] \label{lemma:circ-eigv}
The eigenvalues of a circulant matrix ${\rm Circ}(c_0, c_1, \ldots, c_{n-1})$ are given by
\eq{
    \lambda_i = c_0 + c_1\omega^{i} + c_1\omega^{2i} + \cdots +  c_{n-1}\omega^{(n-1)i},\;\;\;i=0,1,\cdots, n-1 \label{circ-eigvalue}
}
where $\omega_{i}$ is the $i$-th root of unity under $n$-order, i.e.,
$\omega_{i} = \exp(2\pi j\frac{i}{n})$. (Note we use $j$ for imaginary number instead of $i$.)
\end{lemma}
{\bf Proof}. From the convolution theorem of Discrete Fourier Transform (DFT), we know that, for arbitrary $n$-dimension vector $c$ and $v$,  DFT of the $n-$circulant convolution of $c$ and $v$ equals to the element-wise multiplication of DFT of $c$ and DFT of $v$:
\eq{
   \mathfrak{F}(c \otimes v) = \mathfrak{F}(c)\odot  \mathfrak{F}(v)\;\;\; \forall c, v
}
where  $\odot$ means the Hadamard product.
Introduce the  DFT Matrix:
\eq{
    \mathcal{F} \triangleq 
    \begin{bmatrix}
        1      & 1          &   1  & \ldots & 1 \\[1mm]
        1      & \omega^{1} &  \omega^{2}  &     & \omega^{n-1} \\
        1      & \omega^{2} &  \omega^{4}  &     & \omega^{2(n-1)} \\
        \vdots & \vdots     &  \vdots  & \ddots  & \vdots\\
        1      & \omega^{n-1} &  \ldots  & & \omega^{(n-1)^2} \\
    \end{bmatrix}
}
It can be verified that
\eq{
     {\rm Diag}(\mathcal{F}c) \mathcal{F}v = (\mathcal{F}c) \odot (\mathcal{F}v)= \mathcal{F}(c \otimes v) = \mathcal{F} C v= \frac{1}{n} \mathcal{F} C  \mathcal{F}^\dagger \mathcal{F} v \label{12jgs}
}
where ${\rm Diag}(x)$ means the diagonal matrix built from vector $x$ and $\mathcal{F}^\dagger$ denotes the conjugate transpose of $\mathcal{F}$. In \eqref{12jgs}, we utilize the two identities that $x\odot y = {\rm Diag}(x) y$ and $x \otimes y = Xy$, where $X$ is the circulant matrix generated by $x$. Since \eqref{12jgs} holds for any vector $v$, we must have
\eq{
    \left(\frac{1}{\sqrt{n}}\mathcal{F}\right) C  \left(\frac{1}{\sqrt{n}}\mathcal{F}^\dagger\right) = {\rm Diag}(\mathcal{F}c)
}
Hence, any circulant matrix $C$ can be diagonalized by DFT matrix $\mathcal{F}$ and the corresponding eigenvalues are the DFT of the generating vector $c$. Expanding the expression for each element in $\mathcal{F}c$ will lead to \eqref{circ-eigvalue} immediately.
$\qed$.

With this powerful tool, we are ready to prove the spectral gap of exponential graph in Proposition \ref{prop-spectral-gap-expo}. To make the proof easier to follow, we split the proof into two parts. The first part is for special $n=2^\tau$ case. After that, we present the proof for arbitrary number $n$.


{\bf Proof of Proposition \ref{prop-spectral-gap-expo} (special $n=2^\tau$ case)}. First, we note by definition the combination matrix $W^{\rm exp}$ is a circulant matrix:
\eq{
       W^{\rm exp} = {\rm Circ}\left(\frac{1}{\tau+1}, \frac{1}{\tau+1},  \ldots, 0,  \frac{1}{\tau+1}, 0, \ldots\right)
}
Resorting to lemma \ref{lemma:circ-eigv}, we can immediate conclude that all eigenvalues of exponential graph have the following form:
\eq{
    \lambda_{i} = \frac{1}{\tau + 1} + \frac{1}{\tau + 1} \omega_i + \frac{1}{\tau + 1} \omega_i^2 + \frac{1}{\tau + 1} \omega_i^{2^2} + \ldots + \frac{1}{\tau + 1} \omega_i^{2^{\tau-1}}, \;\;\;i = 0, 1, ..., N-1
}
where $\omega_{i} = \exp{(2\pi j \frac{i}{N})}$ is the $i$-th root of unity under $N$-order. The magnitude of each eigenvalue is:
\eq{
    |\lambda_{i}| = \sqrt{\left( \frac{1}{\tau+1}+ \frac{1}{\tau+1}\sum_{n=0}^{\tau-1}\cos(\frac{2\pi i }{N}2^n)\right)^2 + \left( \frac{1}{\tau+1}\sum_{n=0}^{\tau-1}\sin(\frac{2\pi i}{N}2^n)\right)^2 }
}
However, it is not obvious which eigenvalue has the second largest magnitude. It is easy to see that
\eq{
    \lambda_{0} =  &\frac{1}{\tau+1} + \frac{1}{\tau+1}1 + \ldots + \frac{1}{\tau+1} 1 = 1
}
Recall that the eigenvalues of doubly stochastic matrix $W$ must be equal to or smaller  than $1$, we know $\lambda_{0}$ is the largest eigenvalue in magnitude. Next, it is also not hard to check that
\eq{
    \lambda_{n/2} =& \frac{1}{\tau+1} + \frac{1}{\tau+1} \omega_{n/2} + \ldots + \frac{1}{\tau+1} \omega_{n/2}^{2^{\tau-1}}\nn\\
    =&\frac{1}{\tau+1} + \frac{1}{\tau+1}(-1) + \frac{1}{\tau+1}(-1)^2 \ldots + \frac{1}{\tau+1} (-1)^{2^{\tau-1}}\nn\\
    =&\frac{\tau-1}{\tau+1}
}
where $n/2$ must be integer since $n=2^\tau$. As long as we can show that there is no other eigenvalue $\lambda_i$ lying between $\frac{\tau-1}{\tau+1}$ and $1$, we can claim that $\frac{2}{\tau+1}$ is the spectral gap of exponential graph.

Consider two cases for the rest $\lambda_i$:
\begin{enumerate}
    \item If $i$ is an odd number, we know that :
    \eq{
      \omega_i^{2^{\tau-1}} = \exp(2\pi j \frac{i 2^{\tau-1}}{2^{\tau}}) =  \exp(2\pi j \frac{i}{2}) = (-1)^i = -1
    }
    Then applying the triangle inequality, we know that 
    \eq{
        |\lambda_{i}| = \underbrace{\left|\frac{1}{\tau + 1} \omega_i + \frac{1}{\tau + 1} \omega_i^2 + \frac{1}{\tau + 1} \omega_i^{2^2} + \ldots + \frac{1}{\tau + 1} \omega_i^{2^{\tau-2}}\right|}_{{\tau-1}\; {\rm terms}} \leq \frac{\tau-1}{\tau+1}
    }
    Therefore, we conclude that the magnitude of any $|\lambda_i|$, where $i$ is an odd number, must not lie between $\frac{\tau-1}{\tau+1}$ and $1$. 
    \item If $i$ is an even number and $i$ is not zero. We can assume its prime factor decomposition has the following format:
    \eq{
      i = 2^{t'}p_1^{t_1}p_2^{t_2}\cdots p_\ell^{t_\ell}
    }
    where $p_\ell$ is some prime number except 2 and $t_\ell$ is the corresponding order. Because we know $i$ is strictly smaller than $2^\tau$ and $2$ is smallest prime number, we can claim that $ t' \leq {\tau-1}$. Since $i$ is some even number larger than 0, we also know $t' > 0$. These two conditions implies that among the index set $\{0, 1, 2,\cdots, \tau-1\}$, we can always find a number $\tau'$ such that $\tau' + t' = \tau - 1$. We evaluate :
    \eq{
        \omega_i^{2^{\tau'}} =& \exp\left(2\pi j \frac{i 2^{\tau'}}{2^{\tau}}\right) \nn\\
        = & \exp\left(2\pi j \frac{2^{\tau'} 2^{t'}p_1^{t_1}p_2^{t_2}\cdots p_\ell^{t_\ell}}{2^\tau}\right) \nn\\
        =&\exp\left(2\pi j \frac{p_1^{t_1}p_2^{t_2}\cdots p_\ell^{t_\ell}}{2}\right)\nn\\
        =& -1
    }
    Again, using the triangle inequality, we also can conclude that the magnitude of any $|\lambda_i|$, where $i$ is an even number, must not lie between $\frac{\tau-1}{\tau+1}$ and $1$. 
\end{enumerate}
So combining above two cases, we complete the proof that there is no other eigenvalue having the magnitude that is larger than $(\tau-1)/(\tau+1)$ and smaller than $1$.
$\qed$



{\bf Proof of Proposition \ref{prop-spectral-gap-expo} (the general cases)}.  The first several steps are the same as we did in previous proof. Next, we just need to show that there is no eigenvalue lying between $\frac{\tau-1}{\tau+1}$ and $1$.

Among in the index set $\{0, 1, 2,\cdots, \tau-1\}$, we select two numbers, denoting them as set $S$. Using the triangle inequality, we have
\eq{
    |\lambda_{i}| \leq& \frac{1}{\tau + 1} \left|\sum_{t=0, t\notin S}^{\tau-1} w_i^{2^t}\right| + \frac{1}{\tau + 1} \left|1 + \sum_{t\in S} w_i^{2^t}\right| \nn\\
    \leq & \frac{\tau-2}{\tau + 1}  + \frac{1}{\tau + 1} \left|1 + \sum_{t\in S} w_i^{2^t}\right|,\;\;\; \forall i
}
As long as we show that for all feasible $i$ but $0$, there always exist a set $S$ such that 
\eq{
    \left|1 + \sum_{t\in S} w_i^{2^t}\right|  \leq 1
}
we establish the upper bound for the second largest eigenvalue. The key to solve it is notice that, for $\alpha \in [0.25, 0.75]$, we have
\eq{
 |1+e^{2\pi j \alpha}+e^{2\pi j 2\alpha}| =|e^{2\pi j \alpha}(e^{-2\pi j \alpha} + 1 + e^{2\pi j \alpha})| = |1+\cos(2\pi\alpha)|\leq 1
}
\begin{figure}[h!]
    \centering
    \includegraphics[width=0.4\textwidth]{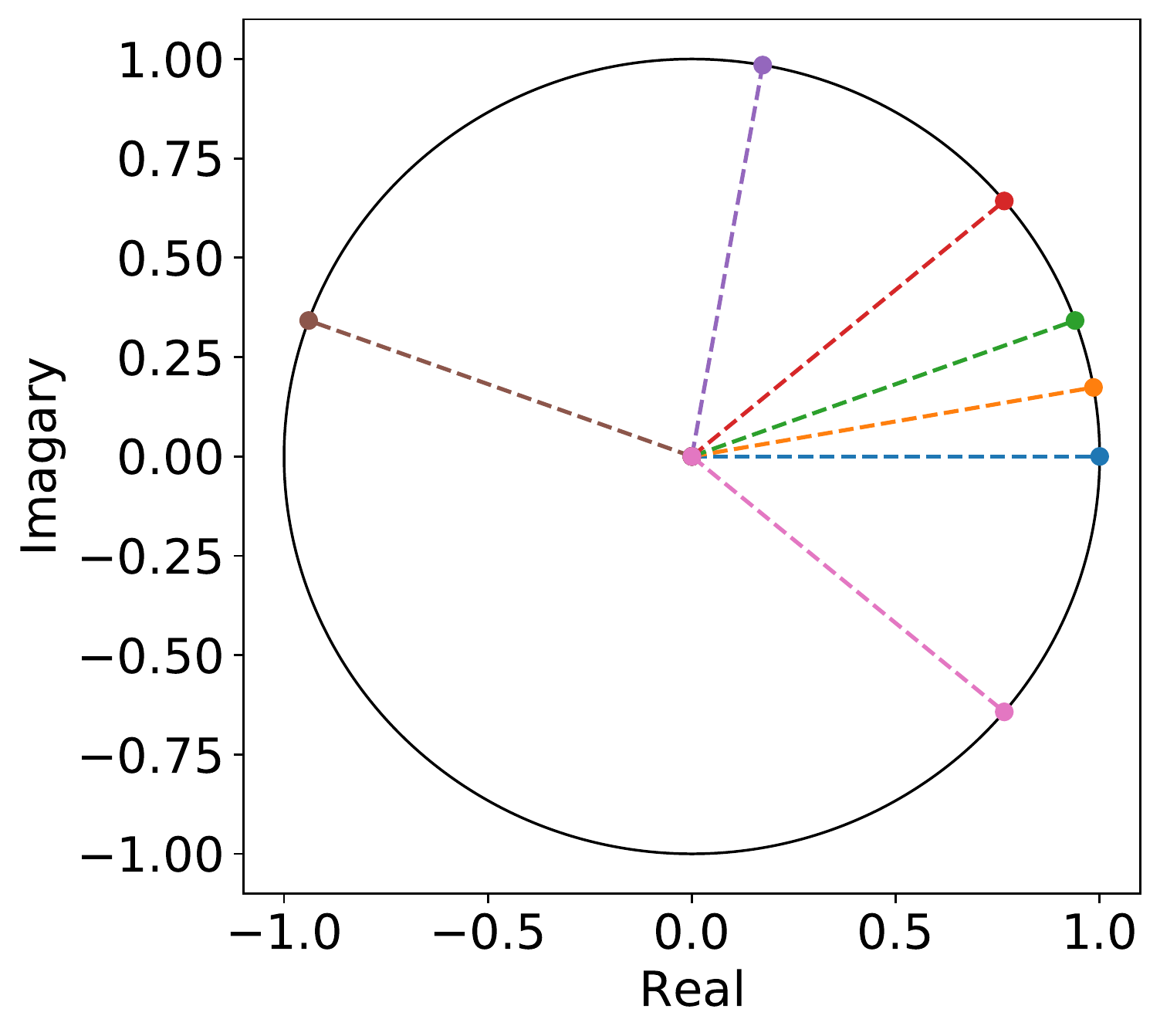}\hspace{2mm}
    \includegraphics[width=0.4\textwidth]{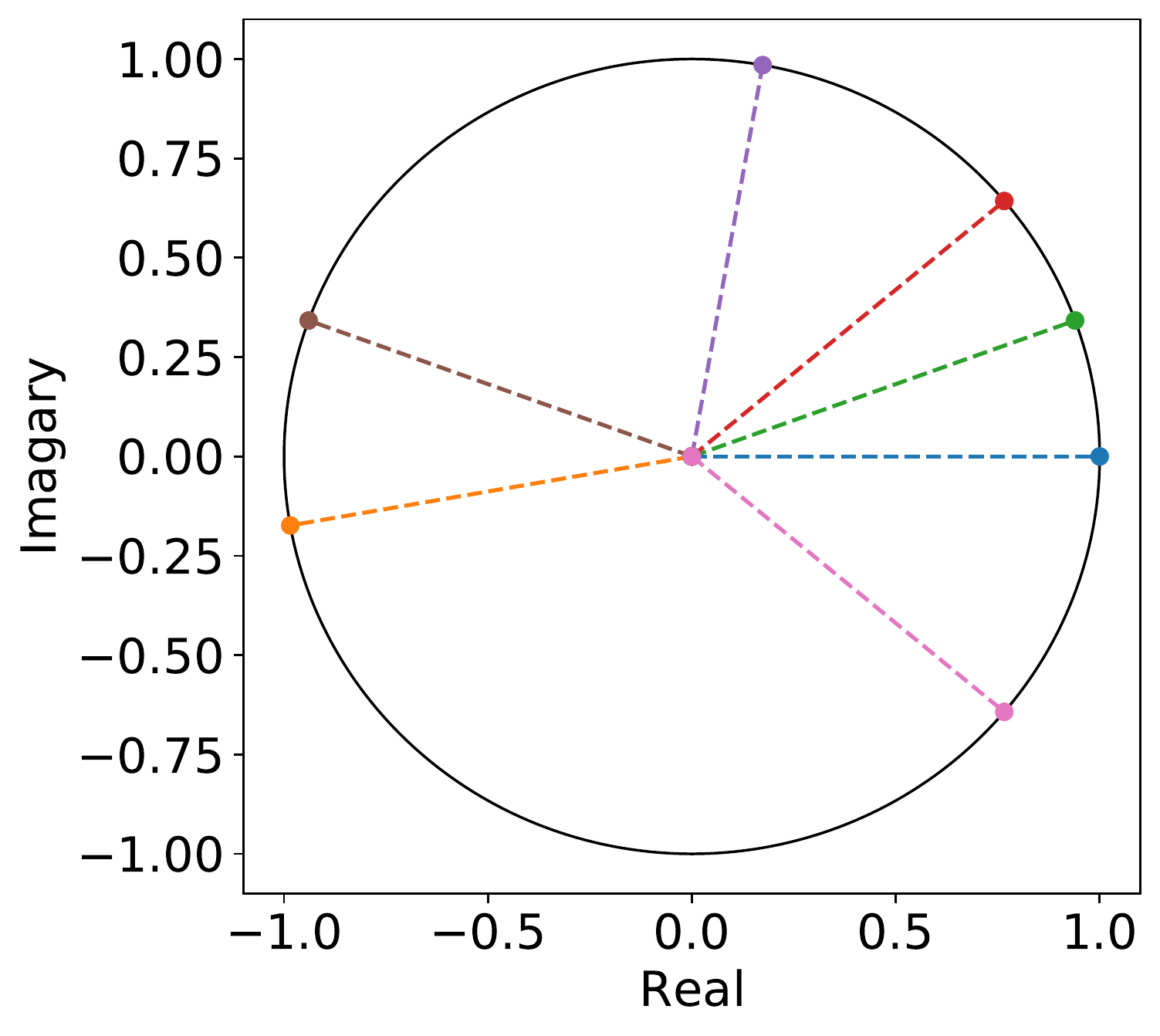}\vspace{-2mm}
    \caption{The position of each $\{\omega_i^{2^t}\}_{t=0}^{\tau-1}$ in the complex plane. The left figure shows the case that $i=1$ and $n=36$ and the right one shows the case that $i=19$ and $n=36$.}
\end{figure}

Next, we discuss case-by-case:
\begin{itemize}
    \item If $\frac{3n}{4}  \geq i \geq  \frac{n}{4}$, we can simply choose $S = \{0, 1\}$. We know
    \eq{
        \left|1 + w_i +  w_i^{2}\right| = \left|1 + \exp({2\pi j \frac{i}{n}}) +  \exp({2\pi j \frac{2i}{n}})\right| \leq 1 \label{18nv0s}
    }
     \item If $\frac{n}{4} > i > 1$. Because $2^\tau \geq n $, there exist a $t$ satisfying $\frac{3n}{4}  \geq  i 2^t \geq \frac{n}{4} $ and $t\leq \tau-2$. Choosing $S=\left\{t, t+1\right\}$ yields the desired inequality as we have in \eqref{18nv0s}: $ \left|1 + \sum_{t\in S} w_i^{2^t}\right|  \leq 1$
     \item If $n - 1 > i > \frac{3n}{4}$. Due the circular symmetry, we know that $w_i = w_{n-i}^*$, where we use $x^*$ as the conjugate of complex number $x$. It implies
     \eq{
        \left|1 + w_i +  w_i^{2}\right| =& \left|1 + w^*_{n-i} +  (w_{n-i}^{n-2})^*\right|\nn\\
        =&\left|1 + w_{n-i} + w_{n-i}^{n-2}\right| \label{9hfgw3}
    }
    Notice $n-i$ belongs to the range $(1, \frac{n}{4})$, we can immediate conclude that \eqref{9hfgw3} is smaller than 1.
     \item If $i=1$. We need to use a different argument to select the index set $S$ since the $t$ satisfying $\frac{3n}{4}  \geq  i 2^t \geq \frac{n}{4}$ may equal $\tau-1$. However, we still can select $S=\{\tau-2, \tau-1\}$:
     \eq{
      \left|1 + \exp^{2\pi j \frac{2^{\tau-2}}{n}} + \exp^{2\pi j \frac{2^{\tau-1}}{n}}\right|
      =& \left|\exp^{-2\pi j \frac{2^{\tau-2}}{n}} + 1 + \exp^{2\pi j \frac{2^{\tau-2}}{n}}\right| \nn\\
      =& \left| 1 + \cos\left(2\pi\frac{2^{\tau-2}}{n}\right)\right|
     }
     Since $2^\tau \geq n$, we know $\frac{2^{\tau-2}}{n} \geq \frac{1}{4}$. And we also know $\frac{2^{\tau-2}}{n} \leq \frac{1}{2}$, otherwise it indicates that $2^{\tau-1} \geq n$, which contradicts the assumption that $\tau =\lceil \log_2(n)\rceil$. Hence, we can conclude that $
         \cos\left(2\pi\frac{2^{\tau-2}}{n}\right) \leq 0
     $.
     
     \item If $i=n-1$. Use the conjugate argument then apply the similar procedure as $i=1$.
\end{itemize}
Above 5 cases cover all possible choices of $i$ for eigenvalues(except $0$). Hence, we conclude that for arbitrary $n$, the second largest magnitude of eigenvalue of exponential graph is bounded by $\frac{\tau-1}{\tau+1}$.

This bound is attained if $n$ is an even number, $\lambda_{n/2}$ is that desired eigenvalue. Based on the numerical experiment, we know it if $n$ is an odd number, this bound cannot be attained. Unfortunately, we  neither have a closed form solution for the spectral gap nor know which eigenvalue will becomes the second largest magnitude of eigenvalue. 

Lastly, we need to show that the $\ell_2$ matrix norm $\|W^{\rm exp} - \frac{1}{n} \one\one^T\|^2_2$ is the same value as spectral gap. To prove that, we resort to the Discrete Fourier Transform again
\eq{
    \|W^{\rm exp} - \frac{1}{n} \one\one^T\|^2_2 \overset{(a)}{=}& {\rm eig}_1 \Big((W^{\rm exp})^TW^{\rm exp} - \frac{1}{n} \one\one^T\Big) \nn\\
    \overset{(b)}{=}& {\rm eig}_1 \Big(\frac{1}{n^2}\cF D^\dagger \cF^\dagger \cF D \cF^\dagger - \frac{1}{n} \one\one^T\Big) \nn\\
    \overset{(c)}{=}& {\rm eig}_1 \Big(\frac{1}{n}\cF \big(D^\dagger D  - {\rm Diag}\{1, 0, 0,\cdots, 0\} \big)\cF^\dagger \Big) \nn\\
    =& {\rm eig}_1 \Big(\big(D^\dagger D  - {\rm Diag}\{1, 0, 0,\cdots, 0\} \big) \Big) \nn\\
    =& \rho(W^{\rm exp})^2
}
where ${\rm eig}_1(\cdot)$ means the largest eigenvalue of matrix, step (a) is because $\|X\|_2^2=\|X^TX\|_2 = {\rm eig}_1(X^TX)$, step (b) applied the eigenvalue decomposition of $W^{\rm exp}$ through the DFT, step (c) follows the fact that $\one$ is the first column of $\cF$.
$\qed$

\subsection{Comparing static exponential graph with commonly-used topologies} \label{apx.sub.static-compare}

\subsubsection{Details of each graph and the associated weight matrix}\label{app-details-topo}

\begin{figure}[h!]
	\centering
	\includegraphics[width=0.8\textwidth]{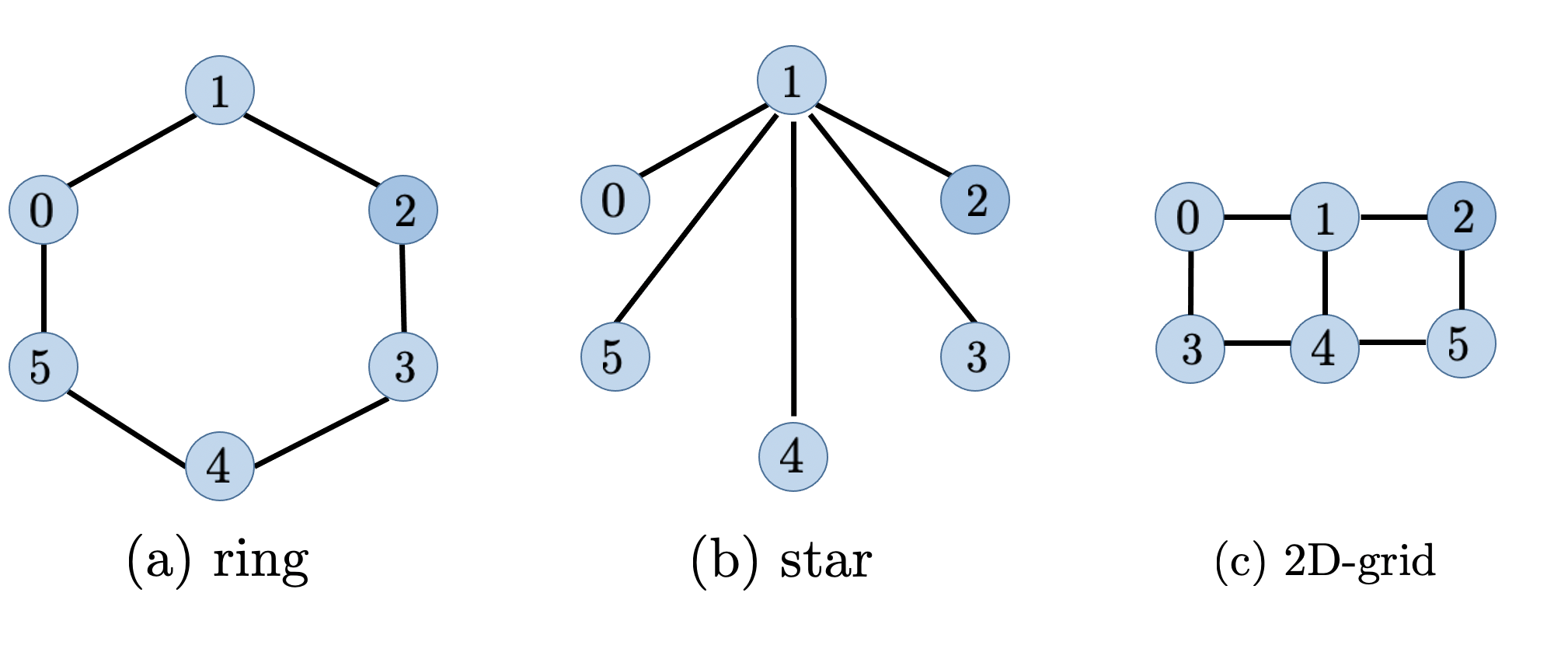} 
	\includegraphics[width=0.8\textwidth]{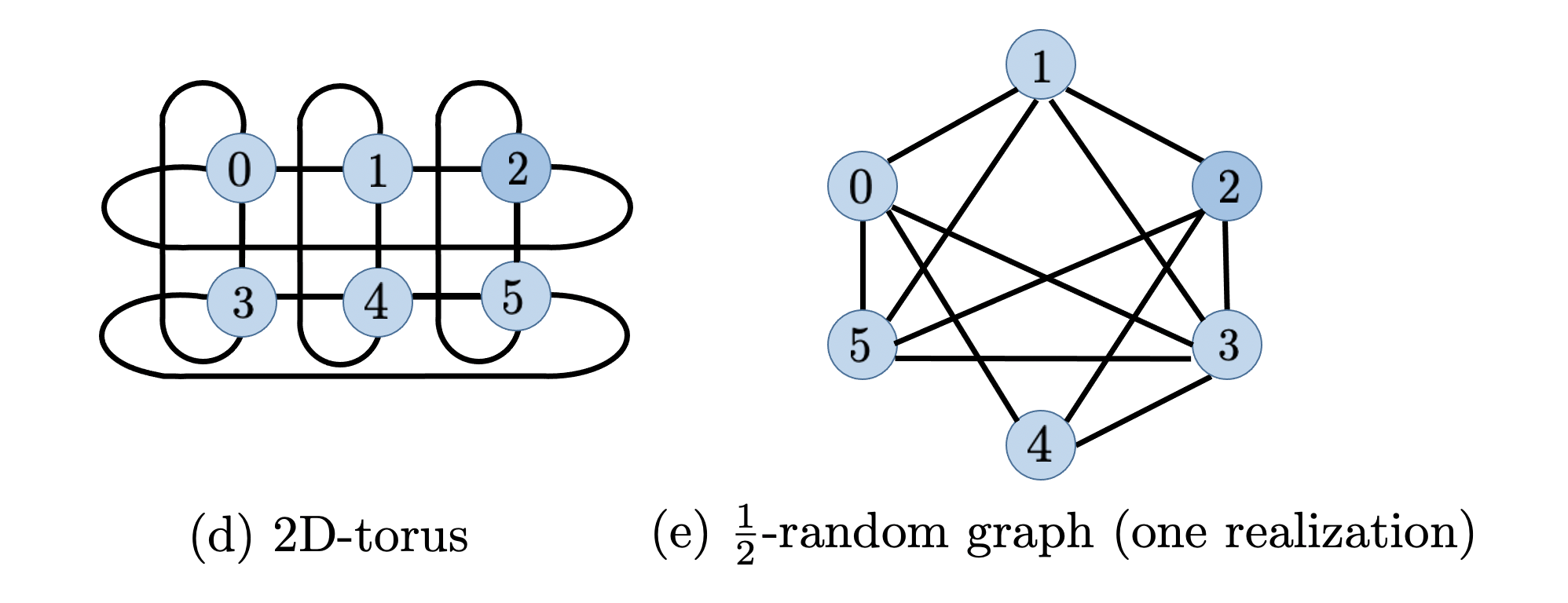}
	\includegraphics[width=0.85\textwidth]{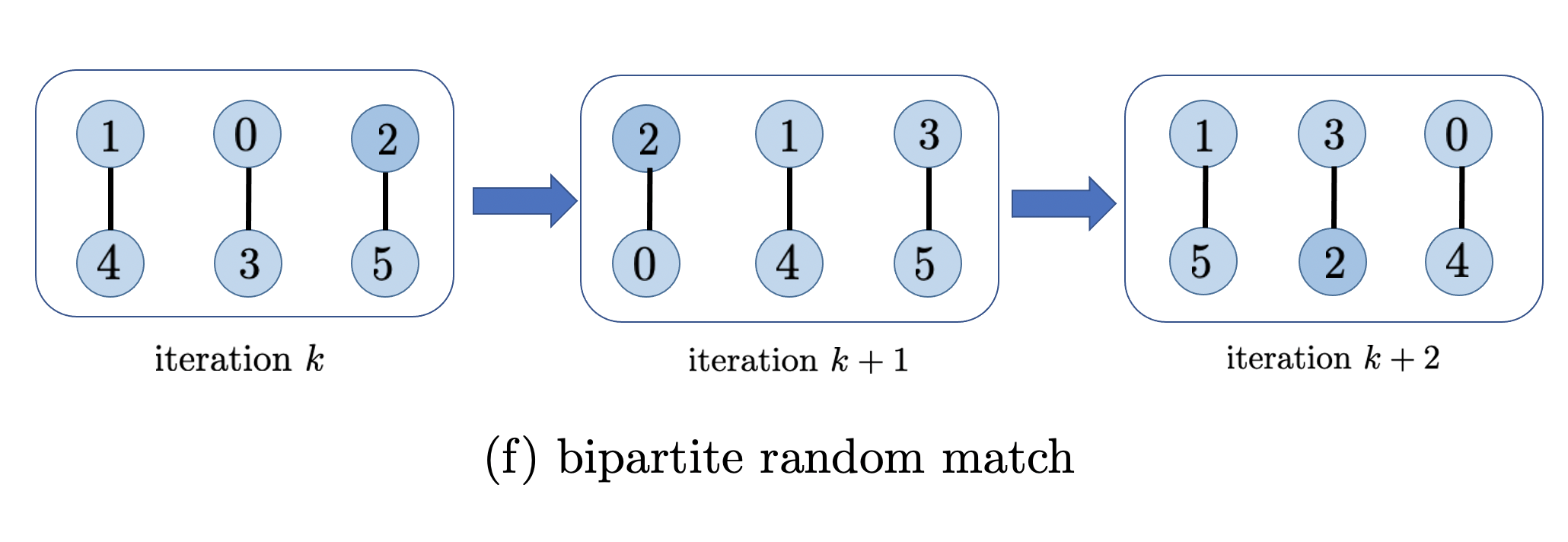}
	\caption{\small The shape of the $6$-node topologies discussed in Sec.~\ref{app-details-topo}.}
	\label{Fig:topologies}
\end{figure}

\begin{itemize}
    \item \textbf{Ring.} The ring topology is undirected, and is illustrated in Fig.~\ref{Fig:topologies}(a). Its weight matrix is generated according to the Metropolis rule \cite[Eq. (8)]{nedic2018network}, which is symmetric. 
    
    \item \textbf{Star.} The star topology is undirected, and is illustrated in Fig.~\ref{Fig:topologies}(b). Its weight matrix is generated according to the Metropolis rule, which is symmetric. Note that DmSGD with star graph still conducts partial averaging per iteration. It is different from parallel SGD that utilizes parameter-server (which is also of the star shape) to conduct global averaging.
    
    \item \textbf{2D-grid.} The 2D-grid topology is undirected, and is illustrated in Fig.~\ref{Fig:topologies}(c). Its weight matrix is generated according to the Metropolis rule, which is symmetric.
    
    \item \textbf{2D-torus.} The 2D-torus topology is undirected, and is illustrated in Fig.~\ref{Fig:topologies}(d). Its weight matrix is generated according to the Metropolis rule, which is symmetric.
    
    \item $\frac{1}{2}$-\textbf{random graph.} Consider a random $n$-node graph generated with each edge populating independently with probability $p=\frac{1}{2}$. Let $A \in \RR^{n\times n}$ be the symmetric adjacency matrix of the graph, with $A_{ij} = 1$ if nodes $i$ and $j$ are connected, and $0$ otherwise. We let $W = A/d_{\max}$ where $d_{\max} = \max_i \sum_{j} A_{ij}$. By union bound and
Bernstein’s inequality, one can easily derive $d_{\max}$ concentrates around $(n-1)/2$ with probability 1. It is derived in \cite[Proprosition 5]{nedic2018network} that the $\frac{1}{2}$-random graph has $1-\rho = O(1)$. A realization of one $\frac{1}{2}$-random graph is depicted in Fig.~\ref{Fig:topologies}(e). It is observed that the random graph is rather dense. 

    \item \textbf{Bipartite random match graph.} We assume $n$ is even. Bipartite random match graph is undirected and time-varying. To generate such graph at iteration $k$, we first randomly permute the index $(1, 2, \ldots, n)$ to $(\sigma^{(k)}(1), \sigma^{(k)}(2), \ldots, \sigma^{(k)}(n))$, where $\sigma^{(k)}(\cdot)$ denotes the permutation function at iteration $k$. Next we let node $\sigma^{(k)}(2j+1)$ be the neighbor of node $\sigma^{(k)}(2j)$ for each $j=0,\cdots, n/2-1$. It is obvious that, at iteration $k$, each node only exchanges information with one neighbor. The spectral gap of the bipartite random match graph, to our knowledge, is unknown yet in literature. Fig.~\ref{Fig:topologies}(f)  illustrates a sequence of the bipartite random match graphs.

\end{itemize}
\subsubsection{Comparison with commonly-used topologies}

Table \ref{table-app-topo} summarizes the maximum degree and $1-\rho$ for commonly-used graphs. The spectral gaps of ring, star, grid and torus are discussed in  \cite[Proprosition 5]{nedic2018network}.

\begin{table}[h!]
\caption{Comparison between commonly-used topologies in maximum degree and $1-\rho$.}
\centering
\begin{tabular}{rcc}
\toprule
\multicolumn{1}{c}{\textbf{}} & \textbf{$1-\rho$}         & \textbf{Max-degree} \\ \midrule
\textbf{ring}                 & $O(\frac{1}{n^2})$        & $2$            \vspace{1mm}     \\
\textbf{star}                    & $O(\frac{1}{n^2})$        & $n-1$            \vspace{1mm}      \\
\textbf{2D-grid}              & $O(\frac{1}{n\log_2(n)})$ & $4$              \vspace{1mm}      \\
\textbf{2D-torus}             & $O(\frac{1}{n})$          & $4$              \vspace{1mm}      \\
\textbf{$\frac{1}{2}$-random graph}     & $O(1)$          & $\frac{n-1}{2}$    \vspace{1mm}      \\
\textbf{random match}         & N.A.                      & $1$               \vspace{1mm}     \\
\textbf{static exponential}   & $O(\frac{1}{\log_2(n)})$  & $\log_2(n)$         \\ \bottomrule
\end{tabular}
\vspace{1mm}
\label{table-app-topo}
\end{table}

{\color{black}
\subsubsection{Comparison with random topologies}
\label{app-compare-random-graph}
A random graph is achieved by starting with a set of $n$ nodes and imposing successive edges between them  randomly. Random graph is extensively studied in wireless networks. To show the comparison between the exponential graph and various random graphs studied in \cite{nachmias2008critical,benjamini2014mixing,beveridge2016best,boyd2005mixing}, we first summarize the  differences between scenarios in deep learning and in wireless network and control theory:
\begin{itemize}
    \item \textbf{Topology size.} The GPUs utilized in deep learning are typically very expensive. A topology with tens or hundreds of GPUs is already regarded as a large network. This is different from wireless networks which may consist of thousands of (relatively cheap) sensors or mobile agents. The properties that are very likely to hold for large networks with thousands of nodes (e.g. the connectivity of the random graphs in \cite{nachmias2008critical,benjamini2014mixing,beveridge2016best,boyd2005mixing} with such a large size) may not valid for network with a small or moderate size.
    
    \item \textbf{Topology control.} Decentralized deep learning is typically conducted in data-center GPU clusters. In these clusters, GPUs are connected with high-bandwidth channels (such as InfiniBand, the optical fiber, etc.), and they can be organized in any topology shape. However, the network connectivity in the wireless network is highly sensitive to the geographical location of the nodes, and the radius of their wireless signals. The topology cannot be controlled freely in the latter setting.
    
    \item \textbf{Balanced degree.} Since the topology is in full control for deep learning, the topology design is very important for communication efficiency. In deep learning, we prefer topologies in which all nodes have identical degrees (i.e., the number of neighbors) so that they can finish the communication almost at the same time without waiting for the slowest one. Static and one-peer exponential graphs studied in our paper are such topologies. However, for the random graph in references \cite{nachmias2008critical,benjamini2014mixing,beveridge2016best,boyd2005mixing}, there always exists the possibility to generate a realization with highly unbalanced degrees, especially when the network size is not large.
\end{itemize}

References \cite{nachmias2008critical,benjamini2014mixing,beveridge2016best,boyd2005mixing} studied various random graphs. In this subsection, we will focus on the Erdos-Renyi graph $G(n,p)$ with $p = (1+c)\log(n)/n$ for some $c > 0$, and the 2-D geometric random graph $G(n, r)$ with $r^2 = (1+c)\log(n)/n$ for some $c > 0$. Both random graphs  are widely used in wireless networks. Table \ref{Table-comparison-random-graph} lists the comparison between exponential graph and the random graphs. In the table, it is observed that the E.-R. and geometric random graphs are either equivalent to, or worse than, exponential graphs in either per-iteration communication, or the transient iteration complexity. Moreover, note that the per-iteration communication cost for random graphs is calculated in expectation. In practice, the maximum degree in both random graphs must be greater than the expected degree for each node in the table, which will lead to an even slower per-iteration communication cost than exponential graphs. With the results listed in the above table as well as the other comparison described below, we still recommend using exponential graphs in deep learning.

\begin{table}[]
\caption{Comparison between exponential graph and the random graphs.} 
\label{Table-comparison-random-graph}
\begin{tabular}{rcccc}
\toprule 
\textbf{}       & \textbf{E.-R. Random} & \textbf{Geometric Random} & \textbf{Static Exp.}  & \textbf{O.P. Exp}   \vspace{2mm}  \\ \midrule
Per-iter. comm. & $\tilde{\Omega}(1)$ in expectation   & $\tilde{\Omega}(1)$ in expectation      & $\tilde{\Omega}(1)$   & ${\Omega}(1)$       \vspace{2mm}    \\
Transient iter. & $\tilde{\Omega}(n^3)$ & $\tilde{\Omega}(n^5)$     & $\tilde{\Omega}(n^3)$ & $\tilde{\Omega}(n^3)$  \vspace{2mm}  \\
Connectivity    &     \begin{tabular}[c]{@{}c@{}}Connected when $n$\\ is sufficiently large\end{tabular}                &        \begin{tabular}[c]{@{}c@{}} Connected when $n$\\ is sufficiently large\end{tabular}                      &   \begin{tabular}[c]{@{}c@{}} Always\\Connected\end{tabular}                 &         \begin{tabular}[c]{@{}c@{}}Disconnected \\ for some  iter.$^\dagger$\end{tabular}          \vspace{2mm}      \\
Degree balance  &      \begin{tabular}[c]{@{}c@{}}Can be highly \\ unbalanced \end{tabular}                  &          \begin{tabular}[c]{@{}c@{}}Can be highly \\ unbalanced \end{tabular}                     &             Balanced          &             Balanced              \\ \bottomrule 
\multicolumn{5}{l}{$^\dagger$\footnotesize{While disconnected for some iteration, it is proved to work for DmSGD.}}\\
\end{tabular}
\end{table}
}

\section{One-peer Exponential Graph} \label{apx.one-peer-exp}
\subsection{Weight matrix example} \label{apx.sub.one-peer-weight}
\begin{figure}[h!]
	\centering
	\includegraphics[width=0.75\textwidth]{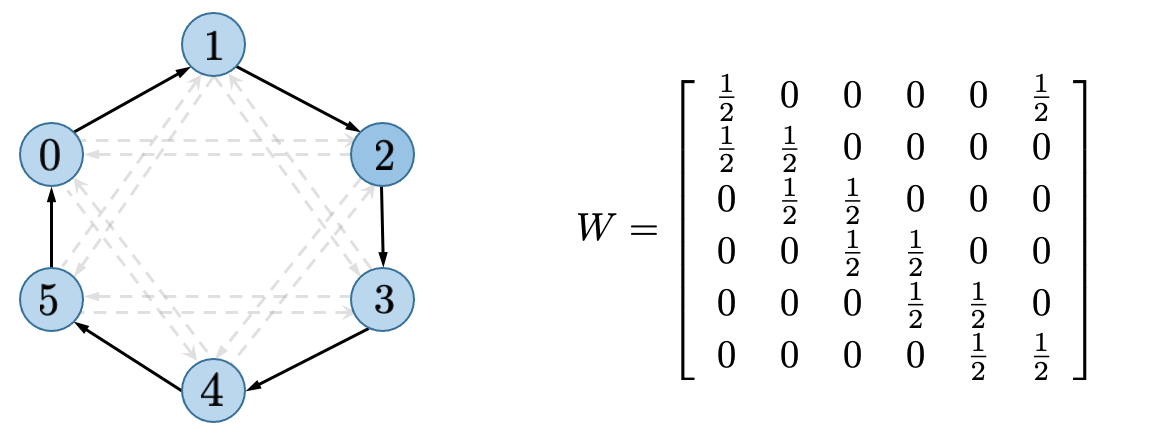} 
	\includegraphics[width=0.75\textwidth]{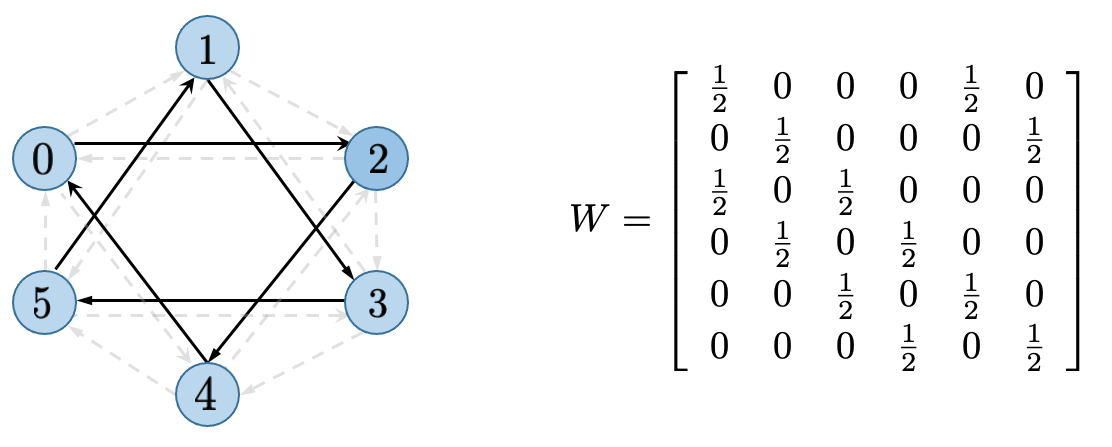}
	\includegraphics[width=0.75\textwidth]{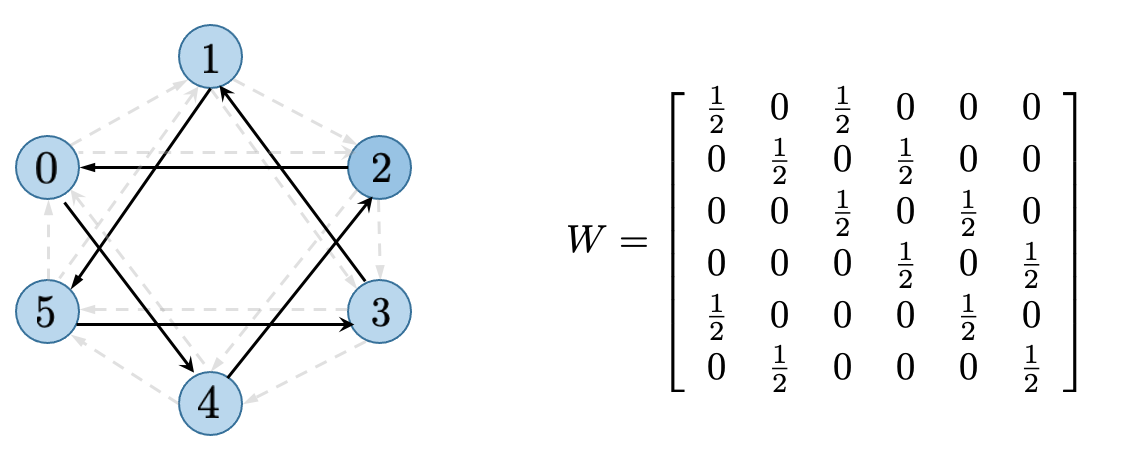}
	\caption{\small Illustration of the $6$-node one-peer exponential graph and its associated weight matrix.}
	\label{Fig:one-peer-exp-weight-matrix}
\end{figure}

Fig.~\ref{Fig:one-peer-exp-weight-matrix} illustrates the weight matrix $W$ defined in \eqref{wij-one-exp} for the $6$-node one-peer exponential graph. 

\subsection{Periodic exact averaging of one-peer exponential graph} \label{apx.sub.exact-avg-one-peer}\vspace{-2mm}
We present and prove a lemma that is more general than Lemma \ref{lm-perioidc-exact-averaging} in the main body. Therefore, its proof also serves the proof of Lemma \ref{lm-perioidc-exact-averaging}.
\begin{lemma}[\sc Exact averaging] \label{exact_average_lemma}
Suppose $W^{(k)}$, $k\ge 0$, are the weight matrices defined in \eqref{wij-one-exp} over the one-peer exponential graph. It holds that each $W^{(k)}$ is doubly-stochastic, i.e., $W^{(k)} \mathds{1} = \mathds{1}$ and $\mathds{1}^T W^{(k)} = \mathds{1}^T$. Furthermore, if there exists an integer $\tau \ge 0$ such that $n = 2^\tau$, then it holds that
\begin{align}\label{finite-time-average}
W^{(k_{\ell})} \cdots W^{(k_2)} W^{(k_1)}= \frac{1}{n}\mathds{1}\mathds{1}^T
\end{align}
as long as $\{\mathrm{mod}(k_1,\tau),\dots,\mathrm{mod}(k_\ell,\tau)\}=\{0,\dots,\tau-1\}$. In particular, the weight matrices associated with one-peer exponential graph can help reach an exact consensus average after all $\tau$ different matrices are each applied at least once.
\end{lemma}

\textbf{Proof.}
The double stochasticity of every $W^{(k)}$ follows directly from their definitions. It is left to establish \eqref{finite-time-average}.

Since $W^{(k_1)}=W^{(k_2)}$ as long as $\mathrm{mod}(k_1,\tau) = \mathrm{mod}(k_2,\tau)$, we can assume all $k_i \in\{0,\dots,\tau-1\}$ without loss of generality.

Since the eigenvectors of all circulant matrices of the same size are the same set of Fourier modes, circulant matrices $W^{(k)}$ for all $k\ge 0$ are simultaneously diagonalizable and their multiplications are commutative, i.e., $W^{(k_1)}W^{(k_2)}=W^{(k_2)}W^{(k_1)}$ for any $k_1,k_2\ge 0$. 
This property, together with the fact $\{k_1,\dots,k_\ell\}=0,\dots,\tau-1$ and the double stochasticity of every $W^{(k)}$, implies it suffices to show
$
    W^{(0)}W^{(1)}\dots W^{(\tau-1)} = \frac{1}{n}\mathds{1}\mathds{1}^T
$
or, for the convenience of argument below,
\begin{align}\label{finite-time-average1}
    (W^{(\tau-1)})^T\dots(W^{(1)})^T (W^{(0)})^T = \frac{1}{n}\mathds{1}\mathds{1}^T.
\end{align}

Consider $y=(W^{(\tau-1)})^T\dots(W^{(1)})^T (W^{(0)})^Tx$. 
(We index the entries of $x,y$ starting from 0, instead of 1, for we use a binary representation below.)
Since all nodes are treated equally, it suffices to show $y_0 = \frac{1}{n}(x_0+\dots+x_{n-1})$ since, through shifting the node indices, this equality implies $y_i = \frac{1}{n}(x_i+\dots+x_{n-1} + x_0 +\dots +x_{i-1})=\frac{1}{n}(x_0+\dots+x_{n-1})$ for $i=1,2,\dots,n-1$.

In a graph with $n=2^\tau$ nodes, \emph{index} the nodes by decimal numbers $0,\ldots,n-1$. Obtain their binary-form numbers:
\begin{align*}
\text{decimal index} &~~|~~ \text{binary index}\\
    0 & = 0 \ldots 0 0\textup{b}\\
    1 & = 0 \ldots 0 1\textup{b}\\
    2 & = 0 \ldots 1 0\textup{b}\\
    & \vdots \\
    n-2 & = 1 \ldots 1 0\textup{b}\\
    n-1 & = 1 \ldots 1 1\textup{b}.
\end{align*}
There are $\tau$ bits in each binary number above, denoted by $b$; the $j$th bit, $j=0,\dots,\tau-1$, from right to left is denoted by $b_j$. For the example of node 2, $b_{0}=0$, $b_{1}=1$, and then $b_{j}=0$ for $j=2,\dots,\tau-1$.

Pick any single $(W^{(k)})^T$ from $k=0,\dots,\tau-1$. The results of applying $x' = (W^{(k)})^T x$ are
\begin{align*}
    y_{b} = \frac{1}{2}(x_{\mathrm{mod}(b,n)} + x_{\mathrm{mod}(b+2^k,n)}),\quad \forall b=0,\dots,n-1.
\end{align*}
In particular, $x' = \frac{1}{2}\sum_{b\in B} x_b$ for $B=\{b:b_j=0~\forall j\neq k\}=\{0,2^k\}$, that is, all bits of $b$ are 0 except for the $k$th bit, which is either 0 or 1.

Now pick $k_1\neq k_2\in \{0,\dots,\tau-1\}$, then $x'' = (W^{(k_2)})^T(W^{(k_1)})^T x$ satisfies
\begin{align*}
    x''_{b} & = \frac{1}{2}(x'_{\mathrm{mod}(b,n)} + x'_{\mathrm{mod}(b+2^{k_2},n)})\quad\text{where}~x'=(W^{(k_1)})^T x\\
    & = \frac{1}{4}(x_{\mathrm{mod}(b,n)} + x_{\mathrm{mod}(b+2^{k_1},n)}+ x_{\mathrm{mod}(b+2^{k_2},n)}+ x_{\mathrm{mod}(b+2^{k_1}+2^{k_2},n)}),\quad \forall b=0,\dots,n-1.
\end{align*}
In particular, $y_0 = \frac{1}{2}\sum_{b\in B} x_b$ for $B=\{b:b_j=0~\forall j\neq k_1,k_2\}$, that is, all bits of $b$ are 0 except for the $k_1$th and $k_2$th bits, which are either 0 or 1.

Using proof by induction, it is easy to show that $z = (W^{(k_\ell)})^T\dots(W^{(k_1)})^T x$ for \emph{distinct} $k_\ell,\dots,k_1\in \{0,\dots,\tau-1\}$ satisfies
\begin{align*}
    y_{0} = \frac{1}{2^{\ell}}\sum_{b\in B}x_b,\quad B= \{b: b_{j}=0,\forall j\not\in \{k_1,\dots,k_\ell\}\}.
\end{align*}

By taking $k_\ell=\tau-1,\dots,k_2=1,k_1=0$ (where $\ell=\tau-1$), we have proved \eqref{finite-time-average1} and the lemma. {\color{white}quadquadquad}  $\qed$

\begin{corollary}
Under the same condition as stated in Lemma \ref{exact_average_lemma}, it also holds that
\begin{align}\label{w-prod-zero}
    \left(W^{(k_{\ell})} - \frac{1}{n}\mathds{1}\mathds{1}^T\right)\cdots \left( W^{(k_2)}- \frac{1}{n}\mathds{1}\mathds{1}^T\right)  \left( W^{(k_1)} - \frac{1}{n}\mathds{1}\mathds{1}^T\right) = 0
\end{align}
as long as $\{\mathrm{mod}(k_1,\tau),\dots,\mathrm{mod}(k_\ell,\tau)\}=\{0,\dots,\tau-1\}$. 
\end{corollary}
{\bf Proof}. Consider the production of two terms:
\eq{
    \left(W^{(k_\ell)} - \frac{1}{n}\mathds{1}\mathds{1}^T\right) \left(W^{(k_{\ell-1})} - \frac{1}{n}\mathds{1}\mathds{1}^T\right) =W^{(k_{\ell})}W^{(k_{\ell-1})}- \frac{1}{n}\mathds{1}\mathds{1}^T
}
Here we utilize the doubly stochastic property of $W^{(k_\ell)}$ that $W^{(k_\ell)}\mathds{1}\mathds{1}^T = \mathds{1}\mathds{1}^T$.  Repeating above process until all terms are merged, we 
\eq{
    &\hspace{-8mm}\left(W^{(k_\ell)}- \frac{1}{n}\mathds{1}\mathds{1}^T\right) \left(W^{(k_{\ell-1})} - \frac{1}{n}\mathds{1}\mathds{1}^T\right) \cdots  \left(W^{(k_1)} - \frac{1}{n}\mathds{1}\mathds{1}^T\right)\nn\\
    &= W^{(k_{\ell})} W^{(k_{\ell-1})} \cdots W^{(k_2)} W^{(k_1)} - \frac{1}{n}\mathds{1}\mathds{1}^T 
}
Last, referring Lemma \ref{exact_average_lemma}, we conclude the l.h.s product in \eqref{w-prod-zero} is an all-zero matrix.
$\qed$

\subsection{More about one-peer exponential graph} \label{apx.sub.one-peer-any-n}
\subsubsection{One-peer exponential graph with the size that is not power 2}\label{app-exp-graph-not-power-2}

\begin{figure}[h!]
    \centering
    \includegraphics[width=0.6\textwidth]{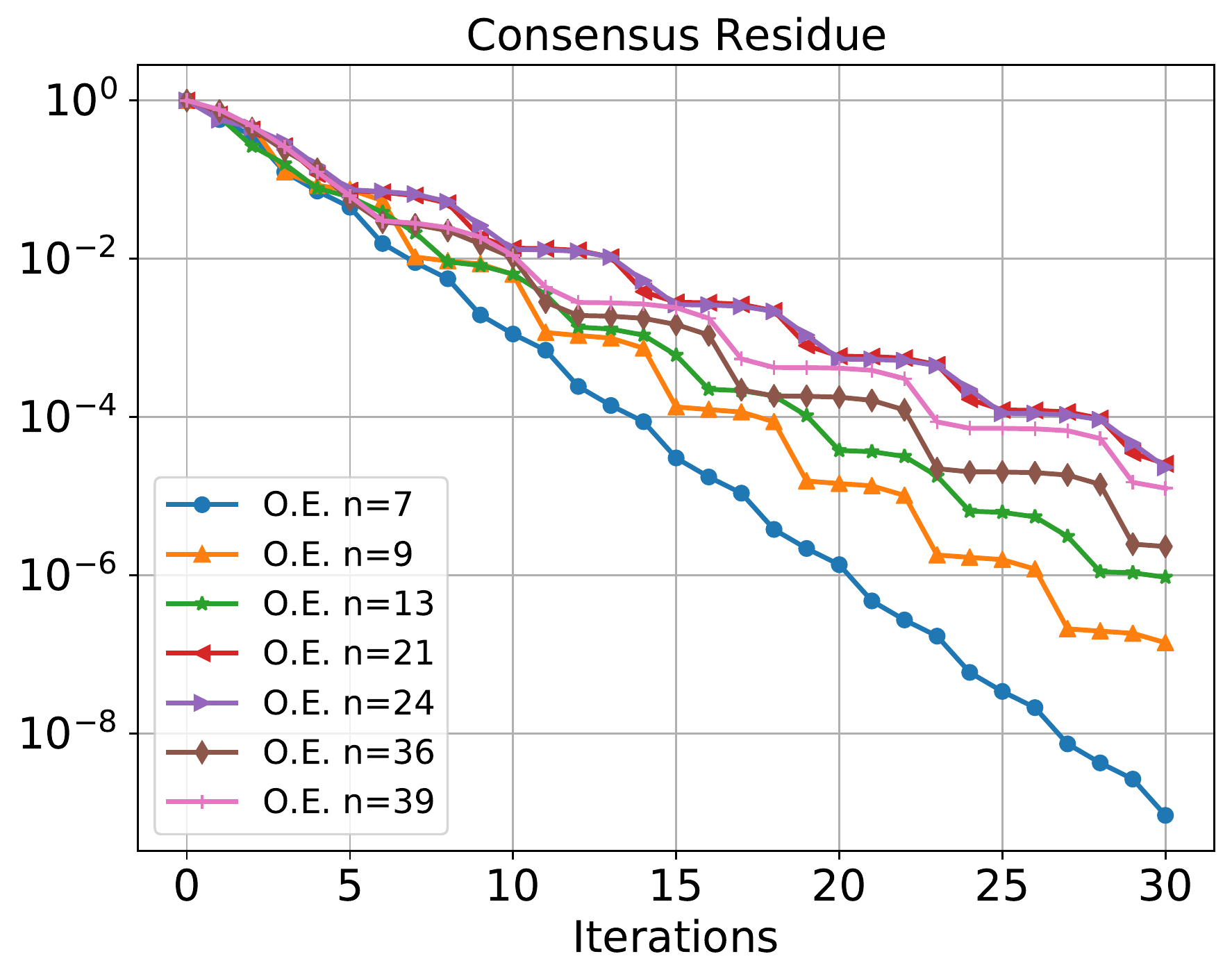}\vspace{-3mm}
    \caption{\small Illustration of how consensus residues decay with iterations for one-peer exponential graph with the size of nodes is not the power of 2.
    }
    \label{fig:dynamic_eig_not_power2}
\end{figure}

\textbf{Numerical validation.} 
First, we numerically examine whether 
one-peer exponential graph can achieve periodic exact-averaging when 
the number of nodes is not the power of $2$. To this end, we consider the same setting as in Fig.~\ref{fig:period-averaging}, and depict how the consensus residue $\|(\Pi_{\ell=0}^k W^{(\ell)} - \frac{1}{n}\mathds{1}\mathds{1}^T)x\|$ decreases as iteration increases in Fig.~\ref{fig:dynamic_eig_not_power2}. It is observed that, when $n$ is not a power of $2$, one-peer exponential graphs can only achieve the asymptotic, not periodic, exact averaging. 

{\bf A case study: One-peer exponential graph with $3$ nodes}. We provide an example to show that it is impossible to achieve the periodic exact averaging that when the size of nodes is 3. In this case, the period is $\lceil \log_2(3) \rceil = 2$. Due to the symmetry between the nodes, the product of two one-peer exponential graph weight matrices has the form
\eq{W^{(1)}W^{(0)} = &
    \begin{bmatrix}
        1-\beta& \beta & \\
        & 1-\beta &\beta \\å
        \beta  & & 1-\beta
    \end{bmatrix}
    \begin{bmatrix}
        1-\alpha & & \alpha \\
        \alpha & 1-\alpha &\\
         & \alpha & 1-\alpha
    \end{bmatrix} \nn\\
    =&\begin{bmatrix}
        1-\alpha - \beta+2\alpha\beta & \beta - \alpha\beta & \alpha - \alpha\beta  \\
        \alpha - \alpha\beta & 1-\alpha - \beta+2\alpha\beta & \beta - \alpha\beta  \\
        \beta - \alpha\beta & \alpha - \alpha\beta & 1-\alpha - \beta+2\alpha\beta 
    \end{bmatrix}
}
In order to achieve the exact averaging, the product has to be $\frac{1}{3} \one_3\one_3^T$. Under this requirement, it is easy to derive that 
\eq{
 \alpha = \beta,\;\; \alpha^2 - \alpha + \frac{1}{3} = 0 \;\;  \Longrightarrow \;\;\alpha = \beta = \frac{1}{6}(3\pm j\sqrt{3})
}
However, it is meaningless to let the combination weights to be complex number since the domain of iterate $x_i^{(k)}$ is $\mathbb{R}^{d}$. 

\subsubsection{One-peer exponential graph with uniform sampling and random permutation}\label{app-exp-graph-uniform-sampling}

In the main body, we only consider the one-peer exponential graphs in the cyclic order. However, that is not the only choice of selecting one-peer exponential graphs. Other two popular strategies are random permutation and uniform sampling. It is easy to describe these two strategies by taking an example. 
Consider
\eq{
  \mathbb{W} \triangleq \{W^{(0)}, W^{(1)}, \cdots, W^{(\tau-1)}\}.
}
Uniform sample strategy is at each iteration, one $W^{(t)}$ randomly selected with replacement. While random permutation is at each iteration, one $W^{(t)}$ randomly selected {\it without} replacement. After $\tau$ iterations, $\mathbb{W}$ will reset with $\tau$ element and repeat the sampling without replacement. 

\begin{figure}[t]
    \centering
    \includegraphics[width=0.6\textwidth]{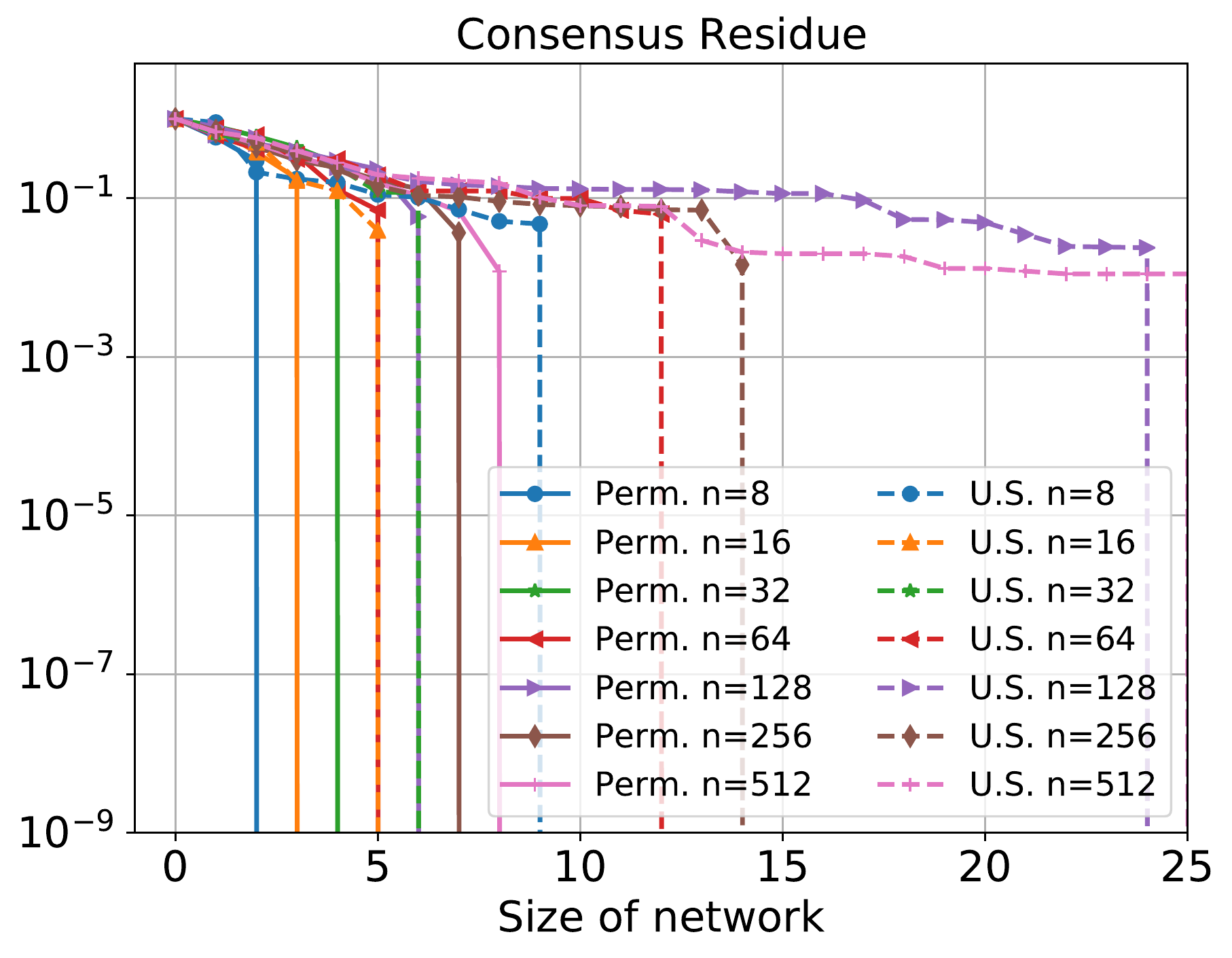}\vspace{-1mm}
    \caption{\small Illustration of how consensus residues decay with iterations for one-peer exponential graph. Perm. stands for random-permutation one-peer exponential graphs and U.S. stands for uniformed sampling one-peer exponential graphs.
    }\label{fig:dynamic_eig_perm}
\end{figure}

With slightly modification of the proof in lemma \ref{lm-perioidc-exact-averaging}, we also show that one-peer exponential graph with random permutation still has the exact averaging property. Meanwhile, one-peer exponential graph with uniform sample may no longer has this property within $\tau$-iterations. With some none-zero probability, the realization of uniform sampling with $\tau$ times can be one permutation order. Obviously, in this realization, uniform sample will have exact convergence property. Under the rest realizations, it cannot since it may miss at least one element. However, with long enough time $t$, the realization of uniform sampling with $t$  times will contain all possible elements in $\mathbb{W}$ with probability one. These claims are validated in the Fig.~\ref{fig:dynamic_eig_perm}.\vspace{-3mm}

\section{Deriving transient iteration complexity for DmSGD} \label{apx.transient}
We copy the convergence rate of DmSGD for non-convex costs in \eqref{dmsgd-convergence} as follows
\begin{align}\label{app-dmsgd-convergence}
\frac{1}{T}\sum_{k=1}^T  \Ex \|\nabla f(\bvx^{(k)})\|^2  =  O\left( \frac{\sigma^2}{\sqrt{nT} } + \frac{n\sigma^2}{T(1-\rho)} + \frac{ n b^2}{T(1-\rho)^2}\right)  \quad
\end{align}
in which $\bar{x}^{(k)}=\frac{1}{n}\sum_{i=1}^n x^{(k)}_i$, and the influence of momentum $\beta$ is ignored. In the following, we will derive the transient iteration complexity of DmSGD for the data-homogeneous and data-heterogeneous scenarios, respectively. 

\begin{itemize}
    \item In the \textbf{data-homogeneous} scenario, it holds that $D_i = D_j$ for any $i$ and $j$, and hence $\nabla f_i(x) = \nabla f_j(x)$. This implies that $b^2 = \frac{1}{n}\sum_{i=1}^n\|\nabla f_i(x) - \nabla f(x)\|^2 = 0$. To reach the linear speedup stage, the iteration $T$ has to be sufficiently large so that the $nT$-term dominates, i.e.,
    \begin{align*}
    \frac{\sigma^2}{\sqrt{nT}} \ge \frac{n\sigma^2}{T(1-\rho)}, \quad \mbox{which is equivalent to} \quad T \ge \frac{n^3}{(1-\rho)^2}.
    \end{align*}
    As a result, the transient iteration complexity of DmSGD is given by $\Omega({n^3}/{(1-\rho)^2})$.
    
    \item In the \textbf{data-heterogeneous} scenario, it holds that $b^2 \neq 0$. To reach the linear speedup stage, the iteration $T$ has to be sufficiently large so that
    \begin{align*}
    \frac{\sigma^2}{\sqrt{nT}} \ge \frac{n b^2}{T(1-\rho)^2}, \quad \mbox{which is equivalent to} \quad T \ge  \frac{n^3 b^4}{(1-\rho)^4 \sigma^4}.
    \end{align*}
    As a result, the transient iteration complexity of DmSGD is given by $\Omega({n^3}/{(1-\rho)^4})$ if the influences of $b^4$ and $\sigma^4$ are ignored. 
\end{itemize}
With the above arguments, we achieve the transient iteration complexity in \eqref{eq-tran-iter-hetero}.

\section{Proof of Theorem \ref{thm-convergence-one-peer}} \label{apx.proof-dmsgd-convergence}

\subsection{Notations and preliminaries} \label{apx.sub.proof-dmsgd-convergence-1}

\textbf{Notations.} We first introduce necessary notations as follows.
\begin{itemize}
	\item $\vx^{(k)} = [(\x_1^{(k)})^T; (\x_2^{(k)})^T; \cdots; (\x_n^{(k)})^T]\in \mathbb{R}^{n\times d}$
	\item ${\color{black}\vm^{(k)} = [(\m_1^{(k)})^T; (\m_2^{(k)})^T; \cdots; (\m_n^{(k)})^T]\in \mathbb{R}^{n\times d}}$
	\item $\nabla F(\vx^{(k)};\bxi^{(k)}) = [\nabla F_1(\x_1^{(k)};\bxi_1^{(k)})^T; \cdots; \nabla F_n(\x_n^{(k)};\bxi_n^{(k)})^T]\in \mathbb{R}^{n\times d}$
	\item $\nabla f(\vx^{(k)}) = [\nabla f_1(\x_1^{(k)})^T; \nabla f_2(\x_2^{(k)})^T; \cdots; \nabla f_n(\x_n^{(k)})^T]\in \mathbb{R}^{n\times d}$ 
	\item {$\bvx^{(k)} = \left(\frac{1}{n}\sum_{i=1}^n \x_i^{(k)}\right)^T \in \mathbb{R}^{d}$}
	\item {$\bvm^{(k)} = \left(\frac{1}{n}\sum_{i=1}^n \m_i^{(k)}\right)^T \in \mathbb{R}^{d}$}
	\item $W=[w_{ij}]\in \mathbb{R}^{n\times n}$.
	\item $\mathds{1}_n = \mathrm{col}\{1,1,\cdots, 1\} \in \RR^n$
	\item {\color{black} Given two matrices $\vx, \vy \in \RR^{n\times d}$, we define inner product $\langle \vx, \vy \rangle = \mathrm{tr}(\vx^T \vy)$, the Frobenius norm $\|\vx\|^2 = \langle \vx, \vx \rangle$, and the $\|\vx\|_2$ as $\vx$'s matrix $\ell_2$ norm. }
\end{itemize}
From the above definitions, it is quick to check that $\bar{\vx}^{(k)}=\frac{1}{n}\mathds{1}^T \vx^{(k)}$ and $\bar{\vm}^{(k)}=\frac{1}{n}\mathds{1}^T \vm^{(k)}$. We adopt the convention \footnote{If you are familiar with the NumPy library, it is the array broadcasting concept [\url{https://numpy.org/doc/stable/user/basics.broadcasting.html}].} that 
\eq{
 \bar{\vx}^{(k)} - \vx^{(k)} \triangleq \left[(\x_1^{(k)} - \frac{1}{n}\sum_{i=1}^n \x_i^{(k)} )^T; (\x_2^{(k)} - \frac{1}{n}\sum_{i=1}^n \x_i^{(k)})^T; \cdots; (\x_n^{(k)} - \frac{1}{n}\sum_{i=1}^n \x_i^{(k)})^T\right]\in \mathbb{R}^{n\times d}
}
Same convention applies when $\bvm^{(k)}$ adds or subtracts with the stacked variables like $\vx^{(k)}$ and $\vm^{(k)}$.

\textbf{Algorithm reformulation.} With the above notations, DmSGD (Algorithm \ref{Algorithm: DmSGD}) can be re-written as a more elegant vector-matrix form. For $k=0,1,\cdots$, DmSGD with one-peer exponential graph will iterate as follows:
\begin{align}
    \vg^{(k)} =\,&  \nabla F(\vx^{(k)}; \bxi^{(k)}), \\
    \vm^{(k+1)} =\,& W^{(k)}(\beta\vm^{(k)}  + \vg^{(k)}  ),  \label{msgd_recursion}\\
    \vx^{(k+1)} =\,& W^{(k)}(\vx^{(k)}  - \gamma \vm^{(k)} ),  \label{msgd_recursion-2}
\end{align} 
where $\vm^{(0)} = 0$, $\vx^{(0)}$ can be set arbitrarily, and $W^{(k)}$ is the weight matrix associated with the one-peer exponential graph defined by \eqref{wij-one-exp}. Note that the weight matrix sequence $\{W^{(k)}\}$ satisfies the periodic exact averaging property, see Lemma \ref{lm-perioidc-exact-averaging}. 

\noindent \textbf{Smoothness.} Since each $f_i(x)$ is assumed to be $L$-smooth in Assumption A.3, it holds that $f(x) = \frac{1}{n}\sum_{i=1}^n f_i(x)$ is also $L$-smooth. As a result, the following inequality holds for any $\x, \y \in \mathbb{R}^d$:
\begin{align}
f(\x) - f(\y) - \frac{L}{2}\|\x - \y\|^2 &\le  \langle \nabla f(\y), \x- \y \rangle \label{sdu-2}
\end{align}

{\color{black}\textbf{Submultiplicativity of the Frobenius norm.} Given matrices $W\in \RR^{n\times n}$ and $\vy\in \RR^{n\times d}$, we have
\begin{align}\label{submulti}
\|W\vy\| \le \|W\|_2 \|\vy\|.
\end{align}
To verify it, by letting $y_j$ be the $j$-th column of $\vy$, we have $\|W\vy\|^2 = \sum_{j=1}^d \|Wy_j\|_2^2 \le \sum_{j=1}^d \|W\|_2^2 \|y_j\|_2^2=\|W\|_2^2\|\vy\|^2$.}

\textbf{DmSGD: the averaged recursion.}
Multiplying $\frac{1}{n}\one_n^T$ to   both sides of \eqref{msgd_recursion} and \eqref{msgd_recursion-2}, we establish the averaged (or centralized) recursion:
\begin{align}
    \bvm^{(k+1)} =\,& \beta\bvm^{(k)} + \bvg^{(k)}\\
    \bvx^{(k+1)} =\,& \bvx^{(k)} - \gamma \bvm^{(k)}
\end{align}
where $\bvg^{(k)} \triangleq \frac{1}{n}\mathds{1}_n^T\vg^{(k)}$. 


\textbf{A critical auxiliary recursion.} We also need to introduce a auxiliary sequence $\{\bvz^{(k)}\}$, which is commonly used for the convergence analysis in momentum methods \cite{yu2019linear, loizou2020momentum, balu2020decentralized}:
\begin{align}\label{xnzznz897}
    \bvz^{(k)} =  \frac{1}{1-\beta}  (\bvx^{(k)} - \beta \bvx^{(k-1)}),\;\;  \bvz^{(0)} =  \frac{1}{\beta}  \bvx^{(0)}
\end{align}
It is easy to validate that \cite[lemma 3]{yu2019linear}:
\begin{align} \label{expand.z_bar}
    \bvz^{(k+1)} - \bvz^{(k)} = -\frac{\gamma}{1-\beta}\bvg^{(k)}
\end{align}
{\color{black}When $\beta = 0$ and $\bar{\vz}^{(0)} = \bvx^{(0)}$, recursion \eqref{xnzznz897} reduces to},
\eq{
     \bvz^{(k)}= \bvx^{(k)},\;\;\; \bvx^{(k+1)} = \bvx^{(k)} -\gamma\bvg^{(k)}
}

\textbf{Main idea to prove Theorem \ref{thm-convergence-one-peer}.} Theorem \ref{thm-convergence-one-peer} can be proved in  two steps. First, we need to establish a decent lemma on how $f(\bvz^{(k)})$ would evolve as iteration increases. Second, we will establish a consensus lemma showing that the consensus distance $\Ex\|\vx^{(k)} - \bvx^{(k)}\|^2$ would gradually decrease to zero. These two lemmas together will lead to the result in Theorem \ref{thm-convergence-one-peer}. 

\subsection{Descent lemma}
\begin{lemma} \label{lm:descent}
Suppose the learning rate satisfies the condition $\gamma \leq\frac{(1-\beta)^2}{2 (1+\beta)L}$, the main recursion of \eqref{msgd_recursion}-\eqref{msgd_recursion-2} under the Assumption A.1 - A.4 has
\eq{
   \frac{1}{T+1}\sum_{k=1}^T 
    \Ex  \|\nabla f(\bvx^{(k)})\|^2 \leq& \frac{2(1-\beta)}{\gamma(T+1)} (\Ex f(\bvz^{(0)}) -  f^\star ) + \frac{\gamma L }{n(1-\beta)} \sigma^2 + \frac{\beta  L \gamma}{n(1-\beta)^2}\sigma^2  \nn\\
     &\;\;\; + \frac{L^2}{T+1}\sum_{k=0}^T \Ex \| \bvx^{(k)} - \vx^{(k)} \|^2
}
where $f^\star$ is the minimum value of the problem \eqref{dist-opt}; $\sigma^2$ and $L$ are the constants defined in Assumptions.
\end{lemma}
{\bf Proof}. Utilizing the $L$-smooth assumption of loss function $f$ -- Eq. \eqref{sdu-2}, we have:
\begin{align}
    &\hspace{-5mm}\Ex f(\bvz^{(k+1)}) \nn\\
    \leq\,& \Ex f(\bvz^{(k)}) + \Ex \langle \bvz^{(k+1)} - \bvz^{(k)}, \nabla f(\bvz^{(k)}) \rangle + \frac{L}{2} \Ex \|\bvz^{(k+1)} - \bvz^{(k)}\|^2 \nn\\
    \stackrel{(a)}{=}\,& \Ex f(\bvz^{(k)}) - \frac{\gamma}{1-\beta} \Ex \langle \bvg^{(k)}, \nabla f(\bvz^{(k)})\rangle  + \frac{\gamma^2L}{2(1-\beta)^2} \Ex \|\bvg^{(k)}\|^2\nn\\
    \stackrel{(b)}{=}\,& \Ex f(\bvz^{(k)}) - \frac{\gamma}{1-\beta} \Ex \left\langle \frac{1}{n}\sum_{i=1}^n \nabla f_i(\bx_i^{(k)}), \nabla f(\bvz^{(k)}) \right\rangle + \frac{\gamma^2L}{2(1-\beta)^2} \Ex \|\bvg^{(k)}\|^2 \label{main-recursion-expand}
\end{align}
where step (a) expands $\bvz^{(k+1)} - \bvz^{(k)}$ according to \eqref{expand.z_bar} and  step (b) utilizes the unbiased and independent assumption of gradient noise (Assumption A.2):
\eq{
    &\hspace{-5mm}\Ex \langle\bvg^{(k)}, \nabla f(\bvz^{(k)})\rangle\nn\\
    =& \Ex \left\langle \frac{1}{n}\sum_{i=1}^n \nabla F_i(\bx_i^{(k)}\,;\, \bxi_i) - \frac{1}{n}\sum_{i=1}^n \nabla f_i(\bx_i^{(k)}) + \frac{1}{n}\sum_{i=1}^n \nabla f_i(\bx_i^{(k)}), \nabla f(\bvz^{(k)}) \right\rangle\nn\\
    =& \Ex \left\langle \frac{1}{n}\sum_{i=1}^n \nabla F_i(\bx_i^{(k)}\,;\, \bxi_i) - \frac{1}{n}\sum_{i=1}^n \nabla f_i(\bx_i^{(k)}), \nabla f(\bvz^{(k)}) \right\rangle + \Ex \left\langle \frac{1}{n}\sum_{i=1}^n \nabla f_i(\bx_i^{(k)}), \nabla f(\bvz^{(k)}) \right\rangle \nn\\
    =&  \Ex \left\langle \frac{1}{n}\sum_{i=1}^n \nabla f_i(\bx_i^{(k)}), \nabla f(\bvz^{(k)}) \right\rangle \label{gh9s}
}
Next, we focus on bounding the middle term in \eqref{main-recursion-expand}. First, we expand it into:
\begin{align} \label{sefesg}
    &\hspace{-8mm}- \Ex \left\langle \frac{1}{n}\sum_{i=1}^n \nabla f_i(\bx_i^{(k)}), \nabla f(\bvz^{(k)}) \right\rangle  \\
    =&\; - \Ex \left(\frac{1}{n}\sum_{i=1}^n \nabla f_i(\bx_i^{(k)})\right)^T (\nabla f(\bvz^{(k)}) - \nabla f(\bvx^{(k)})) - \Ex \left(\frac{1}{n}\sum_{i=1}^n \nabla f_i(\bx_i^{(k)})\right)^T  \nabla f(\bvx^{(k)}) \nn
\end{align}
To bound the first term in \eqref{sefesg}, we apply the Young's inequality -- $a^T b \leq \frac{1}{2\epsilon}\|a\|^2 +  \frac{\epsilon}{2}\|b\|^2 $:
\eq{
    &\hspace{-9mm}-\left(\frac{1}{n}\sum_{i=1}^n \nabla f_i(\bx_i^{(k)})\right)^T (\nabla f(\bvz^{(k)}) - \nabla f(\bvx^{(k)}))\nn\\
    \leq&\, \frac{1}{2\epsilon}\|\frac{1}{n}\sum_{i=1}^n \nabla f_i(\bx_i^{(k)})\|^2 + \frac{\epsilon}{2}\|\nabla f(\bvz^{(k)}) - \nabla f(\bvx^{(k)})\|^2 \nn\\
    \leq&\,\frac{1}{2\epsilon}\|\frac{1}{n}\sum_{i=1}^n \nabla f_i(\bx_i^{(k)})\|^2  + \frac{\epsilon L^2}{2} \| \bvz^{(k)} - \bvx^{(k)}\|^2\nn\\
    = &\,\frac{1}{2\epsilon}\|\frac{1}{n}\sum_{i=1}^n \nabla f_i(\bx_i^{(k)})\|^2  + \frac{\epsilon L^2
    \beta^2\gamma^2}{2(1-\beta)^2} \| \bvm^{(k)}\|^2
}
where the last equality relied on the observation that 
\eq{
    \bvz^{(k)} - \bvx^{(k)} = \frac{\beta}{1-\beta} [\bvx^{(k)} - \bvx^{(k-1)}] = -\frac{\beta\gamma}{1-\beta} \bvm^{(k)}
}

If we choose $\epsilon = \frac{(1-\beta)^2}{\gamma\beta L}$, we obtain:
\eq{\label{289hg23}
    -\left(\frac{1}{n}\sum_{i=1}^n\nabla f_i(\bx_i^{(k)})\right)^T \hspace{-1.5mm}(\nabla f(\bvz^{(k)}) - \nabla f(\bvx^{(k)})) \leq
    \frac{\beta L \gamma }{2(1-\beta)^2}\|\frac{1}{n}\sum_{i=1}^n \nabla f_i(\bx_i^{(k)})\|^2  + \frac{
    \beta L \gamma}{2} \| \bvm^{(k)}\|^2
}
To bound the second term in \eqref{sefesg}, we use the identity that $a^T b = \frac{1}{2}\left(\|a\|^2 + \|b\|^2 -  \|a - b \|^2  \right)$:
\eq{\label{23gjio23t}
    &\left(\frac{1}{n}\sum_{i=1}^n \nabla f_i(\bx_i^{(k)})\right)^T\nabla f(\bvx^{(k)})\nn \\
    =&\frac{1}{2} \left(\|\nabla f(\bvx^{(k)})|\|^2 + \|\frac{1}{n}\sum_{i=1}^n \nabla f_i(\bx_i^{(k)}) \|^2 - \| \nabla f(\bvx^{(k)}) - \frac{1}{n}\sum_{i=1}^n f_i(\bx_i^{(k)}) \|^2\right)\nn\\
    \geq&\frac{1}{2} \left(\|\nabla f(\bvx^{(k)})|\|^2 + \|\frac{1}{n}\sum_{i=1}^n \nabla f_i(\bx_i^{(k)}) \|^2 - \frac{L^2}{n}\sum_{i=1}^n\| \bvx^{(k)} - \bx_i^{(k)} \|^2\right) \nn\\
     =&\frac{1}{2} \left(\|\nabla f(\bvx^{(k)})|\|^2 + \|\frac{1}{n}\sum_{i=1}^n \nabla f_i(\bx_i^{(k)}) \|^2 - L^2\| \bvx^{(k)} - \vx^{(k)} \|^2\right)
}
Noting \eqref{289hg23} and \eqref{23gjio23t} hold for all realization. Substituting them back to \eqref{sefesg} and simplifying the terms, the middle term in \eqref{main-recursion-expand} is bounded as follows
\eq{
    &\hspace{-1mm} - \frac{\gamma}{1-\beta} \Ex \left(\frac{1}{n}\sum_{i=1}^n \nabla f_i(\bx_i^{(k)})\right)^T (\nabla f(\bvz^{(k)})) \nn\\
    &\leq   \frac{\gamma^2 \beta L }{2(1-\beta)^3}\Ex \|\frac{1}{n}\sum_{i=1}^n \nabla f_i(\bx_i^{(k)})\|^2  + \frac{
    \beta L \gamma^2}{2(1-\beta)} \Ex \| \bvm^{(k)}\|^2 + \frac{L^2\gamma}{2(1-\beta)n}\sum_{i=1}^n\Ex\| \bvx^{(k)} - \bx_i^{(k)} \|^2\nn\\
    &\hspace{4mm} - \frac{\gamma}{2(1-\beta)}\Ex\|\nabla f(\bvx^{(k)})\|^2 - \frac{\gamma}{2(1-\beta)} \Ex\|\frac{1}{n}\sum_{i=1}^n \nabla f_i(\bx_i^{(k)}) \|^2 \nn\\
    &=   \frac{
    \beta L \gamma^2}{2(1-\beta)} \Ex \| \bvm^{(k)}\|^2 + \frac{L^2\gamma}{2(1-\beta)}\Ex\| \bvx^{(k)} - \vx^{(k)} \|^2\nn\\
    &\hspace{4mm} - \frac{\gamma}{2(1-\beta)}\Ex\|\nabla f(\bvx^{(k)})\|^2 - \frac{\gamma}{2(1-\beta)} \left( 1 - \frac{\gamma \beta L}{(1-\beta)^2}\right)\Ex\|\frac{1}{n}\sum_{i=1}^n \nabla f_i(\bx_i^{(k)}) \|^2 \label{giofs}
}
The third term in the main recursion \eqref{main-recursion-expand} can be bounded as similar as we did in \eqref{gh9s}:
\eq{ 
    \Ex \|\bvg^{(k)}\|^2  =& \Ex \left\|\frac{1}{n}\sum_{i=1}^n \nabla f_i(\bx_i^{(k)})  + \frac{1}{n}\sum_{i=1}^n \nabla F_i(\bx_i^{(k)}\,;\, \bxi_i) - \frac{1}{n}\sum_{i=1}^n \nabla f_i(\bx_i^{(k)}) \right\|^2\nn\\
     \stackrel{(a)}{=}& \Ex \left\|\frac{1}{n}\sum_{i=1}^n \nabla f_i(\bx_i^{(k)}) \right\|^2  + \Ex \left\|\frac{1}{n}\sum_{i=1}^n \nabla F_i(\bx_i^{(k)}\,;\, \bxi_i) - \frac{1}{n}\sum_{i=1}^n \nabla f_i(\bx_i^{(k)})  \right\|^2\nn\\
    \stackrel{(b)}{=}& \Ex \left\|\frac{1}{n}\sum_{i=1}^n \nabla f_i(\bx_i^{(k)}) \right\|^2  + \frac{1}{n^2}\sum_{i=1}^n \Ex \left\|\nabla F_i(\bx_i^{(k)}\,;\, \bxi_i) - \nabla f_i(\bx_i^{(k)}) \right\|^2\nn\\
    \leq& \Ex \left\|\frac{1}{n}\sum_{i=1}^n\nabla f_i(\bx_i^{(k)})\right\|^2 + \frac{1}{n}\sigma^2 \label{ghsdkh}
}
where the step (a) that separates the norm square term into the sum of two terms is thanks to the independent assumption of gradient noise over the past data; the step (b) is one of the key step that relied on the independent assumption of gradient noise across the agents.
Substituting \eqref{giofs} and \eqref{ghsdkh} back to main recursion \eqref{main-recursion-expand} and re-organize the terms, we establish
\eq{
    \frac{\gamma}{2(1-\beta)}\Ex\|\nabla f(\bvx^{(k)})\|^2 \leq\,& \Ex f(\bvz^{(k)}) -  \Ex f(\bvz^{(k+1)}) + \frac{\gamma^2L}{2(1-\beta)^2}\Ex \left\|\frac{1}{n}\sum_{i=1}^n\nabla f_i(\bx_i^{(k)})\right\|^2 \nn\\
    &\;\; + \frac{
    \beta L \gamma^2}{2(1-\beta)} \Ex \| \bvm^{(k)}\|^2 + \frac{L^2\gamma}{2(1-\beta)}
    \Ex\| \bvx^{(k)} - \vx^{(k)} \|^2
    \nn\\
    &\;\; - \frac{\gamma}{2(1-\beta)} \left( 1 - \frac{\gamma \beta L}{(1-\beta)^2}\right)\Ex\|\frac{1}{n}\sum_{i=1}^n \nabla f_i(\bx_i^{(k)}) \|^2  + \frac{\gamma^2L}{2(1-\beta)^2n}\sigma^2 \label{23gdsh}
}

Next step is to expand the momentum term $\Ex \| \bvm^{(k)}\|^2 $ back to the first iteration:
\eq{
    \bvm^{(k)} 
    =&\, \sum_{t=0}^{k-1}\beta^{k-1-t} \bvg^{(t)}
}
Taking the expectation and norm square on both sides, we have
\eq{
    \Ex \|\bvm^{(k)} \|^2 \stackrel{(a)}{\leq} & \Ex\left\|\frac{s_k}{s_k}\sum_{t=0}^{k-1}\beta^{k-1-t}\bvg^{(t)}\right\|^2\nn\\
    \stackrel{(b)}{\leq}& s_k \sum_{t=0}^{k-1} \beta^{k-1-t} \Ex\left\|\bvg^{(t)}\right\|^2 \nn\\
    \stackrel{(c)}{\leq}& s_k \sum_{t=0}^{k-1} \beta^{k-1-t} \Ex\left\| \frac{1}{n}\sum_{i=1}^n\nabla f_i(\bx_i^{(t)})\right\|^2 + \frac{s_k^2}{n}\sigma^2 \label{2giosd}
}
where in step (a), we define $s_k \triangleq \sum_{t=0}^{k-1} \beta^{k-1-t} $ as the sum of weights; we applied the Jensen's inequality in the step (b); step (c) used the conclusion from  \eqref{ghsdkh}. Note the sum of weight $s_k$ is bounded by constant:
\eq{
 s_k = \frac{1-\beta^k}{1-\beta} \leq \frac{1}{1-\beta} \label{s_k.upper-bound}
}

Plug it back to \eqref{23gdsh}, we have
\eq{
    \frac{\gamma}{2(1-\beta)}\Ex\|\nabla f(\bvx^{(k)})\|^2 \leq\,& \Ex f(\bvz^{(k)}) -  \Ex f(\bvz^{(k+1)}) + \frac{\gamma^2 L}{2n(1-\beta)^2}\sigma^2 + \frac{
   s_k^2 \beta  L \gamma^2}{2n(1-\beta)}\sigma^2  \nn\\
    &\;\; + \frac{
    s_k  \beta  L \gamma^2}{2(1-\beta)} \sum_{t=0}^{k-1} \beta^{k-1-t} \Ex\left\| \frac{1}{n}\sum_{i=1}^n\nabla f_i(\bx_i^{(t)})\right\|^2 \nn\\
    &\;\;\;+ \frac{L^2\gamma}{2(1-\beta)}\Ex\| \bvx^{(k)} - \vx^{(k)} \|^2
    \nn\\
    &\;\; - \frac{\gamma}{2(1-\beta)} \left( 1 -\frac{\gamma L}{1-\beta} - \frac{\gamma \beta L}{(1-\beta)^2}\right) \Ex \|\frac{1}{n}\sum_{i=1}^n \nabla f_i(\bx_i^{(k)}) \|^2
    \label{hgbuis}
}
Taking the average of \eqref{hgbuis} from time $k=0$ to $k=T$, we have
\eq{
    &\hspace{-5mm}\frac{\gamma}{2(T+1)(1-\beta)}\sum_{k=0}^T  \Ex\|\nabla f(\bvx^{(k)})\|^2 \nn\\
    &\leq \frac{1}{T+1} (\Ex f(\bvz^{(0)}) -  \Ex f(\bvz^{(T+1)})) + \frac{\gamma^2 L}{2n(1-\beta)^2}\sigma^2 + \frac{\beta  L \gamma^2}{2n(1-\beta)^3}\sigma^2\nn\\
    &\;\; + \frac{1}{T+1}\sum_{k=0}^T   \frac{
    s_k  \beta  L \gamma^2}{2(1-\beta)} \sum_{t=0}^{k-1} \beta^{k-1-t} \Ex\left\| \frac{1}{n}\sum_{i=1}^n\nabla f_i(\bx_i^{(t)})\right\|^2 \nn\\
     &\;\;\;+ \frac{L^2\gamma}{2(1-\beta)}\frac{1}{T+1}\sum_{k=0}^T\Ex\| \bvx^{(k)} - \vx^{(k)} \|^2
    \nn\\
    &\;\; - \frac{\gamma}{2(1-\beta)} \left( 1 -\frac{\gamma L}{1-\beta} - \frac{\gamma \beta L}{(1-\beta)^2}\right) \frac{1}{T+1}\sum_{k=0}^T \Ex \|\frac{1}{n}\sum_{i=1}^n \nabla f_i(\bx_i^{(k)}) \|^2 \label{df9hgre}
}
Focus on the term in the second line of r.h.s of \eqref{df9hgre}:
\eq{
    &\hspace{-6mm}\frac{1}{T+1}\sum_{k=0}^T   \frac{
    s_k  \beta  L \gamma^2}{2(1-\beta)} \sum_{t=0}^{k-1} \beta^{k-1-t} \Ex\left\| \frac{1}{n}\sum_{i=1}^n\nabla f_i(\bx_i^{(t)})\right\|^2\nn\\
    \stackrel{(a)}{\leq}& \frac{\beta  L \gamma^2}{2(1-\beta)^2} \frac{1}{T+1}\sum_{k=0}^T 
     \sum_{t=0}^{k-1} \beta^{k-1-t} \Ex\left\| \frac{1}{n}\sum_{i=1}^n\nabla f_i(\bx_i^{(t)})\right\|^2\nn\\
    \stackrel{(b)}{=}& \frac{\beta  L \gamma^2}{2(1-\beta)^2} \frac{1}{T+1}   
     \sum_{t=0}^{T-1} \sum_{k=t+1}^T\beta^{k-1-t} \Ex\left\| \frac{1}{n}\sum_{i=1}^n\nabla f_i(\bx_i^{(t)})\right\|^2\nn\\
    \stackrel{(c)}{\leq}& \frac{\beta  L \gamma^2}{2(1-\beta)^3} \frac{1}{T+1}\sum_{k=0}^{T}
    \Ex\left\| \frac{1}{n}\sum_{i=1}^n\nabla f_i(\bx_i^{(k)})\right\|^2
}
where step (a) uses the upper bound of $s_k$ in \eqref{s_k.upper-bound}; step (b) switches the order of two summations; step (c), again, uses the upper bound of $s_k$ and re-align the index of summation due to non-negativity of each term. Hence, we establish
\eq{
    &\hspace{-5mm}\frac{\gamma}{2(T+1)(1-\beta)}\sum_{k=1}^T  \Ex\|\nabla f(\bvx^{(k)})\|^2 \nn\\
    &\leq \frac{1}{T+1} \left(\Ex f(\bvz^{(0)}) -  \Ex f(\bvz^{(T+1)})\right) + \frac{\gamma^2 L}{2n(1-\beta)^2}\sigma^2 + \frac{\beta  L \gamma^2}{2n(1-\beta)^3}\sigma^2\nn\\
     &\;\;\;+ \frac{L^2\gamma}{2(1-\beta)}\frac{1}{T+1}\sum_{k=0}^T\Ex\| \bvx^{(k)} - \vx^{(k)} \|^2
    \nn\\
    &\;\; - \frac{\gamma}{2(1-\beta)} \left( 1 -\frac{\gamma L}{1-\beta} - \frac{2\gamma \beta L}{(1-\beta)^2}\right) \frac{1}{T+1}\sum_{k=0}^T \Ex \|\frac{1}{n}\sum_{i=1}^n \nabla f_i(\bx_i^{(k)}) \|^2 \label{23ghgsd}
}
In order to discard the $ \Ex \|\frac{1}{n}\sum_{i=1}^n \nabla f_i(\bx_i^{(k)}) \|^2$, the step-size has to be small enough so that the coefficient is negative. To achieve that, we need $ 1 -\frac{\gamma L }{1-\beta} - \frac{2\gamma \beta L}{(1-\beta)^2} \geq 0$. The idea is we can require last two terms bounded by two constants, which sum up to 1. Suppose we require that:
\eq{
    \frac{\gamma L}{1-\beta} \leq  \frac{1-\beta}{1-\beta^2}\;\; \Longrightarrow&\;\;\gamma \leq \frac{1-\beta}{(1+\beta) L} \label{g3d-1}\\
    \frac{2\gamma \beta L}{(1-\beta)^2} \leq \frac{\beta(1-\beta)}{1-\beta^2} \;\;\Longrightarrow&\;\; \gamma \leq\frac{(1-\beta)^2}{2 (1+\beta)L}\label{g3d-2}
}
Since $1>\beta \geq 0$, \eqref{g3d-2} is always smaller than \eqref{g3d-1}. So as long as $\gamma \leq\frac{(1-\beta)^2}{2 (1+\beta)L}$, we can safely discard the last terms in \eqref{23ghgsd}. 

Finally, we arrive at the conclusion in the lemma by noting $f^\star$ is the minimum value of the problem:
\eq{
   \frac{1}{T+1}\sum_{k=0}^T 
    \Ex  \|\nabla f(\bvx^{(k)})\|^2 \leq& \frac{2(1-\beta)}{\gamma(T+1)} (\Ex f(\bvz^{(0)}) -  f^\star ) + \frac{\gamma L }{n(1-\beta)} \sigma^2 + \frac{\beta  L \gamma}{n(1-\beta)^2}\sigma^2  \nn\\
     &\;\;\; + \frac{L^2}{T+1}\sum_{k=0}^T \Ex \| \bvx^{(k)} - \vx^{(k)} \|^2
}
A few comments about this bounds: the historical average of gradient at the average trajectory $\bvx^{(k)}$ is bounded by the excess risk at the initial value, the gradient noise, and the average of the consensus residue over the time.
$\qed$

\subsection{Consensus lemma} \label{apx.sub.proof-dmsgd-convergence-3}
Before we can bound the consensus residue of the DmSGD algorithm, we transform the main recursion \eqref{msgd_recursion} and \eqref{msgd_recursion-2} into the following consensus residue form, which is much easier for analysis.

Because of the periodic exact averaging property, we can view the main recursion in every $\tau$ iterations as reference point. Recall $\tau = \ln(n)$ which is an integer. We define $m = \lfloor k/\tau \rfloor -1$. (More precisely, $m$ should be a function of $k$. $m(k)$ would be more proper but we choose $m$ to light the notation). Apparently, it holds that $2\tau > k - m\tau \ge \tau$ . It implies from iteration $k$ to $m\tau$ it must contain as least one period. 

\begin{lemma} If we expand the recursion from iteration $k$ to the previous period $m\tau$, it has following concise form due to exact averaging property:
\eq{
     \vx^{(k)}  - \bvx^{(k)} &= -\gamma\sum_{t=m\tau}^{k-1} \left(\sum_{j=t+1}^{k-1}\beta^{j-1 - t} \right)  \left(\prod_{i=t}^{k-1} \ww^{(i)} \right) (\vg^{(t)} - {\bar{\vg}}^{(t)}), \;\;\; \forall k \geq \tau
}
where $\ww^{(i)} \triangleq W^{(i)}-\frac{1}{n}\mathds{1}\mathds{1}^T$.
\end{lemma}


\noindent \textbf{Proof}: Recalling that decentralized momentum SGD in \eqref{msgd_recursion} subtract it by the centralized recursion:
\eq{
    \vx^{(k)}  - \bvx^{(k)}&= W^{(k-1)}(\vx^{(k-1)} - \bvx^{(k-1)} - \gamma (\vm^{(k-1)} - \bvm^{(k-1)}))\nn\\
    &= \left(W^{(k-1)}-\frac{1}{n}\mathds{1}\mathds{1}^T\right)(\vx^{(k-1)} - \bvx^{(k-1)} - \gamma (\vm^{(k-1)} - \bvm^{(k-1)}))
}
where we utilized the average of the average value is still the average value: $ W^{(k-1)} \bvx^{(k-1)} = \bvx^{(k-1)}$ and $ W^{(k-1)} \bvm^{(k-1)} = \bvm^{(k-1)}$. For the short notation, we denote that
\eq{
    \ww^{(k-1)} := W^{(k-1)}-\frac{1}{n}\mathds{1}\mathds{1}^T
}
For any $k\ge \tau$, we can always expand the recursion into $m\tau$:
\begin{align}
 \vx^{(k)}  - \bvx^{(k)} &= \ww^{(k-1)}(\vx^{(k-1)} - \bvx^{(k-1)} - \gamma (\vm^{(k-1)} - \bvm^{(k-1)})) \nonumber \\
&= \prod_{i=m\tau}^{k-1} \ww^{(i)} (\vx^{(m\tau)} - \bvx^{(m\tau)}) - \gamma \sum_{j=m\tau}^{k-1} \prod_{i=j}^{k-1} \ww^{(i)} (\vm^{(j)} - \bvm^{(j)}) \nonumber \\
&\stackrel{(a)}{=} -\gamma \sum_{j=m\tau}^{k-1} \prod_{i=j}^{k-1} \ww^{(i)}  (\vm^{(j)} - \bvm^{(j)}) \nn\\
&= -\gamma \sum_{j=m\tau}^{k-1} \prod_{i=j}^{k-1} W^{(i)}  (\vm^{(j)} - \bvm^{(j)}) \label{j9ioa}
\end{align}
where step (a) discards the first term because of the periodic exact averaging property in Lemma \ref{lm-perioidc-exact-averaging}.  To evaluate the sum of production term in \eqref{j9ioa}, we first expand the momentum term according the recursion \eqref{msgd_recursion} until iteration $m\tau$
\begin{align}
\vm^{(j)}  =\,& \beta^{j - m\tau} \prod_{i=m\tau}^{j-1}  W^{(i)} \vm^{(m\tau)}  + \sum_{t=m\tau}^{j-1}\beta^{j-1 - t} \prod_{q=t}^{j-1}  W^{(q)}  \vg^{(t)} 
\end{align}
Multiplying $\prod_{i=j}^{k-1} W^{(i)}$ on both sides and note we can exchange the order of $\prod_i$ and $\sum_t$ when their index is not dependent:
\eq{
    \prod_{i=j}^{k-1} W^{(i)} \vm^{(j)} =& \beta^{j - m\tau} \prod_{i=j}^{k-1} W^{(i)} \prod_{i=m\tau}^{j-1}  W^{(i)} \vm^{(m\tau)}  + \sum_{t=m\tau}^{j-1}\beta^{j-1 - t} \prod_{i=j}^{k-1} W^{(i)} \prod_{q=t}^{j-1}  W^{(q)}  \vg^{(t)} \nonumber\\
    =&\beta^{j - m\tau}  \prod_{i=m\tau}^{k-1} W^{(i)}\vm^{(m\tau)} + \sum_{t=m\tau}^{j-1}\beta^{j-1 - t} \prod_{i=t}^{k-1} W^{(i)}\vg^{(t)} \nonumber\\
    =&\beta^{j - m\tau}  \bvm^{(m\tau)} + \sum_{t=m\tau}^{j-1}\beta^{j-1 - t} \prod_{i=t}^{k-1} W^{(i)}\vg^{(t)}
}
where the last equality is, again, thanks to the periodic exact averaging property.
We can establish the similar conclusion for average momentum term:
\eq{
     \prod_{i=j}^{k-1} W^{(i)}   \bvm^{(j)}  = \bvm^{(j)} =& \beta^{j - m\tau}\bvm^{(m\tau)} + \sum_{t=m\tau}^{j-1}\beta^{j-1 - t} {\bar{\vg}}^{(t)} \nn\\
    =& \beta^{j - m\tau}\bvm^{(m\tau)} + \sum_{t=m\tau}^{j-1}\beta^{j-1 - t}  \prod_{i=t}^{k-1} W^{(i)} {\bar{\vg}}^{(t)}
}
Combining above two,  we get
\eq{
    \prod_{i=j}^{k-1} W^{(i)}  (\vm^{(j)} - \bvm^{(j)})=& \sum_{t=m\tau}^{j-1}\beta^{j-1 - t}  \prod_{i=t}^{k-1} W^{(i)} (\vg^{(t)} - {\bar{\vg}}^{(t)}) \nn\\
    =& \sum_{t=m\tau}^{j-1}\beta^{j-1 - t}  \prod_{i=t}^{k-1} \ww^{(i)} (\vg^{(t)} - {\bar{\vg}}^{(t)}) \label{si.gje}
}
Substituting \eqref{si.gje} back to \eqref{j9ioa}, we establish
\eq{
    \vx^{(k)}  - \bvx^{(k)} &= -\gamma\sum_{j=m\tau}^{k-1} \sum_{t=m\tau}^{j-1}\beta^{j-1 - t}  \prod_{i=t}^{k-1} \ww^{(i)} (\vg^{(t)} - {\bar{\vg}}^{(t)}) 
}
Note we can switch the order of two summations:
\eq{
    \sum_{j=m\tau}^{k-1} \sum_{t=m\tau}^{j-1} \equiv \sum_{t=m\tau}^{k-1} \sum_{j=t+1}^{k-1}
}
By above identity, we can group the coefficients and finally arrive at
\eq{
     \vx^{(k)}  - \bvx^{(k)} &= -\gamma\sum_{t=m\tau}^{k-1} \left(\sum_{j=t+1}^{k-1}\beta^{j-1 - t} \right)  \left(\prod_{i=t}^{k-1} \ww^{(i)} \right) (\vg^{(t)} - {\bar{\vg}}^{(t)})  \label{ns0dfea}
}
$\qed$

With this simplified consensus residue form \eqref{ns0dfea}, we ready to present the consensus lemma.
\begin{lemma}[Consensus Lemma]\label{lm:consensus} Suppose the learning rate satisfies the condition $\gamma \leq\frac{1-\beta}{6 L \tau }$ and Assumption A.1 - A.4 holds, the consensus residue have
\eq{
   \frac{1}{T+1}\sum_{k=0}^{T}\ExNorm{\vx^{(k)}  - \bvx^{(k)} } \leq \frac{8\tau \gamma^2 \rho_{\rm max}^2}{(1-\beta)^2}(\sigma^2+ 4\tau b^2) + \frac{2}{(T+1)}\sum_{k=0}^{\tau-1}\ExNorm{\vx^{(k)}  - \bvx^{(k)}}
}
where $\sigma^2$ and $b^2$ are the constants defined in Assumptions for gradient noise and data heterogeneous respectively; the spectral gap $ \rho_{\rm max}^2$ is defined as 
\eq{
     \rho_{\rm max}^2 = \max_{i\in [0, \tau-1]}\left\|\ww^{(i)} \right\|^2_2 \leq 1
}
\end{lemma}
\noindent {\bf Proof}.
Taking norm and expectation on both sides of \eqref{ns0dfea}, we obtain
\eq{
 &\hspace{-8mm}\ExNorm{\vx^{(k)}  - \bvx^{(k)} } \nn\\
 =& \gamma^2 \ExNorm{ \sum_{t=m\tau}^{k-1} \left(\sum_{j=t+1}^{k-1}\beta^{j-1 - t} \right)  \left(\prod_{i=t}^{k-1} \ww^{(i)} \right) (\vg^{(t)} - {\bar{\vg}}^{(t)}) } \nn\\
 \leq& \underbrace{2 \gamma^2 \ExNorm{ \sum_{t=m\tau}^{k-1} \left(\sum_{j=t+1}^{k-1}\beta^{j-1 - t} \right)  \left(\prod_{i=t}^{k-1} \ww^{(i)} \right)  (\nabla F(\vx^{(t)}) - \nabla f(\vx^{(t)}) )}}_{:=(A)} \nn\\
 &\;\; + \underbrace{2 \gamma^2 \ExNorm{ \sum_{t=m\tau}^{k-1} \left(\sum_{j=t+1}^{k-1}\beta^{j-1 - t} \right)  \left(\prod_{i=t}^{k-1} \ww^{(i)} \right)  (\vg^{(t)} - {\bar{\vg}}^{(t)} - \nabla F(\vx^{(t)}) + \nabla f(\vx^{(t)}))}}_{:=(B)} \label{2389hgh}
}
where the inequality is due to Jensen's inequality. First, let's exam the second term in \eqref{2389hgh}, which contains the gradient noise only
\eq{
    (B) \stackrel{(a)}{=}& 2 \gamma^2\sum_{t=m\tau}^{k-1}  \ExNorm{ \left(\sum_{j=t+1}^{k-1}\beta^{j-1 - t} \right)  \left(\prod_{i=t}^{k-1} \ww^{(i)} \right)  (\vg^{(t)} - {\bar{\vg}}^{(t)} - \nabla F(\vx^{(t)}) + \nabla f(\vx^{(t)})) } \nn\\
    \stackrel{(b)}{\leq} &  \frac{2\gamma^2}{1-\beta}\sum_{t=m\tau}^{k-1}  \sum_{j=t+1}^{k-1}\beta^{j-1 - t}  \ExNorm{ \left(\prod_{i=t}^{k-1} \ww^{(i)} \right)  (\vg^{(t)} - {\bar{\vg}}^{(t)} - \nabla F(\vx^{(t)}) + \nabla f(\vx^{(t)})) }\nn\\
    \stackrel{(c)}{\leq}&  \frac{2\gamma^2}{1-\beta}\sum_{t=m\tau}^{k-1}  \sum_{j=t+1}^{k-1}\beta^{j-1 - t}  \left\|\prod_{i=t}^{k-1} \ww^{(i)} \right\|^2_2 \ExNorm{ \vg^{(t)} - {\bar{\vg}}^{(t)} - \nabla F(\vx^{(t)}) + \nabla f(\vx^{(t)}) }\nn\\
    \stackrel{(d)}{\leq} &  \frac{2\gamma^2}{1-\beta}\sum_{t=m\tau}^{k-1}  \sum_{j=t+1}^{k-1}\beta^{j-1 - t}  \left\|\prod_{i=t}^{k-1} \ww^{(i)} \right\|^2 _2 \ExNorm{ \vg^{(t)} - \nabla F(\vx^{(t)}) }\nn\\
     \stackrel{(e)}{\leq}  &  \frac{2\gamma^2}{1-\beta}\sum_{t=m\tau}^{k-1}  \sum_{j=t+1}^{k-1}\beta^{j-1 - t}  \left\|\prod_{i=t}^{k-1} \ww^{(i)} \right\|^2_2 \sigma^2\nn\\
     \stackrel{(f)}{\leq}  & \frac{4\tau \gamma^2 \rho_{\rm max}^2}{(1-\beta)^2} \sigma^2 \label{239hd2}
}
where the step (a) is thanks to the independent properties of gradient noise; in the step (b), we apply the Jensen's inequality and loosen the sum of weights to $1/(1-\beta)$; step (c) utilized the submultiplicative property of norm; by noting that \eq{
    {\bar{\vg}}^{(t)} - \nabla f(\vx^{(t)}) = \frac{1}{n} \one_n \one_n^T \left(\vg^{(t)} - \nabla F(\vx^{(t)})\right) \label{2dbs}
}
step (d) applies the inequality $\|x - \bar{x}\|^2 \leq \|x\|^2$; step (e) is because of Assumption A.2; step (f) define that
\eq{
    \rho_{\rm max}^2 \triangleq \max_{k,t}  \left\|\prod_{i=t}^{k-1} \ww^{(i)} \right\|^2_2 \;\;\; \forall k \geq \tau,  t \in [m\tau, k-1]
}
It is easy to that for any $i$:
\eq{
    \left\|\ww^{(i)} \right\|^2_2 = \lambda_{\rm max}\left( (W^{(i)})^TW^{(i)} - \frac{1}{n}\one_n\one_n^T\right) \leq 1
}
where the inequality is thanks to the property of doubly stochastic matrix. (Noting $(W^{(i)})^TW^{(i)}$ is just a symmetric doubly stochastic matrix).
So using the sub-multiplicity property of matrix norm, $\rho_{\rm max}^2$ also equals to the following definition:
\eq{
    \rho_{\rm max}^2 := \max_{i\in[0, \tau-1]}\left\|\ww^{(i)} \right\|^2_2
}
We will revisit this quantity numerically later. In most of case, this $\rho_{\rm max}^2$ can be omitted since it equals to 1, but we keep it for the place-holder. Next, we can use the similar procedure for the first term in \eqref{2389hgh}. The difference is that the first step, we use Jensen's inequality to take the summation over $t$ out of the norm since we can no longer use the independent assumption about the noise: 
\eq{
    (A) \leq&  \frac{4\tau \gamma^2}{(1-\beta)^2}\sum_{t=m\tau}^{k-1}  \sum_{j=t+1}^{k-1}\beta^{j-1 - t}  \left\|\prod_{i=t}^{k-1} \ww^{(i)} \right\|^2_2 \Ex \|\nabla F(\vx^{(t)}) - \nabla f(\vx^{(t)})\|^2\nn\\
    \leq&  \frac{4\tau \gamma^2 \rho_{\rm max}^2}{(1-\beta)^2}\sum_{t=m\tau}^{k-1}   \Ex \|\nabla F(\vx^{(t)}) - \nabla \cf(\bvx^{(t)}) + \nabla \cf(\bvx^{(t)}) - \nabla f(\bvx^{(t)}) \nn\\
    &\hspace{31mm}+ \nabla f(\bvx^{(t)}) - \nabla f(\vx^{(t)})\|^2\nn\\
    \stackrel{(a)}{\leq}&  \frac{8\tau \gamma^2 \rho_{\rm max}^2}{(1-\beta)^2}\sum_{t=m\tau}^{k-1}   \left(\Ex \|\nabla F(\vx^{(t)}) - \nabla \cf(\bvx^{(t)}) + \nabla f(\bvx^{(t)}) - \nabla f(\vx^{(t)})\|^2  \right.\nn\\
    &\hspace{28mm} \left. + \Ex\|\nabla \cf(\bvx^{(t)}) - \nabla f(\bvx^{(t)}) \|^2\right)\nn\\
    \stackrel{(b)}{\leq}&  \frac{8\tau \gamma^2 \rho_{\rm max}^2}{(1-\beta)^2}\sum_{t=m\tau}^{k-1}   \left(\Ex \|\nabla F(\vx^{(t)}) - \nabla \cf(\bvx^{(t)})\|^2  + \Ex\|\nabla \cf(\bvx^{(t)}) - \nabla f(\bvx^{(t)}) \|^2\right)\nn\\
    \stackrel{(c)}{\leq}&  \frac{8\tau \gamma^2 \rho_{\rm max}^2}{(1-\beta)^2}\sum_{t=m\tau}^{k-1}   \left(L^2\Ex\|\vx^{(t)}-\bvx^{(t)}\|^2  + b^2\right)\nn\\
    \leq &  \frac{8\tau \gamma^2 \rho_{\rm max}^2 L^2}{(1-\beta)^2}\sum_{t=m\tau}^{k-1} \Ex\|\vx^{(t)}-\bvx^{(t)}\|^2  + \frac{16\tau^2 \gamma^2 \rho_{\rm max}^2}{(1-\beta)^2}b^2 \label{g9hsdwe}
}
where step (a) applied Jensen's inequality; step (b) is similar as \eqref{2dbs} by applying the inequality $\|x - \bar{x}\|^2 \leq \|x\|^2$; step (c) utilize the $L$-smoothness assumption and the data heterogeneous assumption (Assumption A.1 and A.3);

Plugging \eqref{239hd2} and \eqref{g9hsdwe} back to \eqref{2389hgh}, we establish
\eq{
    \hspace{-8mm}\ExNorm{\vx^{(k)}  - \bvx^{(k)} }\leq  \frac{8\tau \gamma^2 \rho_{\rm max}^2 L^2}{(1-\beta)^2}\sum_{t=m\tau}^{k-1} \Ex\|\vx^{(t)}-\bvx^{(t)}\|^2  + \frac{4\tau \gamma^2 \rho_{\rm max}^2}{(1-\beta)^2}(\sigma^2+ 4\tau b^2), \hspace{5mm} \forall k \geq \tau
}
Taking average over iteration $k$ from $0$ to $T$, we have
\eq{
&\hspace{-4mm}\frac{1}{(T+1)}\sum_{k=0}^{T}\ExNorm{\vx^{(k)}  - \bvx^{(k)} } \nn\\
&\leq  \frac{8\tau \gamma^2 \rho_{\rm max}^2 L^2}{(1-\beta)^2}\frac{1}{T+1}\sum_{k=\tau}^{T} \sum_{t=m\tau}^{k-1} \Ex\|\vx^{(t)}-\bvx^{(t)}\|^2 + \frac{4\tau \gamma^2 \rho_{\rm max}^2}{(1-\beta)^2}(\sigma^2+ 4\tau b^2) \nn\\
&\;\;\;\;\;\;{}+ \frac{1}{(T+1)}\sum_{k=0}^{\tau-1}\ExNorm{\vx^{(k)}  - \bvx^{(k)} } 
}
One key observation is that for arbitrary term $\psi_t$, there exists a non-negative sequence $\left\{d_k\right\}$ which is uniformly bounded by $2\tau$ such that
\eq{
    \sum_{k=\tau}^T\sum_{t=m\tau}^{k-1} \psi_t = \sum_{k=0}^T d_k \psi_k,\;\;\; \forall \psi_t
}
It implies
\eq{
\frac{1}{T+1}\sum_{k=0}^{T}\ExNorm{\vx^{(k)}  - \bvx^{(k)} } \leq & \frac{16\tau^2 \gamma^2 \rho_{\rm max}^2 L^2}{(1-\beta)^2}\frac{1}{T}\sum_{k=0}^{T}  \Ex\|\vx^{(k)}-\bvx^{(k)}\|^2 + \frac{4\tau \gamma^2 \rho_{\rm max}^2}{(1-\beta)^2}(\sigma^2+ 4\tau b^2) \nn\\
&\;\;\;\;\;\;{}+ \frac{1}{(T+1)}\sum_{k=0}^{\tau-1}\ExNorm{\vx^{(k)}  - \bvx^{(k)} } 
}
We can conclude that
\eq{
    \frac{1}{T+1}\sum_{k=0}^{T}\ExNorm{\vx^{(k)}  - \bvx^{(k)} } \leq& \left(1-\frac{16\tau^2 \gamma^2 \rho_{\rm max}^2 L^2}{(1-\beta)^2}\right)^{-1}  \frac{4\tau \gamma^2 \rho_{\rm max}^2}{(1-\beta)^2}(\sigma^2+ 4\tau b^2) \\
    &\;\;\;\;\;\;{}+ \left(1-\frac{16\tau^2 \gamma^2 \rho_{\rm max}^2 L^2}{(1-\beta)^2}\right)^{-1}  \frac{1}{(T+1)}\sum_{k=0}^{\tau-1}\ExNorm{\vx^{(k)}  - \bvx^{(k)} } \nn
}
where the step-size $\gamma$ has to be small enough. Supposing
\eq{
    \frac{16\tau^2 \gamma^2 \rho_{\rm max}^2 L^2}{(1-\beta)^2} \leq \frac{1}{2} \Longrightarrow \gamma\leq\frac{1-\beta}{6 L \tau \rho_{\rm max}} \label{gsdhs}
}
it guarantees that
\eq{
     \frac{1}{T+1}\sum_{k=0}^{T}\ExNorm{\vx^{(k)}  - \bvx^{(k)} } \leq \frac{8\tau \gamma^2 \rho_{\rm max}^2}{(1-\beta)^2}(\sigma^2+ 4\tau b^2) + \frac{2}{(T+1)}\sum_{k=0}^{\tau-1}\ExNorm{\vx^{(k)}  - \bvx^{(k)} } \label{23gjigs}
}
Since $\rho_{\rm max}\leq1$, \eqref{gsdhs} can be further relaxed into the condition $ \gamma\leq\frac{1-\beta}{6 L \tau} $.
$\qed$

As we seen in \eqref{23gjigs}, there is an extra terms of $\sum_{k=0}^{\tau-1}\ExNorm{\vx^{(k)}  - \bvx^{(k)}}$ due to the initial phase. But it is easy to see the impact of this is small since it only contain the initial $\tau$-iterations results and coefficient is diminished by $T$. When the $T$ is large enough, the extra term is almost negligible. Moreover, we can use some warm-up strategy, such as allreduce, that forces all agents' iterates in the first period are the same, i.e. $\sum_{k=0}^{\tau-1}\ExNorm{\vx^{(k)}  - \bvx^{(k)}} = 0$. Under this situation, we immediately obtain the following corollary.

\begin{corollary}
Under the same assumptions as lemma \ref{lm:consensus} and using the all-reduce warm-up strategy at the first $\tau$ iterations, it holds
\eq{
    \frac{1}{T+1}\sum_{k=0}^{T}\ExNorm{\vx^{(k)}  - \bvx^{(k)} } \leq \frac{8\tau \gamma^2 \rho_{\rm max}^2}{(1-\beta)^2}(\sigma^2+ 4\tau b^2) 
}
\end{corollary}
{\bf Proof}. Replacing $\sum_{k=0}^{\tau-1}\ExNorm{\vx^{(k)}  - \bvx^{(k)}}$ by 0 gives the conclusion immediately. $\qed$

\begin{figure}[h!]
    \centering
    \includegraphics[width=0.55\textwidth]{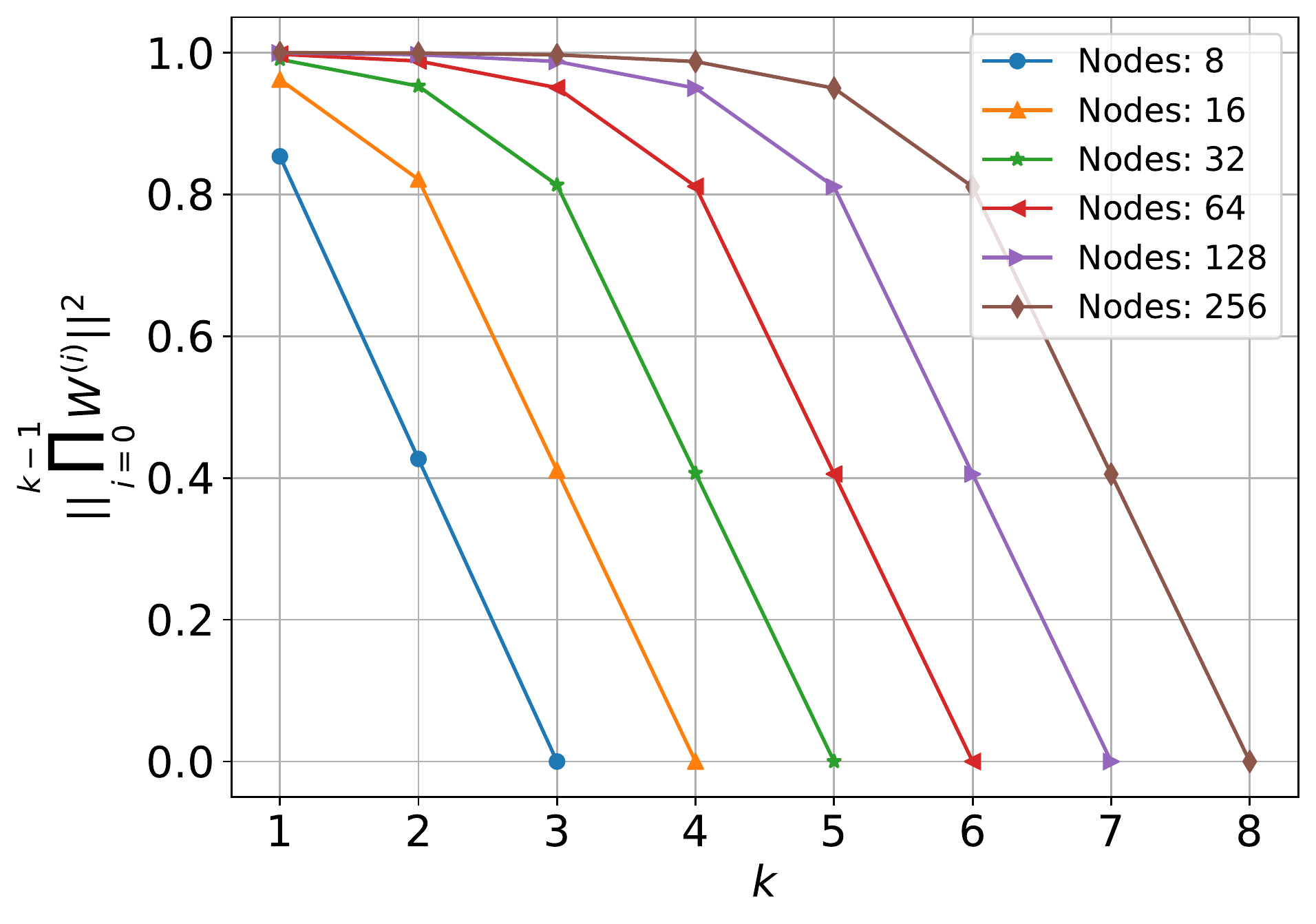}
    \caption{The value of $\|\prod_{i=0}^{k-1}\ww^{(i)}\|^2$ evolves with $k$ over different number of nodes.}\label{fig:dynamic_rho}
\end{figure}

Lastly, we revisit the quantity $\rho_{\rm max}^2$ by numerical experiment here. Looking at the \eqref{239hd2} again, we relax our bounds by simply taking the maximum value of all $\|\prod_{i=t}^{k-1}\ww^{(i)}\|^2_2$. But this value can be much smaller than $\rho_{\rm max}^2$. We just validate them by the numerical experiment in Fig.~\ref{fig:dynamic_rho}.

\subsection{Proof of the convergence Theorem \ref{thm-convergence-one-peer}} \label{apx.sub.converg-theorem}
Finally, we are ready to present the convergence theorem  about the decentralized momentum SGD over one-peer exponential graph. Substituting the conclusion of the descent lemma \ref{lm:descent} into the consensus lemma\ref{lm:consensus}, we immediately establish
\eq{
    \frac{1}{T+1}\sum_{k=1}^T 
    \Ex  \|\nabla f(\bvx^{(k)})\|^2 \leq& \frac{2(1-\beta)}{\gamma(T+1)} (\Ex f(\bvz^{(0)}) -  f^\star ) + \frac{\gamma L }{n(1-\beta)} \sigma^2 + \frac{\beta  L \gamma}{n(1-\beta)^2}\sigma^2  \nn\\
     &\;\;\; + \frac{8\tau \gamma^2 L^2 \rho_{\rm max}^2}{(1-\beta)^2}(\sigma^2+ 4\tau b^2) + \frac{2L^2}{(T+1)}\sum_{k=0}^{\tau-1}\ExNorm{\vx^{(k)}  - \bvx^{(k)} }
}
where the learning rate $\gamma$ requires:
\eq{
     \gamma \leq \min\left\{\frac{1-\beta}{6 L \tau}, \frac{(1-\beta)^2}{2(1+\beta) L} \right\}
}
Simplifying and grouping the terms, we obtain
\eq{
    &\hspace{-12mm}\frac{1}{T+1}\sum_{k=1}^T 
    \Ex  \|\nabla f(\bvx^{(k)})\|^2 \nn\\
    \leq& \frac{2(1-\beta)}{\gamma(T+1)} \left(\Ex f(\bvz^{(0)})
    +\frac{\gamma L^2}{1-\beta}\sum_{k=0}^{\tau-1}\ExNorm{\vx^{(k)}  - \bvx^{(k)} } -  f^\star \right)  \nn\\
     &\;\;\; + \frac{\gamma L }{n(1-\beta)^2} \sigma^2 + \frac{8\tau \gamma^2 L^2 \rho_{\rm max}^2}{(1-\beta)^2}(\sigma^2+ 4\tau b^2)\nn\\
     =& O\left(\frac{(1-\beta)}{\gamma T}\right) + O\left( \gamma \frac{\sigma^2}{n(1-\beta)^2}\right)+  O\left(\frac{\sigma^2\tau\gamma^2}{(1-\beta)^2}\right) + O\left(\frac{b^2 \tau^2\gamma^2}{(1-\beta)^2}\right)
}


If we set the learning rate as $\gamma=O\left(\frac{\sqrt{n(1-\beta)^{3}}}{\sqrt{T}}\right)$, we have
\eq{
    \frac{1}{T}\sum_{k=1}^T  \Ex \|\nabla f(\bvx^{(k)})\|^2  =  
    O\left( \frac{\sigma^2}{\sqrt{(1-\beta) n T} }\right)+  O\left(\frac{n(1-\beta)\sigma^2\tau}{T}\right) + O\left( \frac{ n(1-\beta) b^2 \tau^2}{T}\right) 
}
Last, we derive the transient iteration complexity for the data-homogeneous and data-heterogeneous scenarios, respectively.
\eq{
    \frac{\sigma^2}{\sqrt{(1-\beta) n T}} = \frac{n(1-\beta)\sigma^2\tau}{T}\;\;&\Longrightarrow \;\; T = (1-\beta)^3 n^3 \tau^2 \;\; \mbox{(data-homogeneous)}\\
    \frac{\sigma^2}{\sqrt{(1-\beta) n T}} = \frac{ n(1-\beta) b^2 \tau^2}{T}\;\;&\Longrightarrow \;\; T = (1-\beta)^3 n^3 \tau^4 (b^4 / \sigma^4) \;\;\mbox{(data-heterogeneous)}
}
Absorbing the constants into $\Omega(\cdot)$ notation and replacing $\tau$ by $\log_2(n)$, we establish the transient iteration complexity as stated in Theorem \ref{thm-convergence-one-peer}.
$\qed$

\subsection{Comparison with other commonly-used graphs}
\label{apx.sub.converg-compare}

\subsubsection{Comparison in per-iteration communication and transient iteration (Homogeneous)} 
Table \ref{Talbe-app-trans-iter-homo} summarizes the per-iteration communication and transient iteration complexity of DmSGD with commonly-used topologies. The details of each topology and its associated weight matrix $W$ can be referred to Sec.~\ref{apx.sub.static-compare}. Table \ref{Talbe-app-trans-iter-homo} assumes homogeneous data distributions across all nodes. If the logarithm term can be ignored when $n$ is large, it is observed that both static and one-peer exponential graphs can achieve $\tilde{\Omega}(1)$ per-iteration communication and $\tilde{\Omega}(n^3)$ transient iteration complexity, both of which are nearly best among all compared graphs. Table \ref{Talbe-app-trans-iter-homo} is an extension of Table \ref{tb-main-result} by comparing with more topologies.  

\begin{table}[h!]
\centering
\caption{\small Comparison in per-iteration communication time and transient iteration complexity between decentralized momentum SGD over various commonly-used topologes. The table assumes \textbf{homogeneous} data distributions across all nodes.}
\begin{tabular}{rcc}
\toprule
\multicolumn{1}{c}{\textbf{}}                     & \textbf{Per-iter. Comm.} & \textbf{Transient iter. complexity}  \vspace{0.5mm}\\ \midrule
\textbf{ring}                                     & $\Omega(2)$              & $\Omega(n^7)$                    \vspace{0.5mm}   \\
\textbf{star graph}                               & $\Omega(n)$              & $\Omega(n^7)$                    \vspace{0.5mm}   \\
\textbf{2D-Grid}                                  & $\Omega(4)$              & $\Omega(n^5\log^2_2(n))$         \vspace{0.5mm}   \\
\textbf{2D-Torus}                                 & $\Omega(4)$              & $\Omega(n^5)$                    \vspace{0.5mm}   \\
\textbf{$\frac{1}{2}$-random graph}                         & $\Omega(\frac{n}{2})$    & $\Omega(n^3)$                    \vspace{0.5mm}   \\
\textbf{bipartite random match}                             & $\Omega(1)$              & N.A.                             \vspace{0.5mm}   \\
\textbf{static exponential}                       & $\Omega(\log_2(n))$      & $\Omega(n^3\log_2^2(n))$         \vspace{0.5mm}   \\
\textbf{one-peer exponential} & $\Omega(1)$              & $\Omega(n^3\log_2^2(n))$         \vspace{0.5mm}   \\ \bottomrule
\end{tabular}
\label{Talbe-app-trans-iter-homo}
\end{table}

\subsubsection{Comparison in per-iteration communication and transient iteration (Heterogeneous)} 

Table \ref{Talbe-app-trans-iter-hetero} summarizes the per-iteration communication and transient iteration complexity of DmSGD with commonly-used topologies when data distributions are heterogeneous. Compared to Table \ref{Talbe-app-trans-iter-homo}, it is observed that the transient iteration complexity achieved by each topology in the heterogeneous scenario is typically worse than that in the heterogeneous scenario. Again, if the logarithm term can be ignored when $n$ is large, it is observed that both static and one-peer exponential graphs can achieve $\tilde{\Omega}(1)$ per-iteration communication and $\tilde{\Omega}(n^3)$ transient iteration complexity, both of which are nearly best among all compared graphs.

\begin{table}[h!]
\centering
\caption{\small Comparison in per-iteration communication time and transient iteration complexity between decentralized momentum SGD over various commonly-used topologes. The table assumes \textbf{heterogeneous} data distributions across all nodes.}
\begin{tabular}{rcc}
\toprule
\multicolumn{1}{c}{\textbf{}}                     & \textbf{Per-iter. Comm.} & \textbf{Transient iter. complexity}  \vspace{0.5mm}\\ \midrule
\textbf{ring}                                     & $\Omega(2)$              & $\Omega(n^{11})$                    \vspace{0.5mm}   \\
\textbf{star graph}                               & $\Omega(n)$              & $\Omega(n^{11})$                    \vspace{0.5mm}   \\
\textbf{2D-Grid}                                  & $\Omega(4)$              & $\Omega(n^7\log^4_2(n))$         \vspace{0.5mm}   \\
\textbf{2D-Torus}                                 & $\Omega(4)$              & $\Omega(n^7)$                    \vspace{0.5mm}   \\
\textbf{$\frac{1}{2}$-random graph}                         & $\Omega(\frac{n}{2})$    & $\Omega(n^3)$                    \vspace{0.5mm}   \\
\textbf{bipartite random match}                             & $\Omega(1)$              & N.A.                             \vspace{0.5mm}   \\
\textbf{static exponential}                       & $\Omega(\log_2(n))$      & $\Omega(n^3\log_2^4(n))$         \vspace{0.5mm}   \\
\textbf{one-peer exponential} & $\Omega(1)$              & $\Omega(n^3\log_2^4(n))$         \vspace{0.5mm}   \\ \bottomrule
\end{tabular}
\label{Talbe-app-trans-iter-hetero}
\end{table}

\subsubsection{Exponential graphs endow DmSGD with smaller transient iterations: numerical validation}

In Tables~\ref{Talbe-app-trans-iter-homo}  and \ref{Talbe-app-trans-iter-hetero}, it is observed that exponential graphs endow DmSGD with smaller transient iterations. In this subsection, we validate it with numerical experiments.

We consider a distributed logistic regression problem with each local cost function as 
\eq{
f_i(x) = \frac{1}{M}\sum_{m=1}^M \ln[1 + \exp(-y_{i,m} h_{i,m})^T x ],
}
where $\{h_{i,m}, y_{i,m}\}_{m=1}^M$ are local data samples at agent $i$ with $h_{i,m} \in \mathbb{R}^d$ being the feature vector and $y_{i,m}\in\{+1,-1\}$ being the corresponding label. Each $h_{i,m}$ is generated from the normal distribution $\cN (0; 10 I_d)$. To generate $y_{i,m}$, we first generate an auxiliary random vector $x^\star_{i}\in \RR^d$ with each entry following $\cN(0, 1)$. Next, we generate $y_{i,m}$ from a uniform distribution $\cU(0, 1)$. If $y_{i,m} \le 1/[1 + \exp(- h_{i,m}^T x_i^\star)]$ then $y_{i,m}$ is set as $+1$; otherwise $y_{i,m}$ is set as $-1$. We consider a non-iid scenario in which  $x_i^\star \neq x_j^\star\ \forall i,j$. Each $x_i^\star$ is normalized. We set the number of nodes as 64. 

\begin{figure}[t]
    \centering
    \includegraphics[width=0.55\textwidth]{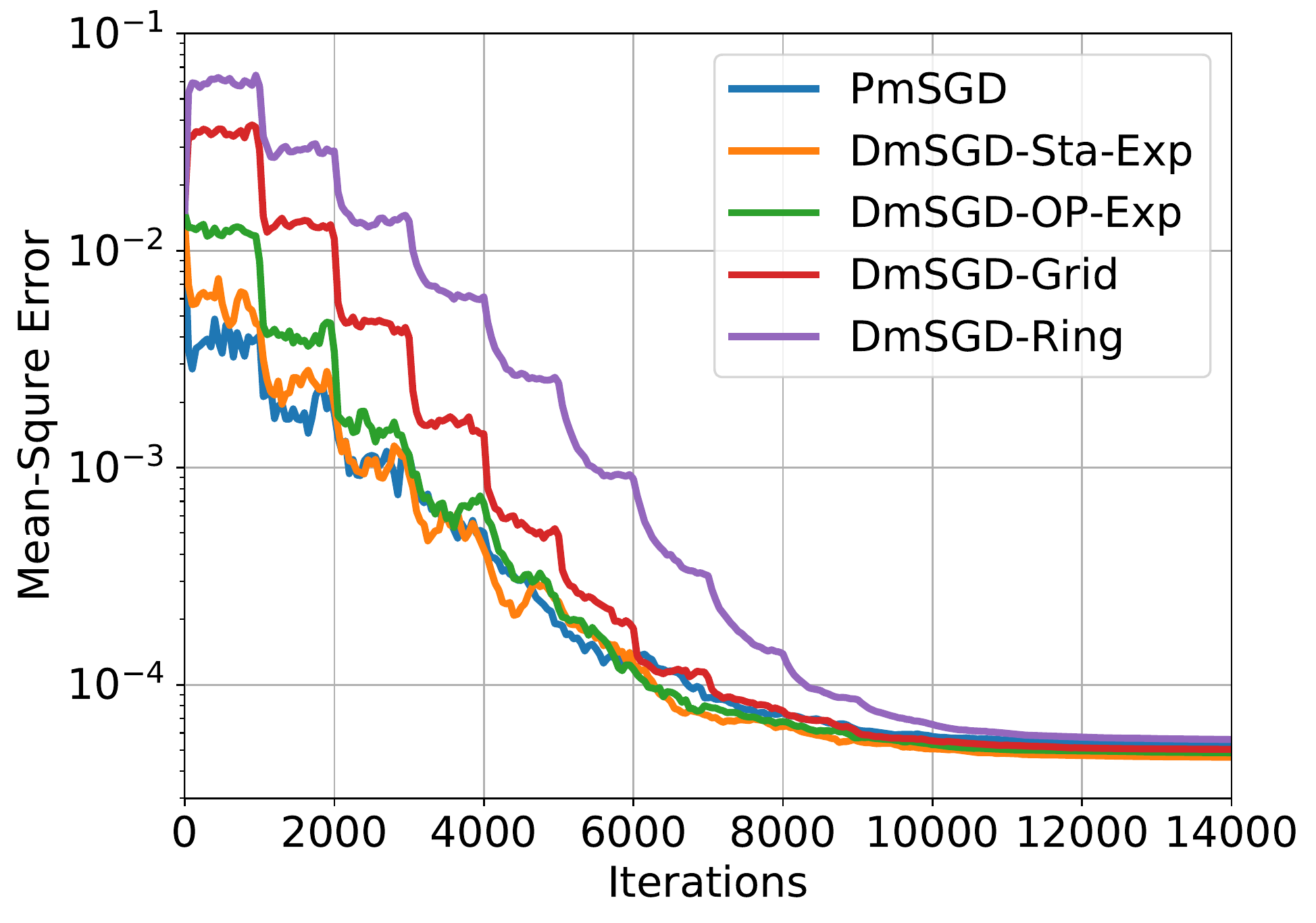}
    \caption{Convergence curves of DmSGD with various topologies. It is observed that DmSGD with exponential graphs has less transient iterations than with other graphs.}\label{fig:tran-iter-compare-d}
\end{figure}

Fig.~\ref{fig:tran-iter-compare-d} illustrates the convergence curves of DmSGD with different topologies as well as parallel momentum SGD (PmSGD). The momentum parameter $\beta = 0.8$. The mean-square-error in the $y$-axis indicates $\frac{1}{n}\sum_{i=1}^n \mathbb{E}\|x_i^{(k)} - x^\star\|^2$. We set $d=10$ and $M=14000$. The step-size $\gamma$ is initialized as $0.2$ and gets decreased by half for every $1000$ iterations.  We repeat all simulations $20$ times and illustrate the mean of all trials. It is observed that DmSGD with static exponential graph converges closely to PmSGD, and it is with the shortest transient iterations. Also, DmSGD with one-peer exponential graph is observed to have slightly longer transient iterations than with the static exponential graph. However, both exponential graphs endow DmSGD with shorter transient iterations than grid and ring. 

\section{More Experiments} \label{apx.experiment} 

\subsection{Details of each topology and the weight matrix} \label{apx.sub.experiment-setup}
The description of the tested topology and its associated weight matrix can be referred to Sec.~\ref{app-details-topo}.

\subsection{DmSGD with exponential graphs when $n$ is not a power of $2$} \label{apx.sub.experiment-not-power2}

Under the same experimental setting as in Sec.~\ref{sec-experiment-implementaion}, this subsection examines the performance of exponential graphs when $n$ is not a power of $2$. As shown in Table~\ref{table-non-integer-log2-nodes-comparison}, one-peer exponential graph can still endow DmSGD with similar, or even better,  training performance compared to its static counterpart. 

\begin{table}[h!]
\caption{\small Comparison of top-1 validation accuracy(\%) when using DmSGD with arbitrary numbers of nodes.}
\begin{center}
\begin{small}
\begin{sc}
\begin{tabular}{ccccc}
\toprule
     nodes & 6(6x8 GPUs) & 9(9x8 GPUs) & 12(12x8 GPUs) & 15(15x8 GPUs) \\
\midrule

static exp. & 76.21 & 75.93 & 75.73 & 76.03 \\
one-peer exp. & 76.16 & 76.17 & 75.85 & 76.19

\\ \bottomrule
\end{tabular}
\end{sc}
\end{small}
\end{center}

\vskip -5mm
\label{table-non-integer-log2-nodes-comparison}
\end{table}

\subsection{Performance with DSGD} \label{apx.sub.experiment-dsgd}

In empirical studies, we conducted a few more experiments to validate how DSGD performs over exponential graphs in deep learning. In Table~\ref{table-dsgd-comparison}, we repeated the same experiment in Table~\ref{table-topo-nodes-comparison} except for the parameter setting $\beta=0$ i.e. eliminating the influence of momentum. It is observed that: 
\begin{itemize}[leftmargin=20pt]
\vspace{-2mm}
\item The accuracy performance of all DSGD scenarios has dropped over 7\% compared to the DmSGD scenarios. This highlights the critical role of the momentum in DSGD for real deep learning experiments. 
\item DSGD over the one-peer exponential graph achieves similar accuracy as the static exponential graph, and both topologies enable DSGD with higher accuracy than the ring topology. This is consistent with the two conclusions listed above. 
\item The training time of DSGD over different topologies is similar to DmSGD listed in Table 2, and we, therefore, omitted it in the following table.
\end{itemize}

\begin{table}[h!]
\caption{\small Comparison of top-1 validation accuracy(\%) when using DSGD with different topologies.}
\begin{center}
\begin{small}
\begin{sc}
\begin{tabular}{cccc}
\toprule
     nodes & 4(4x8 GPUs) & 8(8x8 GPUs) & 16(16x8 GPUs) \\
\midrule
ring & 68.85 & 68.62 & 68.78 \\
static exp. & 69.08 & 68.81 & 68.79 \\
one-peer exp. & 69.01 & 68.94  & 68.85

\\ \bottomrule
\end{tabular}
\end{sc}
\end{small}
\end{center}

\vskip -5mm
\label{table-dsgd-comparison}
\end{table}

\subsection{Example code for implementation}

For the implementation of decentralized methods, we utilize BlueFog, which is an open-source high-performance decentralized deep training framework, to facilitate the topology organization, weight matrix generation, and efficient partial averaging.

\begin{lstlisting}[label={lst.na},
caption={\small Neighbor allreduce functionality for communication.}]
def neighbor_allreduce(tensor: torch.Tensor,
                          self_weight: float,
                          src_weights: Dict[int, float],
                          dst_weights: Dict[int, float]) ->   torch.Tensor:
\end{lstlisting}

One major functionality for decentralized communication is \emph{neighbor\_allreduce}, as listed in Listing \ref{lst.na}, implementing the following equation.
\eq{
  x^{(k+1)}_i = w_{ii}x^{(k)}_i + \sum_{j \in \mathcal{N}_i \backslash i} w_{ij} x^{(k)}_j,
}
The argument \emph{self\_weight} stands for $w_{ii}$, and $w_{ij}$ for communication with the other node $j$ is achieved by either using \emph{src\_weights} in pull-mode or using \emph{dst\_weights} in push-mode.

Listing \ref{lst.exp2} gives two utility functions for one-peer exponential graphs generation. For each node, each call of function \emph{GetOnePeerExpGraphGenerator} provides the one-peer nodes connection information for communication. Passing this connection information to \emph{neighbor\_allreduce} is achieved through updating the member variables of decentralized optimizer in \emph{UpdateOnePeerExpGraph}.

\begin{lstlisting}[label={lst.exp2},
caption={\small Utility functions for the generation of one-peer exponential graphs.}]
# One-peer exponential graph generation
def GetOnePeerExpGraphGenerator(size, self_rank):
    tau = math.ceil(math.log2(size))  # Periodic cycle
    index = 0
    while True:
        send_rank = (self_rank + 2**index) % size
        recv_rank = (self_rank - 2**index) % size
        yield send_rank, recv_rank
        index += 1
        index = index % tau

# Graph update in each iteration
one_peer_exp_graph_gen = GetOnePeerExpGraphGenerator(bf.size(), bf.rank())
def UpdateOnePeerExpGraph(optimizer):
    dst_rank, src_rank = next(one_peer_exp_graph_gen)
    optimizer.dst_weights = {dst_rank: 1.0}
    optimizer.src_weights = {src_rank: 0.5}  # Corresponds to W matrix
    optimizer.self_weight = 0.5    # Corresponds to the diagonal of W matrix
\end{lstlisting}

With that, Listing \ref{lst.train} shows a simplified code for model training. The overall code structure is similar as the traditional model training script, with few modifications for decentralized environment. On line 5, a decentralized optimizer wraps the original SGD optimizer. Under the hood, it registers the  neighbor\_allreduce communication function through the hook mechanism. On line 11, the one-peer exponential graphs get updated in each iteration. After the model forward propagation is computed locally, the backward propagation is performed on line 19. Meanwhile, it also triggers communication using \emph{neighbor\_allreduce}. In order to boost the training efficiency, the time of computation and communication are overlapped as much as possible through the multi-threading. Finally, line 21 updates the model until the communication finishes.

\begin{lstlisting}[label={lst.train},
caption={\small Example of how to train a model using a DmSGD optimizer under a one-peer exponential graph. The communication graph is updated in each iteration.}]
import bluefog.torch as bf
...  # Model and data preparation
# Generate decentralized optimizer
optimizer = optim.SGD(model.parameters(), lr=lr, momentum=momentum)
optimizer = DmSGDOptimizer(optimizer, model=model)
...
for epoch in range(num_epochs):
    # Training the model 
    for data, target in train_loader:
        # Graph update in each iteration
        UpdateOnePeerExpGraph(optimizer)
        data, target = data.cuda(), target.cuda()
        optimizer.zero_grad()
        # Local forward propagation
        output = model(data)
        loss = F.cross_entropy(output, target_batch)
        # Local backward propagation
        # Meanwhile triggering neighbor_allreduce communication
        loss.backward()
        # Model update and wait until the communication finishes
        optimizer.step()
    # Validation
    ...  
\end{lstlisting}